\pdfoutput=1
\documentclass[journal]{IEEEtran}

\usepackage{amsmath,graphicx}
\usepackage{booktabs}
\usepackage{multirow}
\usepackage{xcolor}
\usepackage{amsfonts}
\usepackage[inline]{enumitem}
\usepackage{mathtools}
\usepackage{amsthm}
\usepackage{amssymb}
\usepackage{pgfplots}
\usepackage{tikz}
\usetikzlibrary{arrows,automata,positioning}
\usepackage{caption}
\usepackage{subcaption}
\usepackage{relsize}
\usepackage{url}
\usepackage[hidelinks]{hyperref}   % AMH: check
\usepackage[%
    capitalize,%
    noabbrev,%
]{cleveref}
\usepackage{soul}
\usepackage{siunitx}

\newcommand{\ci}[1]{{\color{black}#1}}

\newcommand{\TNS}[1]{{\color{red} TNS: #1}}

\newcommand{\edit}[1]{{\color{black}#1}}

\usepackage{cite}
\usepackage{url}
\usepackage{amsmath,amsfonts}
\usepackage{booktabs}
\usepackage{multirow, makecell}
\usepackage{footmisc}
\usepackage{pifont}
\usepackage{textcomp}
\usepackage{xcolor}
\usepackage{array}
\usepackage{tabu}

\usepackage{threeparttable}
\usepackage{comment}

\definecolor{customgreen} {RGB}{217	234	212}
\definecolor{customblue}  {RGB}{205	226	242}
\definecolor{customorange}{RGB}{254	228	207}
\definecolor{customred}{RGB}{222 157 155}
\definecolor{sharedcolour}{RGB}{254	228	207}

\newcommand{\norm}[1]{\left\lVert#1\right\rVert}

\newcommand{\kmeans}{k-means}
\newcommand{\lone}{\ensuremath{L_1}}

\newcommand{\tabincell}[2]{\begin{tabular}
{@{}#1@{}}#2\end{tabular}}

\begin{document}
\title{Self-Supervised Speech Representation Learning:\\ A Review}
%\title{Speech and Audio representation learning: A review}
\author{Abdelrahman Mohamed*, Hung-yi Lee*, Lasse Borgholt*, Jakob D. Havtorn*, Joakim Edin, Christian Igel\\Katrin Kirchhoff, Shang-Wen Li,  Karen Livescu, Lars Maaløe, Tara N. Sainath, Shinji Watanabe\thanks{*equal contribution, order is random, remaining sorted alphabetically.}
\thanks{Abdelrahman Mohamed and Shang-Wen Li are with Meta (e-emails: abdo@fb.com, shangwel@fb.com).}
\thanks{Hung-yi Lee is with the Department of Electrical Engineering and the Department of Computer Science \& Information Engineering of National Taiwan University (e-mail: hungyilee@ntu.edu.tw).}
\thanks{Lasse Borgholt is with Corti AI and the Department of Computer Science, University of Copenhagen, Denmark (e-mail: lb@corti.ai).} % lb@corti.ai -> Please use this e-mail to communicate.
\thanks{Jakob D. Havtorn and Lars Maaløe are with Corti AI and the Department of Applied Mathematics and Computer Science, Technical University of Denmark, Denmark (e-mails: jdha@dtu.dk, lm@corti.ai).}
\thanks{Joakim Edin is with Corti AI, Denmark (e-mail: je@corti.ai).}
\thanks{Christian Igel is with the Department of Computer Science, University of Copenhagen, Denmark (e-mail: igel@di.ku.dk).}
\thanks{Katrin Kirchhhoff is with AWS AI Labs, Amazon, Seattle, 98121, USA (email: katrinki@amazon.com).}
\thanks{Karen Livescu is with the Toyota Technological Institute at Chicago, Chicago, IL 60615 USA (e-mail: klivescu@ttic.edu).}
\thanks{Shinji Watanabe is with the Language Technologies Institute, Carnegie Mellon University, Pittsburgh, PA 15213 USA (e-mail: shinjiw@ieee.org).}

}

\maketitle

\begin{abstract}
Although supervised deep learning has revolutionized speech and audio processing, it has necessitated the building of specialist models for individual tasks and application scenarios. It is likewise difficult to apply this to dialects and languages for which only limited labeled data is available. Self-supervised representation learning methods promise a single universal model that would benefit a wide variety of tasks and domains. Such methods have shown success in natural language processing and computer vision domains, achieving new levels of performance while reducing the number of labels required for many downstream scenarios. Speech representation learning is experiencing similar progress in three main categories: generative, contrastive, and predictive methods. Other approaches rely on multi-modal data for pre-training, mixing text or visual data streams with speech. Although self-supervised speech representation is still a nascent research area, it is closely related to acoustic word embedding and learning with zero lexical resources, both of which have seen active research for many years. This review presents approaches for self-supervised speech representation learning and their connection to other research areas. Since many current methods focus solely on automatic speech recognition as a downstream task, we review recent efforts on benchmarking learned representations to extend the application beyond speech recognition. 
\end{abstract}

\begin{IEEEkeywords}
Self-supervised learning, speech representations.
\end{IEEEkeywords}

\IEEEpeerreviewmaketitle

\section{Introduction}

%Hung-yi Lee: I think maybe we need a basic frame image. I happen to have an image that can be used here.
\begin{figure}
    \centering
    \includegraphics[width=0.40\textwidth]{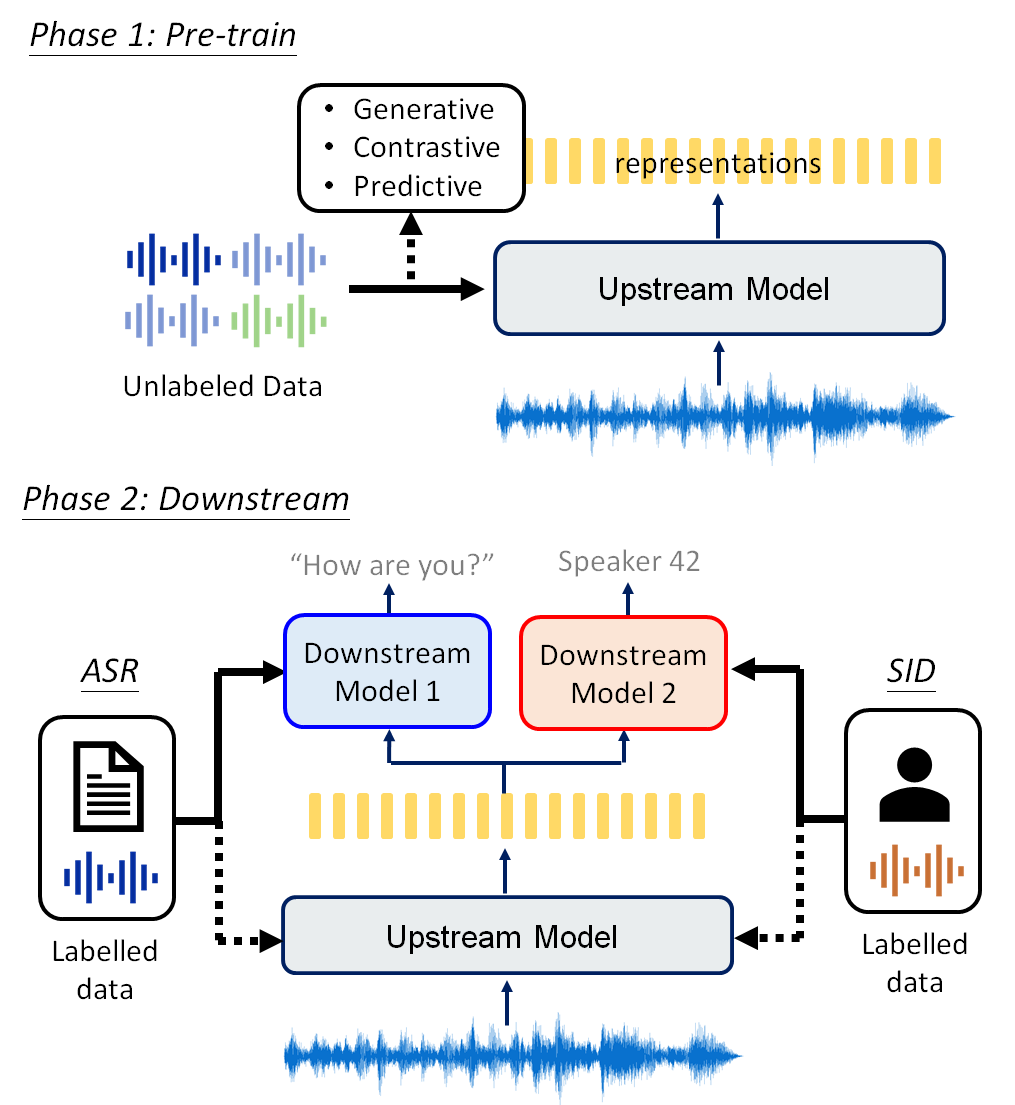}
	 \caption{Framework for using self-supervised representation learning in
	 downstream applications}
    \label{fig:SSL_framework}
\end{figure}

% {\color{blue} Reviewer: Daniel, KL, SW}\\
% {\color{blue} Abdo -- this section is a work-in-progress}\\

% {\color{red} Hung-yi: Last time we have decided to use "pre-train" (with "-")} -- Done\\

% {\color{red} Hung-yi: Fine-tune? or finetune?} --Fine-tune Done\\

% {\color{red} Hung-yi: Fig. 1 or Figure 1} -- Fig. 1  Done\\

% \abdo{comment from Shinji: bring references to intro.}

% \kl{Comments from Nov. 10 meeting: 
% \begin{itemize}
%     \item \abdo{added.} would be good to define representation learning a bit more thoroughly in the intro, and mention that we are considering both discrete and continuous representations.
%     \item \abdo{added.} it would also be good to mention and broadly define "self-supervised" in the intro (as of now it's first mentioned in Section~\ref{sec:thirdwave}), and mention that the term is now used (and we use it) for some techniques that in the past would have just been called "unsupervised".
% \end{itemize}}

% \kl{New comments: 
% \begin{itemize}
%     \item It would be good to include somewhere in the paper a summary of tasks for which SSL pre-trained models have the SOTA performance.  Maybe it is in the paper already but I don't see it.
% \end{itemize}}

Over the past decade, deep learning approaches have revolutionized speech processing
through a giant leap in performance, enabling various real-world applications.
Supervised learning of deep neural networks has been the cornerstone of this
transformation, offering impressive gains for scenarios rich in labeled
data~\cite{lecun2015deep, hinton2012deep, bourlard2012connectionist}. 
Paradoxically, this heavy reliance on supervised learning has restricted progress in
languages and domains that do not attract the same level of labeling
investment. 
%Paradoxically, the heavy reliance on supervised learning bottlenecked the development of novel industrial speech applications and restricted progress in languages and domains that do not entertain the same level of labeling investment. 

To overcome the need for labeled data, researchers have explored approaches that use
unpaired audio-only data to open up new industrial speech use-cases and
low-resource languages~\cite{Kemp1999, csl01_limsi, ma_bbn_06}. Inspired by how
children learn their first language through listening and interacting with
family and surroundings, scientists seek to use raw waveforms and
spectral signals to learn speech representations that capture low-level
acoustic events, lexical knowledge, all the way to syntactic and semantic
information. These learned representations are then used for target downstream
applications requiring a minimal number of labeled data~\cite{Hinton_2007,
LeCun06atutorial, bengio_representation_2013}. 
Formally, representation learning refers to algorithms for extracting latent
features that capture the underlying explanatory factors for the observed
input~\cite{bengio_representation_2013}. 
% Previous research efforts explored latent representations that are binary \cite{}, discrete \cite{} and continuous \cite{}. 
% These latent representations are generally used as inputs to a supervised predictor or a generation module. The whole network generating these representations is frequently used to initialize a supervised classifier~\cite{hinton_2006}. 

%Learning representations approaches generally fall within \textit{unsupervised learning}, which refers to the family of machine learning methods that discover naturally occurring patterns in the training samples without any pre-assigned labels or scores associated with these samples~\cite{jordan_2015}. 
Representation learning approaches are generally considered examples of \textit{unsupervised
learning}, which refers to the family of machine learning methods that discover
naturally occurring patterns in training samples for which there are no pre-assigned
labels or scores~\cite{jordan_2015}. 
The term ``unsupervised'' is used to distinguish this family of methods from
``supervised'' approaches, which assign a label to each training sample, and
``semi-supervised'' approaches, which utilize a small number of training samples
with labels to guide learning using a larger volume of unlabeled samples.
Examples of unsupervised learning techniques include \kmeans{}
clustering~\cite{vq}, mixture models~\cite{MoE}, autoencoders~\cite{hinton_94},
and non-negative matrix factorization~\cite{nmf}. 
\textit{Self-supervised learning} (SSL) is a fast-growing subcategory of
unsupervised learning approaches, which are techniques that utilize
information extracted from the input data itself as the label to learn
representations useful for 
% one or many 
downstream tasks. {\color{black} For example, unsupervised k-means clustering doesn't adhere to this definition of self-supervision since it iteratively minimizes the within-cluster variance during learning.}
In this review, we
focus on self-supervised learning approaches.

\Cref{fig:SSL_framework} outlines self-supervised representation learning in
relation to downstream applications. 
There are two stages in this framework.
In the first stage, we use SSL to pre-train a \textit{representation model},
also called an \textit{upstream model} or a \textit{foundation model}.
In the second stage, downstream tasks use either the learned
representation from the frozen model, or fine-tune the entire pre-trained model
in a supervised phase~\cite{hinton_2006}. 
%e.g., Automatic Speech Recognition (ASR) and Speaker Identification (SID) are two example downstream applications in \cref{fig:SSL_framework}. 
Automatic speech recognition (ASR) and speaker identification (SID) are 
examples of downstream applications in \cref{fig:SSL_framework}.
% There are some task-specific labeled data for each downstream task, and there is a downstream model on top of the upstream model. The labeled data is used to train the corresponding downstream models and optionally fine-tune upstream models.
% If the pre-training stage only leverages unlabeled data, the framework in \cref{fig:SSL_framework} is named self-supervised learning (SSL), which is the focus of this review paper.
% Pre-training with labeled data is out of the scope. Before the term self-supervised learning was widely used, pre-training without any labeled data was usually known as an \texit{unsupervised learning} approach. 

%Lee: in the introduction, we may need a specific example. Below is just an example.
%With unsupervised ASR, imagine that in the future, when an indigenous language family buys an intelligent assistant, even if the assistant does not support the language at the beginning, it would automatically learn to transcribe the new language after learning.

% The speech representation learning approaches evolved over the past few decades through at least three major stages, as discussed in \cref{sec:thirdwave}; static and temporal clustering algorithms \cite{}, energy-based neural models \cite{}, and more recently, self-supervised learning techniques \cite{}. 
% Across all these three waves of representation learning, three main desirable characteristics emerged; features need to be disentangled, invariant, and hierarchical. 
It is considered desirable for learned speech representations to be
disentangled, invariant, and hierarchical.
Since spoken utterances contain much richer information than the corresponding text
transcriptions---e.g., speaker identity, style, emotion, surrounding noise, and
communication channel noise---it is important to learn representations that
disentangle these factors of variation. Furthermore, invariance of the learned
features to changes in background noise and in the communication channel ensures
stability with respect to downstream application scenarios. Learning feature
hierarchies at the acoustic, lexical, and semantic levels supports applications
with different requirements. For instance, whereas a speaker identification task
benefits from a low-level acoustic representation, a speech
translation task requires a more semantic representation of the input
utterance. 

Due to the popularity of SSL, reviews have been published about the
technology in general~\cite{bommasani2021opportunities,ericsson2021selfsupervised,LiuSSLsurvey} as well as its application to natural language processing (NLP)~\cite{rogers-etal-2020-primer,liu2021pretrain,xia-etal-2020-bert,QiuSSLNLPsurvey} and computer vision (CV)~\cite{JingSSLCVsurvey}. \edit{Recently, a brief overview with a general focus on speech representation learning was published \cite{borgholt_22}}. However, none of these overviews focus \edit{exclusively} on SSL for speech processing. Since the speech signal differs greatly from image and text inputs, many  theories  and technologies have been developed to address the unique challenges of speech. One review addresses speech representation learning based on deep learning models~\cite{latif2021deep}, but does not address recent developments in self-supervised learning. This motivates this overview of speech SSL.

%For survey papers on TL to speech processing, please check the reference~\cite{survey_speech_TL}.
%The representation learning methods covered in this review are primarily task-agnostic methods that only benefit from supervised data at the fine-tuning stage, as shown in \cref{}. %Lee: Similar sentence may have been mentioned in the paragraph describing the figure.

The structure of this paper is arranged as follows. \Cref{sec:thirdwave}
briefly reviews the history of speech representation learning, and
\cref{sec:approach} reviews current speech SSL models.
\Cref{section:benchmark} surveys SSL datasets and benchmarks, 
and discusses and compares results from different works. \Cref{analysis}
analyzes successful SSL approaches and offers insights into the
importance of technological innovations. \Cref{sec:zero} reviews 
zero-resource downstream tasks that utilize SSL. 
Finally, \cref{sec:conclusion} summarizes the paper and suggests 
future research directions.

 %Reviewer: Daniel, KL, SW
\section{Historical Context of Representation Learning}

\label{sec:thirdwave}

% {\color{blue} Reviewer: KL, KK, Daniel }\\
In this section we present the historical background of the current surge in
self-supervised representation learning methods in the context of two previous
waves of research work in the 1990s and 2000s. The discussed approaches go
beyond speech to describe the overall landscape of machine learning development during the
past few decades.

%%KK: if you adopt the paragraph below I suggest renaming this as "Features based on generative models" or something
\subsection{Clustering and mixture models}

Initial research in learning latent speech and audio representations involved
simple models in which the training data likelihood was optimized directly
or via the expectation--maximization (EM) algorithm.

Early work used simple clustering methods. For example, in work such
as \cite{Rabiner1979,Wilpon1985}, word patterns were clustered
semi-automatically using techniques such as \kmeans, after which isolated words
were recognized by finding the training cluster closest to 
  the test data.   % AMH: check

Through time, modeling techniques improved such that subword units were
represented by Gaussian mixture models (GMMs) \cite{Gauvain1994}, which facilitated
the modeling of more variability in the input data. GMMs were first built for
context-independent phonemes; state-clustering 
algorithms~\cite{Young1994} then resulted in GMMs for context-dependent phonemes. 
Each latent component of these mixture models acted as a template of a
prototypical speech frame, 
% leading to their deficiency to deal with 
  making it difficult to handle             % AMH: check
large volumes of data with diverse characteristics. 
Furthermore, dynamical models like hidden Markov models (HMMs)~\cite{Bahl1986}
allowed for the processing of continuous speech rather than just isolated word
recognition. These generative GMM and HMM models were trained by maximizing the
likelihood of data given the model, which could be accomplished in either an
unsupervised or a supervised manner.

Another line of research focused on extracting speech features from generative
models. The main objective here was to render the knowledge learned by generative
models accessible to discriminative downstream classifiers, or to map
variable-length sequences to fixed-length representations. Feature vectors
were derived from the parameters of trained GMM models. In the case of {\em
Fisher vectors}, the features were the normalized gradients of the
log-likelihood with respect to the model parameters (mixture weights, means, and
variances) of the Gaussian mixtures. An extension of this approach (likelihood
ratio score space) used the derivative of the log-likelihood ratio of two
models, e.g., a background model and a foreground model. Examples of their use
in speech processing include speech recognition~\cite{smith01,venkata03} and
speaker recognition~\cite{wan03}. Subsequent techniques in speaker and language
verification~\cite{dehak11a,dehak11b} similarly extracted parameters
(concatenated means) from trained background GMMs as representations that were
then combined with low-rank projections of speaker/session- or language-specific
vectors.

\subsection{Stacked neural models}
\label{subsec:stack}

%{\color{blue} Tara}\\
More recently, representation learning has seen a shift of focus towards neural models,
which, compared to GMMs and HMMs, offer distributed representations with more
capacity to model diverse input signals into efficient latent binary codes. 
Examples of early techniques include restricted Boltzmann machines
(RBM)~\cite{hinton_2006}, denoising autoencoders~\cite{DAE}, noise contrastive
estimation (NCE)~\cite{gutmann2012noise}, sparse coding~\cite{Olshausen1996,
sparse_lee, sivaram2010sparse}, and energy-based methods~\cite{Ranzato2007}.
Many of these techniques have also been applied to CV and NLP problems, which
provided inspiration for their application to speech.
%Examples of techniques include Restricted Boltzmann Machine (RBM) \cite{hinton2012practical,Srivastava2013}, Neural Auto Encoders \cite{Lecun2012}, Noise Contrastive Estimation (NCE) \cite{chen2015recurrent},  Sparse Coding \cite{sivaram2010sparse} and Energy-based method \cite{Ranzato2007}. Note that many of these techniques were also tried for CV and NLP problems, which provided inspiration to try them for ASR as well.

Higher-capacity neural models were achieved by stacking several neural network
layers to build progressively higher-level concept representations.
However, these deeper networks also increased the training complexities. For
example, approximate training methods such as contrastive divergence
\cite{hinton_cd_2002} were a practical technique to 
% make       RBM training more efficient. 
  streamline RBM training.                   % AMH: check
Furthermore, deep networks had non-convex objective functions, which
often resulted in long training times compared to GMMs, which are trained
using full batches instead of mini-batch learning.

% http://proceedings.mlr.press/v2/ranzato07a/ranzato07a.pdf, 
%https://auai.org/uai2013/prints/papers/166.pdf, % https://www.cs.cmu.edu/~rsalakhu/papers/neco_DBM.pdf
% problems with this trend: approximate gradient, failure modes, long training time, no clear wins in downstream applications. 

\subsection{Learning through pretext task optimization}

% {\color{blue} Hung-yi + Abdo}\\

%Note on 12/09: We will mainly describe the speech part but link to CV and NLP. 
%discuss instance classification, contrastive losses,\\

%Currently, the network learns a function that maps input to desired representations by solving a \textit{pretext} task.
%After pre-training, learned representation models could be applied to downstream tasks through feature-based representation extraction or fine-tuning as part of the downstream model.  
%By optimizing carefully designed pretext tasks, neural models can capture latent representations which generalize well across a wide range of downstream applications, usually under few-shot or even zero-shot conditions. 
A more recent trend is learning networks that map the input to desired
representations by solving a \textit{pretext} task. Such studies have several
characteristics:
(1)~All layers are trained end-to-end to optimize a single pretext task instead
of relying on layer-wise pre-training
(2)~Past stacked networks typically had only a few layers, but 
very deep networks with more than ten layers are now common.
(3)~It is common to evaluate a representation model on a wide range of tasks.
For example, in NLP, a representation model is usually assessed on GLUE,
which comprises nine tasks~\cite{GLUE}, whereas in speech, a representation model can be
evaluated on SUPERB, which comprises ten tasks~\cite{yang21c_interspeech}, 
as described in detail in \cref{sec:benchmark}.

The cornerstone of this third wave is the design of a pretext task, which
allows the model to efficiently leverage knowledge from unlabeled data.
The pretext task should be challenging enough for the model to learn high-level
abstract representations and 
% not be too amiable to exploit low-level shortcuts.
  not be so easy as to encourage the exploitation of low-level shortcuts.  % AMH: check
Early breakthroughs included end-to-end learning of deep neural architectures
via pretext tasks for restoring the true color of black-and-white
images~\cite{colorizing}, joint learning of latent representations and their
cluster assignments~\cite{deepcluster}, and the prediction of the relative positions of
image patches~\cite{context_pred}. Other popular approaches include variational
autoencoders (VAEs)~\cite{vae, rezende2014stochastic}. \edit{While typical autoencoders learn data
representations using unsupervised objectives by reconstructing the input
after passing it through an information bottleneck, VAEs estimate a neural model of a probability density function (pdf) that approximates the unknown “true” distribution of the observed data, for which we only have access to independently identically distributed (iid) samples. It is also important to mention dynamical VAEs \cite{Girin2021}, which is an extension of VAE for sequential data such as speech.}

In the SSL context, a pretext task related to autoencoding is to generate
an object from its partial information. Such tasks are widely used in NLP, for
example, using 
the previous tokens 
in a sentence to predict the next token such as in
ELMo~\cite{peters2018deep}, the GPT
series~\cite{alex2019GPT2}, and %brown2020gpt3 alex2018GPT,
Megatron~\cite{shoeybi2020megatronlm}, or predicting the masked tokens in a
sentence such as with the bidirectional encoder representations from Transformers
(BERT) series~\cite{jacob2019BERT,Liu2019RoBERTa}. %,albert
%ELMo~\cite{peters2018deep}, GPT series~\cite{alex2018GPT,alex2019GPT2,brown2020gpt3}, Megatron~\cite{shoeybi2020megatronlm} are gigantic LM which predict the next token. 
%BERT~\cite{jacob2019BERT}, RoBERTa~\cite{Liu2019RoBERTa}, ALBERT~\cite{albert} masked the input token, and the models are learned to predict the masked tokens. 
Another common pretext task in the third wave is contrastive
learning~\cite{oord2018representation}, in which a model learns to identify a
target instance from a set of negative samples. 
This approach has become especially popular in the
CV context~\cite{pmlr-v119-chen20j,he2020momentum,chen2020improved,SwAV}.
In this survey, we will mainly focus on techniques for pretext task
optimization for speech processing, and discuss these techniques in detail
in \cref{sec:approach}.

\subsection{Other related work}
A closely related area of research that is not covered in this review is
semi-supervised pre-training methods such as pseudo-labeling (that is,
self-training). Pseudo-labeling (PL) relies on a supervised teacher model to
label a large volume of speech-only data, which is then used to augment the
initial labeled data to train a student model~\cite{Kemp1999, csl01_limsi,
ma_bbn_06, hari_1mhour}. PL has been successful and widely adopted in the
speech community since the 1990s. Other proposed variations of PL include
augmenting speech-only data with noise to improve robustness, iterating
over the PL process to improve teacher labeling quality, and training student
models with more parameters than their original teachers to capture the
complexities in vastly larger speech-only data~\cite{park2020improved,
xu2020iterative, xiao_scaling_2021}. 
Both SSL and PL leverage unlabeled speech-only data.
One distinguishing factor in PL is the utilization of supervised data for a
specific task during model pre-training, which limits the model's focus to
a single (or at best a few) downstream tasks. 
SSL, in turn, is an attempt to learn task-agnostic representations to benefit
a wide range of tasks. 

Transfer learning (TL) is another closely related area of research for
pre-training speech models. TL transfers knowledge captured by models
trained on one task to different but related tasks~\cite{caruana1997multitask}. 
The past few decades have seen active research on TL and its
extension to multitask learning for more general representations. 
Multilingual and cross-lingual supervised models have proven superior in
low-resource speech recognition tasks~\cite{Cui2015MultilingualRF}.
SSL can be regarded as a type of TL because knowledge learned from pre-training
is used for different downstream tasks.
This survey paper focuses on SSL, and not all TL technologies for speech. 
One survey indeed addresses TL for speech processing~\cite{survey_speech_TL}
but does not include current SSL technologies for speech.  

\section{Speech Representation Learning Paradigms} \label{sec:approach}
% {\color{blue} Reviewer: Hung-yi, Daniel , Abdo}\\
% {\color{blue} Hung-yi + Abdo}\\
%Start by discussing central challenges for self-supervised representation learning in speech: 1)Can't do instance classification in variable-length sequences. 2)No pre-defined vocabulary. 3)No pre-defined segmentation. \\
%New approaches offer deep contextualization of learned representations. \\
% {\color{red} first version for a list of symbols to use -- let's keep updating it as needed -- Abdo}\\
Due to the characteristics of speech, SSL pretext tasks developed for CV
and NLP may not directly apply to speech.
Below we summarize the characteristics of speech as compared to CV and NLP.
\begin{itemize}
\item \textit{Speech is a sequence.} 
Unlike CV, in which an image usually has a fixed size representation, it is
natural to represent a speech utterance as a variable-length sequence. 
Therefore, pretext tasks developed for CV cannot generally be directly
applied to speech.

\item \textit{Speech is a long sequence without segment boundaries.} 
Both text and speech can be represented as sequences. From this viewpoint, it
is natural to apply learning approaches developed for text directly to
speech. 
% However, the sequence representations of speech are usually much longer than those of text and they contain no natural boundaries at which they can be split into segments.
In NLP, morpheme-like tokens are widely used as sequence units in
pre-training. The standard BERT takes 512 morpheme-like tokens as
input, usually covering a paragraph including several sentences. 
% In speech processing, the audio signal is recorded as a waveform represents the amplitude of pressure. 
However, speech signals consist of sound pressure measurements with thousands
of samples per second, resulting in sequences much longer than those for text. Even
spectral representations which reduce the sequence length can have hundreds of
frames per second.
%This raw signal is often converted to a spectral representation where each frame represents 20-25 milliseconds of audio computed with overlaps at 10 millisecond strides.
% a frame is the typical sequence unit in speech, and each frame usually represents a length of 10 milliseconds. 
% Therefore, an utterance is usually represented as a long sequence with more than a thousand frames. 
Processing such sequences with typical neural network architectures like
Transformers can result in problems with running time and memory requirements. 
One could gather consecutive frames to form shorter segments,
but unlike text, there is no obvious segmentation for unlabeled
speech.
\item \textit{Speech is continuous.} 
In NLP, it is common to use a pretext task that models a categorical
distribution of masked or future inputs. Since text is easily broken down into
individual tokens such as words, subwords, or characters, it is
straightforward to define a finite vocabulary for such tasks.
%The generative-based pretexts are widely applied in NLP.
%Because a sentence is composed of tokens, sentence generation can be formulated as a series of classification tasks.  
However, this idea does not apply to speech modeling because speech
signals are continuous; 
% thus, such a vocabulary is unavailable.
  in this sense there is no such thing as a speech vocabulary.   % AMH: check
\item \textit{Speech processing tasks are diverse.}
Building generalizable self-supervised representation models for diverse speech
processing tasks is challenging.
Speech contains rich, hierarchical information, and different speech tasks
may require mutually orthogonal information.
For example, speech recognition requires a model that extracts content information
but ignores speaker information; in contrast, speaker recognition 
requires a model that extracts speaker information but removes content information.
Therefore, it is challenging to define a self-supervised model whose
representations are suitable for both speech recognition and speaker
recognition. Analogous considerations apply within CV and NLP.
\end{itemize}
%(Although we list the problems here, we do not highlight the solutions in the following paragraph.) 

\begin{figure*}[t!]
    \centering
    \includegraphics[width=0.98\textwidth]{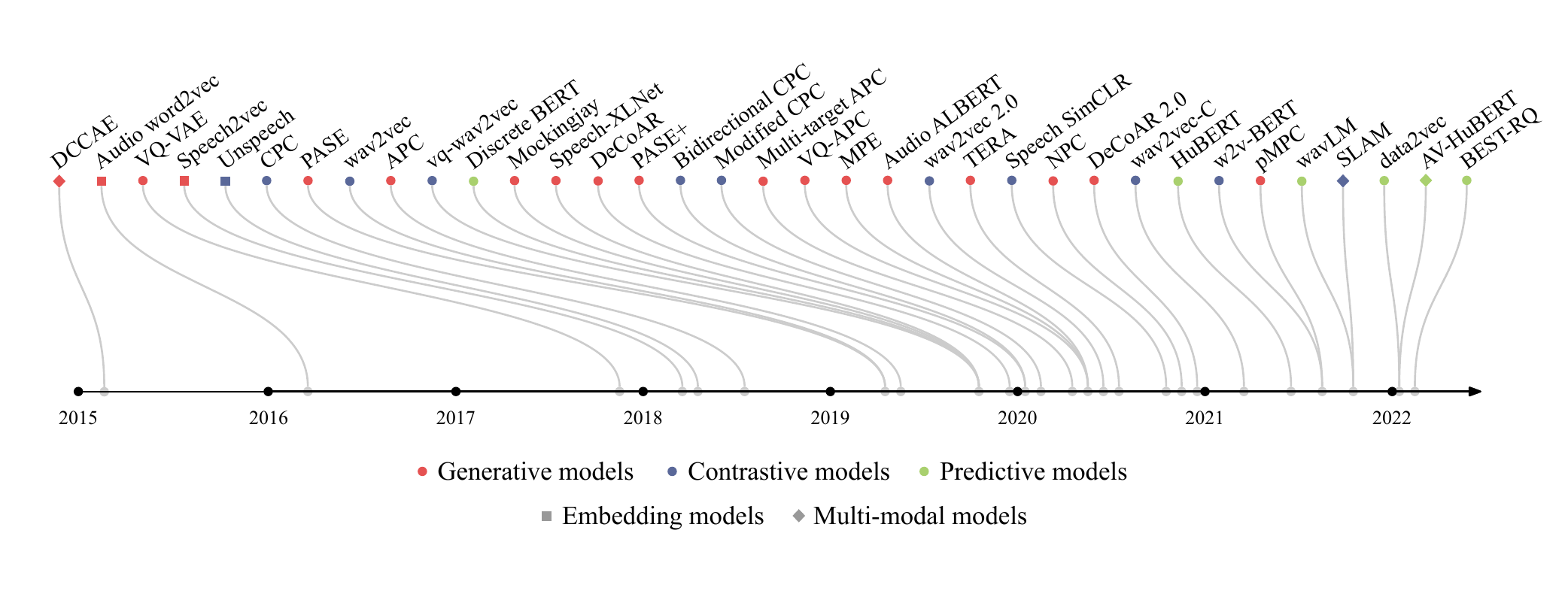}
	 \caption{A selection of models listed according to first publication date
	 on arXiv or conference submission date when this clearly precedes the
	 former. The models are categorized as generative, contrastive, or predictive.
	 In addition, some models are characterized as embedding models or
	 multi-modal models, although most learn frame-level
	 representations from speech only. Some models use a mixture of generative
	 and contrastive tasks. For instance, PASE and PASE+ use a multi-task setup,
	 but find that generative tasks are the most important for downstream
	 task performance~\cite{pascual2019learning}.}
    \label{fig:timeline}
\end{figure*}

In the sections below, we group modern SSL pretext tasks designed for speech
into three main categories: \textit{generative} approaches,
\textit{contrastive} approaches and \textit{predictive} approaches. 
%For each category, we provide a brief motivation and then describe prominent models that belong to it. Lastly, we discuss a set of challenges related to the category.
\Cref{fig:timeline} shows a timeline of the models covered in these sections
with each model colored according to our categorization. 
\Cref{table:pretext} summarizes model pretext tasks along within the categories.

%\input{tables/model}

% \begin{center}
% \begin{tabular}{ |c|c| } 
%  \hline
%  $s$ & a single waveform sample \\
%  $S$ & a sequence of waveform samples \\
%  $x$ & a single feature vector (MFCC or learned)\\
%  $X$ & a sequence of feature vectors \\
%  $\widetilde{X}$ & a data augmented sequence of feature vectors \\
%  $z$ & a single latent vector \\
%  $Z$ & a sequence of latent vectors \\
%  $h$ & a single hidden vector representation \\
%  $H$ & a sequence of hidden vector representation \\
%  $m$ & a binary mask for a single time step \\
%  $M$ & a binary mask for the whole sequence \\
%  $y$ & a single supervised label \\
%  $Y$ & a sequence of supervised labels \\
%  $\tau$ & temperature variable \\ 
%  $f()$ & an encoding network - use $f_1()$ $f_2()$ if multiple\\
%  $g()$ & a decoding network \\
%  $P()$ & Probability \\
%  $E()$ & Expectation \\
%  $L()$ & Loss \\
%  $\mathcal{L}$ & Likelihood \\
%  \hline
% \end{tabular}
% \end{center}

\subsection{Notation}
\label{sec:notation}

To efficiently describe the different approaches, we use a simple
notation. Models are assumed to consist of functions $f(\cdot)$ and $g(\cdot)$, where \edit{$f(\cdot)$  denotes the representation model to be used after pre-training and $g(\cdot)$ is an auxiliary module needed only to support the pretext task. For instance, in a classic autoencoder, $f(\cdot)$ would denote the encoder and $g(\cdot)$ the decoder. For more complex models, these functions might consist of several components indicated by sub-indices $f_1(\cdot) \dots f_N(\cdot)$. As we will see, many self-supervised models use masking, which replaces some parts of the input or a hidden representation  by zeros or a learned vector. We use $m(\cdot)$ to denote a function that applies such masking to its input. Similar to $g(\cdot)$, this function is only used during pre-training. 

Given an acoustic input $X =\{x_1,x_2, ..., x_T\}$, $f(\cdot)$ outputs a representation $H =\{h_1,h_2,...,h_T\}$. The input~$X$ may be either the raw waveform samples or a sequence of spectral feature vectors. Both are viable options in practice. For simplicity, we do not distinguish between the two in our notation.  

While $f(\cdot)$ always takes an acoustic input, the input to $g(\cdot)$ can be either the acoustic signal or another learned representation. Most importantly, $g(\cdot)$ produces an output that is used for the pretext task but is not  used by $f(\cdot)$ to produce the representation $H$. Hence, $g(\cdot)$ can be discarded after pre-training. Finally, $f(\cdot)$ commonly downsamples the temporal dimension, but again, this is not crucial to understand the models, so consider only a single temporal \ci{scale} $t\in\{1,\dots, T\}$ for notational convenience.

%A few other conventions should be noted here. 
We use $Q = \{q_1,q_2, ..., q_T\}$ to denote representations that are quantized via codebook learning. Alternatively, discrete representations may take the form of one-hot vectors, or the equivalent integer IDs, which we denote by $C = \{c_1,c_2, ..., c_T\}$. We use a circumflex to denote that, for instance, $\hat{x}_t$ is an approximation of $x_t$. Finally, we often use a subscript when defining a loss, $\mathcal{L}_i$, to imply that the total loss is computed as a sum over $i$, unless otherwise stated.

For some models, we will refer to $H$ as a \emph{contextualized} representation which means that each $h_t$ is a function of some, linguistically speaking, long sub-sequence of $X$ spanning at least several phonemes. Usually, $h_t$ depends on the entire input $X$ or all previous timesteps $X_{[1,t]}$. In contrast, a \emph{localized} representation is one that only depends on a short part of the input $X_{[t - u,t + u]}$, where $u \geq 0$. The distinction between contextualized and localized may become fuzzy if $u$ is large, however, this is rarely the case.}

\edit{After pre-training,} the representation model $f(\cdot)$ can be fine-tuned for a downstream task directly or used to extract features which are fed to another model, as visualized in \cref{fig:SSL_framework}. It is not uncommon to use the output representation~$H$, but often representations from hidden layers of $f(\cdot)$ are better suited \cite{pasad2021}.

\subsection{Generative approaches}
\label{sec:generative}

% {\color{blue} Hung-yi}\\

\subsubsection{Motivation}

In this category, the pretext task is to generate, or reconstruct, the input data based on some limited view. This includes predicting future inputs from past inputs, masked from unmasked, or the original from some other corrupted view.  ``Generative'' as used in this paper hence refers to models that target the original input in their pretext task. Note that this differs from generative models, which learn \ci{distributions that allow} to sample new data.
% It is important to note that many of the approaches discussed in this subsection are ''reconstructive``, i.e., they focus on reconstructing the input signal rather than learning a probabilistic model for sampling novel sequences\footnote{Generative self-supervised models are not the same as generative probabilistic models that permit sampling new data.}. 

%\footnote{The lengths of $X$ and $H$ are not necessary to be equal. Usually, $H$ is shorter than $X$. For simplicity of discussion, we assume the lengths of $X$ and $H$ are the same in this subsection.} % Any one care about this?

\subsubsection{Approaches}

%\subsubsection{Autoregressive reconstruction}
%\subsubsection{Autoregressive Prediction}
%\label{sssec:autoregressive}

\paragraph{Autoencoding} 
Since their introduction in the mid-1990s~\cite{hinton_94}, autoencoders (AEs)
have played an essential role in learning distributed latent representations of
sensory data. 
As described above, AEs consist of an encoder and decoder; the pretext task
is to reconstruct the given input. The most common type of AE places an
information bottleneck on the latent representation by simply having fewer
hidden units available than input features. This forces the model to discard
low-level details and disccourages the learning of trivial solutions. Other models add
regularization to the latent space to further improve the quality of the
learned representations.
% An AE converts an input utterance $X$ to a sequence of representations $H$ and then uses a decoder network to convert $H$ into an approximate reconstruction of the input $X$. Regularization is required to learn general and efficient latent representations and to avoid learning a trivial identity mapping. 
For instance, denoising autoencoders (DAEs) learn latent representations by
reconstructing from input corrupted by noise~\cite{DAE}. 
\edit{The Variational Autoencoder (VAE) is a probabilistic version of the AE which defines the latent representation via a posterior distribution over stochastic latent variables \cite{vae,rezende2014stochastic}. VAEs have been applied to speech in numerous works \cite{chung_recurrent_2015, fraccaro_sequential_2016, hsu2017learning, hsu2017unsupervised, aksan_stcn_2019}}.
The vector-quantized variational autoencoder (VQ-VAE) is another model in this
category~\cite{vqvae};
% The first method to sucessfully learn meaningful discrete latent representations was the Vector Quantised Variational Autoencoder (VQ-VAE)~\cite{vqvae}. 
it extends the original VAE~\cite{vae} with a novel
parameterization of the posterior distribution for discrete latent
representations. 
\edit{The VQ-VAE has been instrumental in generative speech modelling and recent work on generative spoken language modeling has successfully combined the idea of a discrete latent space with self-supervised learning \cite{SpeechResynthesis_IS21, pGSLM, GSLM}.}
% Forcing the use of discrete latent representations encourages the model to focus on salient input features, e.g., phonemes in speech, rather than using the capacity to model channel noise and imperceptible local input details.

Specifically, in the VQ-VAE, the continuous representation vector $h_t$ at the output of the encoder is quantized by mapping it to a codebook vector, which is then used as the input to the decoder. This operation is non-differentiable and the \edit{gradients of the loss with respect to the encoder parameters} must be obtained by approximation. In the VQ-VAE this is done using the straight-through estimator~\cite{bengio2013estimating}, i.e., the gradients \edit{with respect to the} encoder output are taken to be equal to those \edit{with respect to the} decoder input \ci{(i.e., the quatization step is ignored)}. Given a learned codebook $A\in\mathbb{R}^{K \times D}$, where $K$ is the codebook size and $D$ is the dimensionality of each codebook vector $a_k$, the quantized representation $q_t$ of $h_t$ is obtained as
\begin{align}
    q_t = a_k, \text{ where } k=\arg\min_j \norm{h_t - a_j}_2 .
\end{align}
\edit{The decoder $g(\cdot)$ is an autoregressive model that takes $q_{[1,t]}$ as input to generate $x_t$ \cite{oord2016wavenet}}. 
Codebook learning is facilitated by a two-term auxiliary loss similar to
classical vector quantization dictionary 
learning~\cite{burton_generalization_1983, soong_vector_1985}. 
Gradients for the codebook vectors are given solely by a term that moves
codebook vectors $a_k$ closer to the non-quantized vectors $h_t$. A so-called
\emph{commitment term} is added to ensure that non-quantized vectors do not grow
unboundedly by enforcing the encoder to keep them close to a codebook vector.
This commitment term is optimized only by the encoder. The
total VQ-VAE loss \edit{for a single timestep} is
% \begin{align}
%     %\mathcal{L} = \log p(x | q) +  \underset{\text{vq}}{\underbrace{\norm{\text{sg}\left[v\right] - A}_2^2}} + \underset{\text{commitment}}{\underbrace{\beta\norm{v - \text{sg}\left[A\right]}_2^2}} \enspace ,
%     \resizebox{.44\textwidth}{!}{%
%     $\mathcal{L}_t = \underset{\text{encoder+decoder}}{\underbrace{\log p(x_t | q_{[1,t]})}} +  \underset{\text{codebook}}{\underbrace{\norm{\mathrm{sg}\left[h_t\right] - A}_2^2}} + \beta\underset{\text{encoder}}{\underbrace{\norm{h_t - \mathrm{sg}\left[A\right]}_2^2}} \;,$
%     }
%     \label{eq: vector quantization losses}
% \end{align}
% \begin{align}
%     \mathcal{L}_t =&~
%     \underset{\text{encoder+decoder}}{\underbrace{\log p(x_t | q_{[1,t]})}} + \underset{\text{codebook}}{\underbrace{\text{MSE}\left(\mathrm{sg}\left[H\right], A\right)^2}} + \nonumber\\
%     & \underset{\text{encoder}}{\underbrace{\beta~\text{MSE}\left(H, \mathrm{sg}\left[A\right]\right)^2}} \;,
%     \label{eq: vector quantization losses}
% \end{align}
\edit{
\begin{align}
    \resizebox{.44\textwidth}{!}{%
    $\mathcal{L}_t = \underset{\text{encoder+decoder}}{\underbrace{\log p(x_t | q_{[1,t]})}} + \underset{\text{codebook}}{\underbrace{\text{MSE}\left(\mathrm{sg}\left[h_t\right], A\right)}} + \underset{\text{encoder}}{\underbrace{\alpha~\text{MSE}\left(h_t, \mathrm{sg}\left[A\right]\right)}} \;,$
    }
    \label{eq: vector quantization losses}
\end{align}}%
% \begin{align}
%     \mathcal{L}_t =&~
%     \underset{\text{encoder+decoder}}{\underbrace{\log p(x_t | q_{[1,t]})}} + \nonumber\\
%     & \underset{\text{codebook}}{\underbrace{\frac{1}{KD} \sum_{t=1}^{T}\sum_{k=1}^{K}\sum_{i=1}^{D} \left(\mathrm{sg}\left[h_{t,i}\right] - a_{k,i}\right)^2}} + \nonumber\\
%     & \underset{\text{encoder}}{\underbrace{\frac{\beta}{KD}\sum_{t=1}^{T}\sum_{k=1}^{K}{\sum_{i=1}^{D} \left(h_{t,i} - \mathrm{sg}\left[a_{k,i}\right]\right)^2}}} \;,
%     \label{eq: vector quantization losses}
% \end{align}
\noindent where \edit{$\log p(x_t|q_{[1,t]})$ is a reconstruction likelihood term usually using a categorical distribution,} $\mathrm{sg}[x] = x$ is the so-called stop-gradient operator \edit{which acts as the identity function during the forward pass but \ci{is assumed to have}
partial derivatives all equal to zero during the backward pass}, $\alpha$ is a scalar hyperparameter, 
% and we define $\text{MSE}(H,A) = \frac{1}{KD} \sum_{t=1}^{T}\sum_{k=1}^{K}\sum_{i=1}^{D}\left(h_{t,i} - a_{k,i}\right)^2$. 
\edit{and we define $\text{MSE}(h_t,A) = \frac{1}{KD} \sum_{k=1}^{K}\sum_{i=1}^{D}\left(h_{t,i} - a_{k,i}\right)^2$. 
The loss for a full sequence is the sum or mean over all $\mathcal{L}_t$.} 
% We note that the subtraction of a matrix from a vector in \cref{eq: vector
% quantization losses} implies the use of broadcasting.

These learned discrete representations have been shown to capture high-level
speech information closely related to phonemes, and are useful for
applications such as speaker conversion~\cite{chorowski2019unsupervised}.
Vector quantization is \edit{not} exclusive to VQ-VAE but has seen
widespread application within SSL for regularization purposes and to define
targets for the pretext task. We will cover these applications below.
%We highlight the models using vector-quantization in the column \textbc{{qtz}} in \cref{tab:model-taxonomy}.

\edit{The Gumbel softmax \cite{Gumbel-Softmax} is another frequently used approach for obtaining a discrete representation space, and has also been used for AEs \cite{eloff2019unsupervised}. In addition to the approaches discussed above, several other works on speech representation learning take inspiration from the AE framework \cite{zeiler2013rectified, badino2014auto, badino2015discovering, kamper2015unsupervised, renshaw2015comparison, settle2019_a2w}.}

\begin{comment}
\paragraph{Regularization}
In the generative-based approaches, some regularization in the encoder $f(\cdot)$ or decoder $g(\cdot)$ is required to encourage the model to utilize global speech information and prevent naive copying input when reconstruction. 
The models below have been enhanced by regularization methods. 
\begin{itemize}
%\item VAE, VQ-VAE
    \item DeCoAR 2.0~\cite{ling2020decoar} presents a deep contextualized acoustic representation learning approach with the addition of a vector quantization (VQ) layer.
    \item In VQ-APC~\cite{chung20e_interspeech}, a VQ layer is used with the APC objective, which imposes a bottleneck and forces the model to learn better representations.
    \item Two dropout regularization methods, attention dropout and layer dropout, are introduced to TERA~\cite{luo2021drop}. 
\end{itemize}
\end{comment}

\paragraph{Autoregressive prediction}
\label{par:apc}

Autoregressive predictive coding (APC)~\cite{chung2019unsupervised,
chung2020generative} \edit{takes inspiration from the classic Linear Predictive Coding (LPC) approach for speech feature extraction~\cite{LPC} and autoregressive language models (LM) for text, where the model learns to predict future information from past}.
% This approach can be viewed as training a speech version of an LM.
%Just like an LM for text, this approach uses an encoder model to encode temporal information of past acoustic sequence, and the decoder then predicts future frames.
%In this approach, given $X=\{x_1,x_2, ..., x_T\}$, $\widetilde{X}$ is the acoustic features up to time step $t$, that is, $X_{[0,t]} = \{x_1,x_2, ..., x_{t}\}$.
\edit{A function} $f(\cdot)$ reads the input sequence $X_{[1,t]}$ and \edit{outputs a representation sequence} $H_{[1,t]}$.
The \edit{auxiliary module} $g(\cdot)$ is a linear projection layer which takes the last vector of $H_{[1,t]}$ as input to \edit{approximate} $x_{t+c}$, where $c \geq 1$. Thus, $c$ indicates how many timesteps the model predicts ahead. The \edit{modules} $f(\cdot)$
and $g(\cdot)$ are jointly learned to minimize {the \lone{} loss between $x_{t+c}$ and its approximation $\hat{x}_{t+c}$}. APC is formulated as 
%\begin{align}
    %Z_{[0,t-1]} &= f(X_{[0,t-1]}) \\
    %\hat{\mathbf{x}}_{t+c} &= g(\mathbf{z}_{t-1}) \label{eq:c} \\
    %\mathcal{L}_t &= \lVert \hat{\mathbf{x}}_{t+c} - \mathbf{x}_{t+c} \rVert_1\enspace.
%\end{align}
\begin{align}
    H_{[1,t]} &= f(X_{[1,t]}) , \\
    \hat{x}_{t+c} &= g(h_{t}) \label{eq:c} , \\
    \mathcal{L}_t &= \lVert \hat{x}_{t+c} - x_{t+c} \rVert_1 .
\end{align}
In text-based autoregressive LMs, $c$ is set to $1$ \edit{to enable autoregressive generation}. However, due to the smoothness of the speech signal, neighboring acoustic features are usually similar. Depending on the downstream task, we are often interested in learning so-called \emph{slow features} that typically span multiple input frames~\cite{wiskott2002slow}. Even the smallest linguistic units of speech---phonemes---span $0.07$~seconds on average in the English TIMIT dataset~\cite{garofolo1993timit},   whereas spectrogram frames $\mathbf{x}_t$ are typically computed at $0.01$ second intervals. Thus, simply predicting the next frame constitutes a trivial pretext task for APC; the original work finds that $c=3$ performs well. 
%\begin{equation}
  %  L = \sum_{X} \sum_{t} d(x_{t+c}, g(h_t)),
%   L = d(x_{t+c}, g( f(X_{[0,t-1]}) )),
%\end{equation}
%where $d(x_{t+c}, g(h_t))$ is the distance between $x_{t+c}$ and  $g(h_t)$. 
In \cite{chung2020improved}, the APC objective is extended to multi-target
training. The new objective generates both past and future frames conditioned
on previous context. 
In VQ-APC~\cite{chung20e_interspeech}, quantization is used with the APC
objective, which imposes an information bottleneck serving as a regularizer.

A drawback of APC is that it encodes information only from previous timesteps
and not the entire input.
DeCoAR~\cite{ling2020deep} combines the bidirectionality of the popular NLP model
ELMo~\cite{peters2018deep} and the reconstruction objective of APC to alleviate
this issue and allow encoding information from the entire input. 
%, so it is able to learn deep contextualized acoustic representations. 
It uses a forward LSTM $f_1(\cdot)$ to encode $X_{[1,t]}$ and a backward LSTM
$f_2(\cdot)$ to encode $X_{[t+k,T]}$, where $k>1$: 
\begin{gather}
    H_{[1,t]} = f_1(X_{[1,t]}), \\
    H^\prime_{[t+k,T]} = f_2(X_{[t+k,T]}), \\
    \hat{X}_{[t+1,t+k-1]} = g(h_t, h^\prime_{t+k}).
\end{gather}
% The last vector in $H_{[1,t]}$, $h_{t}$, and the first vector in $H^\prime_{[t+k,T]}$, $h_{t+k}$, are the input to the decoder $g$ used to predict $X_{[t+1,t+k-1]}$. 
The input feature vector used in the downstream tasks is the concatenation of
$h_{t}$ and $h^\prime_{t}$.

\paragraph{Masked Reconstruction}
%\label{sssec:time-only-masked}
%In wav2vec 2.0~\cite{wav2vec2}, time masking is applied in the latent space. But we will not mention this in this section.

%Basic idea
Masked reconstruction is largely inspired by the masked language model (MLM) task from BERT~\cite{jacob2019BERT}. During BERT pre-training, some tokens in the input sentences are masked by randomly replacing them by a learned masking token or another input token. The model learns to reconstruct the masked tokens from the non-masked tokens. Recent work has explored similar pretext tasks \edit{for speech representation learning}. Similar to the DeCoAR model described above, this allows a model to learn contextualized representations that encode information from the entire input. While we here focus on the models that reconstruct the masked input, it is important to note that masking has also been used extensively for contrastive (\cref{contrastive_approaches}) and predictive (\cref{predictive_approaches}) models. 
%In \cref{tab:model-taxonomy}, the models that use masking are highlighted by the \textbc{{msk}} column.

From a high-level perspective, the training phase of models using masked
reconstruction can be formulated as
\begin{align}
    H &= f(m(X)), \\
    \hat{x}_t &= g(h_{t}), \\
    \mathcal{L}_t &= \lVert \hat{x}_{t} - x_{t} \rVert_1 .
\end{align}
\edit{The exact masking policy defined by $m(\cdot)$ differs from model to model and will be discussed further below.}
The function $f(\cdot)$ is typically a Transformer encoder~\cite{liu2020mockingjay,jiang2019improving,liu2020masked}, but recurrent neural networks have also been used~\cite{wang2020unsupervised}. \edit{In general, the Transformer encoder architecture has been adopted widely by self-supervised models for speech within all three surveyed categories.} \edit{The function $g(\cdot)$ is usually a linear projection or a multilayer perceptron (MLP)}. Finally, the loss $\mathcal{L}_t$ is commonly computed only for masked timesteps in order to discourage the model from learning an identity mapping.

%Time masking 
%In the time masking approaches~\cite{liu2020mockingjay,jiang2019improving,liu2020masked}, some of the input frames in $X$ are masked to zero or randomly replaced by other frames. $X$ with $t$-th to $t+k$-th frames masked is denoted as $X_{-[t,t+k]}$. The encoder $f(\cdot)$ is usually a Transformer that encodes the whole utterance. It takes $X_{-[t,t+k]}$ as input and generates $H$. The decoder $g(\cdot)$ is a prediction head, which use $H_{[t,t+k]}$ to predict $X_{[t,t+k]}$. The encoder $f(\cdot)$ and decoder $g(\cdot)$ are jointly learned to minimize the distance between the decoder output and $X_{[t,t+k]}$.
%\begin{equation}
%    L = d(X_{[t,t+k]}, g(f( X_{-[t,t+k]} )) ).
%\end{equation}

The masking policies used in NLP can be adapted to speech by considering a speech \edit{segment} equivalent to a token in a sentence; indeed, the masking strategy of BERT has also been used for speech pre-training~\cite{liu2020mockingjay}.
%However, taking inspiration from NLP is not sufficient, as speech data warrants domain-specific considerations.
In the standard BERT masking policy, each token is masked independently at random. However, for speech, masking a single \edit{sample or spectrogram frame results in a largely} trivial reconstruction task \edit{since, as discussed in paragraph \ref{par:apc}, the smoothness of audio signals may encourage the model to learn to simply interpolate neighboring frames.} Therefore it is common to mask chunks of consecutive frames~\cite{liu2020mockingjay,jiang2021further}. 

We can bring the pretext task closer to the NLP equivalent by using a masking policy where the masked regions of the input correspond to linguistic units. Instead of just masking a fixed number of consecutive frames, pMPC~\cite{yue2021pMPC} selects masked speech frames according to the phonetic segmentation in an utterance. \edit{However, in order to obtain this segmentation, some labeled data is of course needed.}

Whereas most studies use masking along the temporal dimension of the input, speech can also be masked along the frequency dimension when spectral input features are used~\cite{wang2020unsupervised,liu2021tera}. Frequency masking has been shown to improve representations used for speaker classification~\cite{liu2021tera}. 
%The phonetically motivated self-supervised representation learns the speech representation that benefits downstream speech processing tasks.

%TERA
%Previous work mostly explored masking on the temporal axis, and the model learns to reconstruct from corrupted blocks of time steps. Frequency is another alteration for masking. The model can learn to reconstruct from missing blocks of frequency bins~\cite{wang2020unsupervised,liu2021tera}. In Transformer Encoder Representations from Alteration (TERA), the time and frequency alterations are applied together in the pre-training process. This is simply SpecAugment~\cite{spec_augment}. 
%It has been found that each of the alteration methods guides the model to learn a distinct aspect of speech~\cite{liu2021tera}. The time alteration effectively enforces a more accurate phoneme prediction, keyword detection, and speech recognition, as it leads the model to learn richer phonetic content. The frequency alteration effectively improves speaker prediction accuracy, as it leads the model to encode speaker identity.
%The magnitude alteration effectively improves performance for all tasks, as it potentially increases data diversity for pre-training.
Some studies explore alternatives to masking the input directly. In non-autoregressive predictive coding (NPC)~\cite{liu21l_interspeech}, time masking is introduced through masked convolution blocks. Taking inspiration from XLNet~\cite{XLNet}, it has also been suggested that the input be reconstructed from a shuffled version~\cite{song20d_interspeech} to address the discrepancy between pre-training and fine-tuning of masking-based approaches.
%In NLP, the masking tokens only exist in the pre-training phase and never appear in the downstream tasks, making the pre-training and downstream phases have mismatched inputs. To avoid using masks during pre-training, in NLP, XLNet is proposed, which learns by reconstructing from shuffled input. For speech, the speech version of XLNet, Speech-XLNet~\cite{song20d_interspeech}, has also been proposed. 

Regularization methods can further improve on masked reconstruction approaches. \edit{DeCoAR 2.0~\cite{ling2020decoar} uses vector quantization}, which is shown to improve the learned representations. Furthermore, two dropout regularization methods---attention dropout and layer dropout---are introduced with the TERA model~\cite{liu2021tera, luo2021drop}. \edit{Both methods are variations on the original dropout method \cite{srivastava_dropout_2014}.}

% PASE, WaveNet autoencoders, Phase reconstruction, Audio2Vec
%\label{sssec:autoencoder}
%\subsubsection{Autoencoder} 
%TBD: 
% -- Do we have to mention ConvDMM? %The ConvDMM~\cite{khurana20_interspeech} approach learns speech representations with convolutional neural networks and Markov Models.
% -- More about auto-encoder

\paragraph{More Generative Approaches}
Other than the autoregressive and masked reconstruction tasks discussed above, various studies have explored the reconstruction of other targets derived from the input. PASE and PASE+~\cite{pascual2019learning,ravanelli2020multi} use multiple targets, including the waveform, log power spectrum, \edit{mel cepstral coefficients (MFCCs)}, and prosody features. Models that learn acoustic embeddings of small speech segments have targeted future and past spectrogram segments~\cite{chung2018speech2vec, tagliasacchi2019self, tagliasacchi2020pre}, phase information~\cite{quitry2019learning}, and the temporal gap between two segments~\cite{tagliasacchi2019self, tagliasacchi2020pre}.

%\begin{itemize}
% \item Autoencoder~\cite{chorowski2019unsupervised}: The input and output are the same. Usually the regularization introduced above is required. %In these works, the autoencoder framework is designed to encode only phonetic content in latent representation and remove other confounding detail such as speaker identity.
%\item The model learns through reconstructing a spectrogram slice from past and future slices~\cite{tagliasacchi2019self, tagliasacchi2020pre}. %Similar to DecoAR?
%\item The TemporalGap~\cite{tagliasacchi2019self, tagliasacchi2020pre} approach learns through estimating the temporal gap between two short audio segments extracted at random from the same audio clip. %Can this approach be considered as generative?
%\item Representations are learned through reconstructing the phase of the short-time Fourier transform from its magnitude~\cite{quitry2019learning}. %This is audio, not speech.
%\item In PASE~\cite{pascual2019learning,ravanelli2020multi}, a single neural encoder learns to solve multiple self-supervised tasks at once, including reconstruction of waveform, Log power spectrum, MFCC, prosody, and other binary discrimination tasks. %Is this PASE+? I cite PASE+ here, are all the tasks of PASE+ included here. Is this approach generative? 
%\end{itemize}

\subsubsection{Challenges} 

Although successful NLP models like BERT and GPT are based on generative pretext tasks, the progress have not been translated directly to the speech domain. A speech signal encodes more information than text, such as speaker identity and prosodic features, which makes it harder to generate. However, in order to generate all details of the input, the model must encode all information in the speech signal. Hence, a model that learns to perfectly reconstruct its input may not necessarily have learned to isolate the features of interest \edit{\ci{and will encode} redundant information for a given downstream task.} 

There are many choices involved in designing a generative pretext task. For instance, \edit{masking strategy and the choice of input and target representation (e.g., waveform samples or spectral
features)}. These choices influence what the model learns through the pretext task. However, there is little research on the relationship between task design and the information encoded in the learned representations.

%In \cite{slu_bert}, phoneme posterior vectors are used to train a standard BERT~\cite{bert, xlnet} model.
%The phoneme posterior vectors are output from a supervised acoustic model, which requires CTC loss training over the ground-truth phonemes.

% This is about network compression. Perhaps we will have a session about network compression.
%In Audio ALBERT~\cite{audioalbert}, Mockingjay is modified to have shared parameters across Transformer layers.

%Perhpas we will talk about attack defense
%In \cite{mockingjay_defense}, Mockingjay is shown to be effective in defending adversarial black-box attacks.

\begin{table*}[!htb]
    \centering
    \caption{
	 A summary of the approaches in the three categories of
	 self-supervised learning. 
	 Column~(a) lists the names of the models and related references, 
	 column~(b) defines the model input, 
    column~(c) defines any corruption of the input or hidden representation, and
	 column~(d) defines the target of the pretext task; the pretext task itself
	 is described by the overall model category and the main text.
	 $X=\{x_1,x_2,...,x_T\}$ is the input sequence in which $x_t$ can be an
	 acoustic feature vector (e.g., MFCC, filterbank, or spectrogram features)
	 or a waveform sample. 
    $X_{[t_1:t_2]}$ represents $\{x_{t_1},x_{t_1+1},...,x_{t_2}\}$.
	 $X_{-[t_1:t_2]}$ represents $X$ in which the segment
	 $X_{[t_1:t_2]}=\{x_{t_1},x_{t_1+1},...,x_{t_2}\}$ is masked.
	 $x_{t}^{i}$ represents the $i$-th dimension of $x_t$.
	 If $x_t$ is a frame in a spectrogram, then the $i$-th dimension corresponds
	 to a specific frequency bin.
	 $X^{-[f,f+j]}$ refers to a spectrogram $X$ which is masked along the frequency
	 axis from the $f$-th to $(f+j)$-th bin. 
    % $X_{-[t:t+k]}^{-[f,f+j]}$ refers to a spectrogram $X$ but masked along both time and frequency dimensions. 
    % $X^*$ is a temporally permuted version of $X$, that is, the $x_t$ are randomly shuffled to form $X^*$. 
	 We indicate random temporal permutation of a sequence by indexing it with
	 the set $\mathcal{P}_t\triangleq\textsc{permute}([0,t])$, where
	 $\textsc{permute}(\cdot)$ returns a permutation of the given list. 
	 We indicate data augmentation (e.g., reverberation) by the function
	 $\textsc{augment}(\cdot)$. Subscripts indicate different augmentations. 
    % $X^\prime$ is an augmented version of $X$ (e.g., $X$ with reverberation), while $X^{\prime\prime}$ is $X$ adding distortion different from  $X^\prime$.
    $Z$ represents a localized latent representation sequence of $X$. %, and $Z^\prime$ and $Z^{\prime\prime}$ are the latent representation sequences of  $X^\prime$ and $X^{\prime\prime}$, respectively. 
    % $Z$ represents a localized latent representation sequence and $Z^\prime$ is the latent representation sequence of $X^\prime$. 
    $Z^{(l)}$ is $Z$ at the $l$-th layer of the model used to compute it.
	 $\bar{H}$ is the contextualized sequence $H$ obtained from an exponential
	 moving average (EMA) of the model undergoing training with no masking
	 applied.
	 $Q$ represents a sequence of quantized learned representations, and $C$ is
	 a sequence of discrete cluster IDs.
    For contrastive models, we specify only positive targets.
    }
%    \begin{minipage}{\textwidth}  
    \centering
    \renewcommand*\arraystretch{1.2}{
    \begin{tabular}{l|c|c|c}
    \toprule
    \textbf{Model} (a) & \textbf{Input} (b) & \textbf{Corruption} (c) & \textbf{Target} (d) \\
    \midrule
    \midrule
    \multicolumn{4}{c}{\textsc{Generative models}} \\
    \midrule
    \midrule
    Audio Word2vec~\cite{Chung2016AudioWord2Vec}, VQ-VAE~\cite{vqvae}     & $X$ &   \textsc{-}    &  $X$  \\ %It has VQ as extra constraint.
    \midrule  
    Speech2Vec~\cite{chung2018speech2vec}, Audio2Vec~\cite{tagliasacchi2020pre} - skip-gram    & $X_{[t_1,t_2]}$  &     \textsc{-}   &    $X_{[t_0,t_1]}$,$X_{[t_2,t_3]}$     \\
    \midrule 
   Speech2Vec~\cite{chung2018speech2vec}, Audio2Vec~\cite{tagliasacchi2020pre} - cbow    & $X_{[t_0,t_1]}$,$X_{[t_2,t_3]}$   &     \textsc{-}   &    $X_{[t_1,t_2]}$   \\
    \midrule 
     PASE~\cite{pascual2019learning}, PASE+~\cite{ravanelli2020multi}\footnote{PASE uses multiple pretext tasks, but the authors find that reconstruction is most important.}       & $X$ &  \textsc{-}   &  Different modalities of $X$  \\
    \midrule 
    APC~\cite{chung2019unsupervised,chung20e_interspeech}         & $X_{[1,t]}$   & \textsc{-}              & $x_{t+c},\, c\geq1$    \\ %VQ-APC is the same as APC
    \midrule
    Speech-XLNet \cite{song20d_interspeech}     & \multicolumn{2}{c|}{$X_{\mathcal{P}_{t}}$}   &     $x_{i\sim\mathcal{P}^c_{t}}$  \\ %Lee: Use * to represent the permutation, hope it is not strange.
    \midrule  
    DeCoAR~\cite{ling2020deep}     & $X_{[1,t-1]}, X_{[t+k+1,T]}$ & \textsc{-} & $X_{[t,t+k]}$   \\
    \midrule
    Mockingjay~\cite{liu2020mockingjay}, Audio ALBERT~\cite{chi2020audio}, DeCoAR 2.0~\cite{ling2020decoar}   & \multicolumn{2}{c|}{$X_{-[t,t+k]}$}   & $X_{[t,t+k]}$    \\
    \midrule 
    TERA~\cite{liu2021tera}, BMR~\cite{wang2020unsupervised}  & \multicolumn{2}{c|}{$X_{-[t,t+k]}^{-[f,f+j]}$}       & $X$       \\
    \midrule
    pMPC~\cite{yue2021pMPC}      &  \multicolumn{2}{c|}{$X_{-[t,t+k^\prime]}$ ($X_{[t,t+k^\prime]}$ is a phoneme)}       & $X_{[t,t+k^\prime]}$    \\
    \midrule 
    MPE~\cite{liu2020masked} & $X$ &  $Z_{-[t,t+k]}$  & $Z$    \\ %Lee: What is the difference between MPE and NPC??? And I believe it reconstruct the convolutional blocks output (learned target?)
    \midrule
    NPC~\cite{liu21l_interspeech}      & $X$  &   $Z_{-[t,t+k]}$  &    $X$   \\ %Lee: I am not 100% sure it is correct. Please check.
    \midrule
    \midrule
    \multicolumn{4}{c}{\textsc{Contrastive models}} \\
    \midrule
    \midrule
    Unspeech \cite{milde2018unspeech}       &   $X_{[t_1,t_2]}$ &   \textsc{-}   &  $X_{[t_0,t_1]}$,$X_{[t_2,t_3]}$ \\
    \midrule 
    CPC~\cite{oord2018representation}, wav2vec \cite{schneider2019wav2vec}, Modified CPC \cite{riviere2020unsupervised}         & $X_{[1,t]}$   &    \textsc{-}           & $z_{t+c},\, c\geq1$   \\ %Modified CPC is the same as CPC
        \midrule 
    Bidirectional CPC \cite{kawakami2020learning}      & $X_{[1,t]}$ or $X_{[t,T]}$ &  \textsc{-}    &    $z_{t+c}$ or $z_{t-c},\, c\geq1$   \\
    \midrule 
    vq-wav2vec \cite{Baevski2020vq-wav2vec}     &   $X_{[1,t]}$ &   \textsc{-}    &   $q_{t+c},\, c\geq1$   \\ 
    \midrule 
    wav2vec 2.0 \cite{baevski2020wav2vec}, wav2vec-C \cite{sadhu21_interspeech}\footnote{wav2vec-C adds reconstruction loss to wav2vec 2.0.}    & $X$             & $Z_{-[t,t+k]}$          & $Q_{[t,t+k]}$ \\
    \midrule 
    w2v-BERT \cite{w2vbert}     &$X$ &    $Z_{-[t,t+k]}$   &     $Q_{[t,t+k]}$ and $C_{[t,t+k]}$     \\
    \midrule
    Speech SimCLR \cite{SpeechSimCLR}\footnote{Speech SimCLR targets the latent representation of an augmented version of $X$ using a differently augmented $X$, and vice-versa.}    & \multicolumn{2}{c|}{$\textsc{augment}_1(X)$ and $\textsc{augment}_2(X)$}     &    $\textsc{augment}_2(Z)$ and $\textsc{augment}_1(Z)$   \\ 
   \midrule
   \midrule 
    \multicolumn{4}{c}{\textsc{Predictive models}} \\
    \midrule
    \midrule
    Discrete BERT~\cite{Baevski2020vq-wav2vec,baevski2019effectiveness} \footnote{Discrete BERT obtains codes $C$ from vq-wav2vec.}      &   \multicolumn{2}{c|}{$C_{-[t,t+k]}$}   & $C_{[t,t+k]}$  \\
    \midrule 
    HuBERT \cite{hsu2021hubert}\footnote{HuBERT is trained first using cluster IDs of the MFCCs as target and subsequently clusters IDs of the model representations from the last iteration.}, WavLM \cite{chen2021wavlm}\footnote{WavLM simulates noisy/overlapped speech as inputs.}  & $X$             & $Z_{-[t,t+k]}$          & $C_{[t,t+k]}$  \\ 
    \midrule
    data2vec \cite{data2vec}    & $X$             & $Z_{-[t,t+k]}$          & $\sum_{l}\bar{H}^{(l)}_{[t,t+k]}$  \\ 
        \midrule 
BEST-RQ \cite{BEST-RQ}\footnote{BEST-RQ obtains codes $C$ by quantizing acoustic features using a random projection quantizer.}     &  \multicolumn{2}{c|}{$X_{-[t,t+k]}$}      &  $C_{[t,t+k]}$   \\ 
        \midrule 
    \bottomrule
    \end{tabular}
    }
    
 %   \end{minipage}
    \label{table:pretext}
\end{table*}

\subsection{Contrastive approaches}
\label{contrastive_approaches}
\subsubsection{Motivation}
As discussed above, speech contains many entangled features. Thus, \edit{learning to reconstruct the raw speech signal} might not be the best way to discover contextualized latent factors of variations. 
Contrastive models learn representations by distinguishing a target sample (positive) from distractor samples (negatives) given an \emph{anchor representation}. \edit{The pretext task is to maximize latent space similarity between the anchor and positive samples while minimizing the similarity between the anchor and negative samples}. This approach has been used extensively in the general \edit{ML} community \cite{Schultz_Joachims_NeurIPS2004}.

\subsubsection{Approaches}
\paragraph{CPC}
\label{par:cpc}
Contrastive Predictive Coding (CPC)~\cite{oord2018representation} is a prominent example of a contrastive model. 
\edit{CPC uses a convolutional module $f_1(\cdot)$ to produce localized representations $z_t$ with a recurrent module $f_2(\cdot)$ on top that outputs a contextualized representation $h_t$. An anchor representation $\hat{z}_{t,k}$ is obtained via a linear projection $g_k(\cdot)$ of $h_t$. The positives and negatives are sampled from the localized representation $Z$.
%\edit{CPC forms localized representations $z_t$ that depend only on a local window of $X$ and then creates a contextualized representation $h_t$ by running an autoregressive module $f_2(\cdot)$ over $Z_{[1,t]}$. The positives and negatives are then sampled from the localized representation $Z$.
%while the contextualized representation $h_t$ is transformed to an anchor representation $\hat{z}_{t,k}$.
Hence, at a single timestep $t$, CPC forms multiple anchor representations $\hat{z}_{t,k}$ for $k\in\{1,\dots,K\}$ and associates with each one a single positive sample at the corresponding timestep, $z_{t+k}$, $k$ steps in the future:
\begin{align}
    z_t &= f_{1}(X_{[t-u,t+u]}) \label{eq: cpc local representation} \enspace ,\\
    H_{[1,t]} &= f_{2}(Z_{[1,t]}) \enspace , \\
    \hat{z}_{t,k} &= g_k(h_{t}) \enspace.
\end{align}
\noindent Each $z_{t}$ only encodes information from a limited receptive field, while $f_2(\cdot)$ is limited to condition each $h_t$ on previous timesteps $Z_{[1,t]}$. \ci{Without these restrictions}, the model could collapse to a trivial solution. $g_k$ is a unique transformation per \ci{offset} $k$ (e.g., a linear projection).
%where $f_1(\cdot)$ is a convolutional neural network, such that each $z_{t}$ only encodes information from a limited receptive field in $X$, $f_2(\cdot)$ is limited to condition each $h_t$ only on previous timesteps $Z_{[1,t]}$ and $g_k$ is a unique transformation per step $k$ (e.g. a linear projection). 
The loss function measures the similarity between the anchor representation $\hat{z}_{t,k}$ and the positive $z_{t+k}$ normalized by the total similarity to the positive and negatives.}
The approach is similar to previous work on Noise-Contrastive Estimation (NCE)~\cite{gutmann2010noise}. \edit{Minimizing} the loss corresponds to maximizing a lower bound on the mutual information between $h_t$ and $z_{t+k}$ (and in turn $x_{t+k-u:t+k+u}$) and is hence called InfoNCE:
\begin{align}
    \mathcal{L}_{t,k} &= - \log \left(\frac{\exp(\hat{z}_{t,k}^{\text{\tiny T}}z_{t+k})}{\sum_{\ci{i \in \mathcal{I}}} \exp(\hat{z}_{t,k}^{\text{\tiny T}}z_{i})} \right)\enspace .
    \label{eq: cpc loss}
\end{align}
\noindent \edit{Here, $\mathcal{I}$ \ci{is a random subset of $N$} indices  which includes the target index $t+k$ and $N-1$ negative samples drawn from a proposal distribution, e.g., a uniform distribution over $\{1,\dots,T\}$. Including the target index in $\mathcal{I}$ ensures that the loss is a proper categorical cross-entropy and that minimizing it has the previously stated relation to mutual information maximization}. This corresponds to sampling negatives from the same sequence and has been \edit{shown} to give good performance for phoneme classification~\cite{oord2018representation}. The loss is indexed by $k$ to show that CPC targets multiple offsets using different projection layers $g_k(\cdot)$. The authors find \edit{$K=12$} to work well for phoneme classification.

The wav2vec model \cite{schneider2019wav2vec} extends the CPC approach \edit{and uses fully convolutional parameterizations for the modules $f_1(\cdot)$ and $f_2(\cdot)$ with receptive fields of \SI{30}{ms} and \SI{210}{ms}, respectively. While the CPC loss solves a 1-of-$N$ classification task per $(t,k)$, either assigning the anchor to the positive class or (wrongly) to one of the $N-1$ negative classes, the wav2vec loss considers a sequence of $N$ independent binary classifications. That is, the anchor is compared independently to the positive and each negative, and the loss is computed as a sum of the associated log-probabilities,
\begin{align}
    %g_k(\mathbf{c}_{t}, \mathbf{v}_{t+k}) &= \log(\sigma(\mathbf{v}_{t+k}^{\intercal} (\mathbf{W}_k \mathbf{c}_{t} + \mathbf{b}_k)))\\
    \mathcal{L}_{t,k} &= - \log(\sigma(\hat{z}_{t,k}^{\text{\tiny T}}z_{t+k})) + \sum_{i \in \mathcal{I}} \log(1 - \sigma(\hat{z}_{t,k}^{\text{\tiny T}}z_{i})) \enspace .
    \label{eq: wav2vec loss}
\end{align}
\noindent Here, $\sigma(x)=1/(1+\exp(-x))$ is the sigmoid function, $\sigma(\hat{z}_{t,k}^{\text{\tiny T}}z_{t+k})$ is the probability of the anchor being the positive sample and $\sigma(\hat{z}_{t,k}^{\text{\tiny T}}z_{i})$ is the probability of the anchor being the negative sample. Evidently and contrary to CPC, $\mathcal{I}$ must not include the target index $t+k$ as this would cancel out the positive term.}

\paragraph{wav2vec 2.0} 
\edit{The wav2vec 2.0 model combines contrastive learning with masking}. As the CPC model, it uses the InfoNCE loss \cite{oord2018representation} to maximize the similarity between a contextualized representation and a localized representation. \edit{However, instead of using the $z_t$ directly as positive and negatives, it uses a quantization module $g(\cdot)$ to obtain a discrete representation. This has the practical implication that one can avoid sampling negatives from the same category as the positive}. \edit{The model} takes as input a waveform and uses a convolutional module \edit{$f_1(\cdot)$} followed by a \edit{Transformer encoder} \edit{$f_2(\cdot)$}. Masking is applied to the output of the convolutional module: 
\begin{align}
    z_t &= f_1(X_{[t-u,t+u]}) \label{w2v2 f_v} \enspace , \\
    H &= f_2(m(Z)) \label{w2v2 f_c} \enspace , \\
    q_t &= g(z_t) \enspace . \label{w2v2 qtz}
\end{align}
\noindent
%Here, $m(\cdot)$ again defines a masking policy 
% Here the $\circ$ operator denotes the Hadamard product with the mask $M$,
% , $f_1(\cdot)$ is a convolutional neural network with limited receptive field, $f_2(\cdot)$ is a Transformer encoder, 
The quantization module $g(\cdot)$ uses a Gumbel softmax~\cite{Gumbel-Softmax} with a straight-through estimator. Since the quality of the learned representations is contingent on the quality of the quantization, wav2vec 2.0 combines two techniques to learn high-quality codebooks. First, wav2vec 2.0 concatenates quantized representations from multiple codebooks at each timestep, so-called Product Quantization (PQ)~\cite{ProductQuantization}. Also, the \edit{primary} training loss \edit{described below} is augmented with an auxiliary term designed to encourage equal use of all codebook entries. 

In wav2vec 2.0, anchors are taken to be $h_t$ at masked timesteps only, the positive sample is chosen as the quantized vector, $q_t$, at the same timestep, and negatives are sampled from other masked timesteps. The loss is
% For the InfoNCE loss, anchors features are taken from the top layer contextualized Transformer representations for masked regions,  positive samples are the corresponding quantized features, and negative samples are selected from other masked locations in the same utterance. 
%
\begin{align}
    \mathcal{L}_t &= - \log \left(\frac{\exp(S_{\text{c}}(h_{t}, q_{t}))}{\sum_{i \in \mathcal{I}} \exp(S_{\text{c}}(h_{t}, q_{i}))} \right) \enspace , \label{w2v2 loss}
\end{align}
\noindent where $S_{\text{c}}(\cdot)$ is the cosine similarity and $\mathcal{I}$ contains the target index $t$ and negative indices sampled from other masked timesteps.
%\edit{Notably, since the positives and negatives are quantized representations, we can guarentee that representations of the same class as the positive are excluded when sampling the negatives.}

The wav2vec 2.0 approach was the first to reach single-digit \edit{word error rate (WER) on LibriSpeech using only the low-resource Libri-light subsets for fine-tuning a pre-trained model (see \cref{sec:datasets}). It has subsequently} inspired many follow-up studies. The wav2vec-C~\cite{sadhu21_interspeech} approach extends wav2vec 2.0 with a consistency term in the loss that aims to reconstruct the input features from the learned quantized representations, similar to VQ-VAE~\cite{VQVAE2}.
%The wav2vec-C representations achieve, on average, twice the error reduction over baseline and a higher codebook utilization in comparison to wav2vec 2.0 for far-field English voice commands and voice queries. 

\subsubsection{Challenges}
Although representations learned using contrastive approaches have proved effective across a wide range of downstream applications, they face many challenges when applied to speech data. 
One challenging aspect is that the strategy used to define positive and negative samples can also indirectly impose invariances on the learned representations. For example, sampling negatives exclusively from the same utterance as the positive biases the features towards speaker invariance, which may or may not be desired for downstream applications. 
Another standing challenge is that since speech input does not have explicit segmentation of acoustic units, the negative and positive samples do not represent a whole unit of language but rather partial or multiple units, depending on the span covered by each sample. 
Finally, since speech input is smooth and lacks natural segmentation, it can be difficult to define a contrastive sampling strategy that is guaranteed to provide samples that always relate to the anchor as truly positives and negatives in a sound way.

\subsection{Predictive approaches}
\label{predictive_approaches}
\subsubsection{Motivation}
\edit{Similar to the contrastive approaches discussed above, predictive approaches are defined by using a learned target for the pretext task. However, unlike the contrastive approaches, they do not employ a contrastive loss and instead use a loss function such as squared error and cross-entropy. 
Whereas a contrastive loss discourages the model from learning a trivial solution by the use of negative samples, this must be circumvented differently for predictive methods.
%However, such losses do not discourage trivial solutions, in the same way that contrastive losses do.
For this reason, predictive methods compute the targets outside the model's computational graph; usually with a completely separate model. Thus, the predictive setup is somewhat akin to teacher-student training. The first predictive approaches were motivated by the success of BERT-like methods in NLP \cite{jacob2019BERT} as well as the DeepCluster method in CV \cite{Caron_2018_ECCV}.}
%where the model learns a contextual representation that aggregates information from distant timesteps. Predictive approaches for speech representation learning also draw inspiration from the DeepCluster method \cite{Caron_2018_ECCV} for self-supervised visual learning.

\subsubsection{Approaches}

\paragraph{Discrete BERT} \edit{Applying BERT-type training directly to speech input is not possible due to its continuous nature}. The Discrete BERT approach~\cite{baevski2019effectiveness} uses a pre-trained vq-wav2vec model to derive a discrete vocabulary \cite{Baevski2020vq-wav2vec}. The vq-wav2vec model is similar to wav2vec mentioned in Paragraph~\ref{par:cpc} but uses quantization to learn discrete representations. 
%For the LibriSpeech dataset, the discrete vocabulary contains about 13.5k unique codes. 
\edit{Specifically, discrete units $c_t$ are first extracted with the vq-wav2vec model $f_1(\cdot)$ and then used as inputs \emph{and} targets in a standard BERT model $f_2(\cdot)$ with a softmax normalized output layer $g(\cdot)$, 
\begin{align}
    c_t &= f_1(X_{[t-u,t+u]}) \enspace , \\
    H &= f_2(m(C)) \enspace , \\
    \hat{c}_t &= g(h_t) \enspace .
\end{align}
Similar to BERT, the model can then be trained with a categorical cross-entropy loss,
\begin{align}
    \mathcal{L} &= \sum_{t\in \mathcal{M}} - \log p(c_t \mid X) \enspace ,
\end{align}
\noindent where $\mathcal{M}$ is the set of all masked timesteps. During training, only the BERT model's parameters are updated, while the vq-wav2vec model parameters are frozen.} 
% After pre-training, the BERT Transformer model is fine-tuned using a Connectionist Temporal Classification (CTC) \cite{graves2006connectionist} loss for ASR. 
%Following the Libri-light near-zero evaluation policy \cite{kahn2020libri}, fine-tuning is done using 10 minute, 1 hour and 10 hour subsets of LibriSpeech balanced for gender and recording conditions, as well as the standard 100 hour-clean part of LibriSpeech.
Discrete BERT was the first model to demonstrate the effectiveness of self-supervised speech representation learning by achieving a WER of 25\% on the standard test-other subset using a 10-minute fine-tuning set, \edit{setting the direction for many approaches to follow}.

\paragraph{HuBERT} Rather than relying on an advanced representation learning model for discretizing continuous inputs, as Discrete BERT, the Hidden Unit BERT (HuBERT) approach \cite{hsu2021hubert} \edit{uses quantized MFCC features as targets learned with classic \kmeans{}. Thus, to compute the targets, the \kmeans{} model $g_1(\cdot)$ assigns a cluster center to each timestep.  
% \begin{align}
%     c_t &= g_1(X_{[t-w,t+w]}) \enspace .
% \end{align}
% \noindent where $w$ defines the window size used to compute the MFCCs. 
Different from Discrete BERT, HuBERT takes the raw waveform as input, rather than discrete units. This helps to prevent loss of any relevant information due to input quantization.
%Specifically, HuBERT predicts the \edit{pre-computed} \kmeans{} cluster assignment given the masked continuous speech features.
HuBERT uses an architecture similar to that of wav2vec 2.0, with a convolutional module $f_1(\cdot)$ and a Transformer encoder $f_2(\cdot)$, as well as a softmax normalized output layer $g_2(\cdot)$:
%Similar to wav2vec 2.0, the HuBERT model uses a convolutional encoder for the input. Masking is then applied before the Transformer network as in wav2vec 2.0 \namecrefs{w2v2 f_v} \ref{w2v2 f_v} and \ref{w2v2 f_c}.
\begin{align}
    c_t &= g_1(X_{[t-w,t+w]}) \enspace , \\
    z_t &= f_1(X_{[t-u,t+u]}) \enspace , \\
    H &= f_2(m(Z)) \enspace , \\
    \hat{c}_t &= g_2(h_t) \enspace ,
\end{align}
\noindent where $w$ defines the window size used to compute the MFCCs. 
%Since the targets are pre-computed cluster identities, the HuBERT model can directly evaluate the regular categorical cross-entropy loss between the correct \kmeans{} cluster and the predicted one. This is opposed to constrastive methods which require negative samples to avoid degenerating to trivial solutions.
The categorical cross-entropy loss is} computed on both masked, $\mathcal{L}_m$, and unmasked, $\mathcal{L}_u$, timesteps:  
\begin{align}
 %   \mathcal{L}_m &= \sum_{t\in M} - \log p(c_t | \tilde{X}, t)
%    \mathcal{L}_m &= \sum_{t\in M} - \log p(c_t \mid H, t) \enspace ,
    % \mathcal{L}_m &= \alpha\sum_{t\in M} - \log p(c_t \mid H) + (1-\alpha)\sum_{t\notin M} - \log p(c_t \mid H) \enspace , \\
    \mathcal{L}_m &= \sum_{t\in \mathcal{M}} - \log p(c_t \mid X) \enspace , \\
    \mathcal{L} &= \beta
    \mathcal{L}_m + (1 - \beta) \mathcal{L}_u \enspace .
\end{align}
\noindent Again, $\mathcal{M}$ is the set of all masked timesteps, $\beta$ is a scalar hyperparameter and $\mathcal{L}_u$ is computed as $\mathcal{L}_m$ but summing over $t\notin \mathcal{M}$.

Intuitively, the HuBERT model is forced to learn both \edit{an acoustic and a language model. 
First, the model needs to learn a meaningful continuous latent representation for unmasked timesteps which are mapped to discrete units, similar to a classical frame-based acoustic modeling problem. Second, similar to other masked pre-training approaches, the model needs to capture long-range temporal dependencies to make correct predictions for masked timesteps}.

One crucial insight motivating this work is the importance of consistency of the targets which enables the model to focus on modeling the sequential structure of the input. \edit{Importantly though, for HuBERT, pre-training is a two-step procedure. The first iteration is described above.
%As described, in the first iteration of pre-training, \kmeans{} clustered MFCC features are used as targets, $c_t$.
Once completed, a second iteration of pre-training follows. Here, representations from a hidden layer of the model from the first iteration are clustered with \kmeans{} to obtain new targets $c_t$.}
%Here, the final model of the first iteration is used to encode latent representations of the full dataset. These are then quantized using \kmeans{} to form a new set of target vectors, replacing $c_t$ from the first iteration.
%Alternating between learning offline discrete targets using \kmeans{} and training the model using masked prediction ensures training stability while preserving representation quality.

\edit{For HuBERT, only two iterations are needed to match or outperform the previous state-of-the-art results for low-resource speech recognition. And combining the HuBERT approach with the wav2vec 2.0 approach, the w2v-BERT model has managed to improve results even further~\cite{w2vbert}.}

%\edit{The w2v-BERT approach~\cite{w2vbert} combines contrastive and predictive approach and optimizes the wav2vec 2.0 contrastive loss and the HuBERT masked prediction loss jointly}. It showed 5\% to 10\% relative WER reduction on the test-clean and test-other subsets of Librispeech, and more than 30\% relative improvement for real-world in-house voice search compared to a wav2vec 2.0 baseline. 

\paragraph{WavLM}

\edit{WavLM emphasizes spoken content modeling and speaker identity preservation~\cite{chen2021wavlm}. It is largely identical to HuBERT, but introduces two useful extensions.}

\edit{First, it extends the Transformer self-attention mechanism with a so-called \emph{gated relative position bias}. The bias is added prior to the softmax normalization of the attention weights. For the attention weight at $i,j$, the bias is computed based on the input to the Transformer layer at the current time step $i$ and also incorporates a relative positional embedding for $i-j$. The authors find that this extension improves performance on phoneme and speech recognition tasks.}

%First, it extends the Transformer self-attention with a content-based gated relative position bias which improves the model's capability on recognition tasks compared to the convolutional context embedding \edit{used} in HuBERT and wav2vec 2.0. The gates adjust the relative position bias by conditioning on the current speech content \edit{which differs from the usual approach of learning a fixed positional encoding that is independent of the input}. Intuitively, the same distance offset between two frames tends to play different roles if one frame is silent and the other belongs to a speech segment.
%WavLM uses 320 bucketed positions for the range of possible offsets.
%The relative positional embedding parameters are shared across all layers.

\edit{Second, it uses an utterance mixing strategy where signals from different speakers are combined to augment the training data.} \edit{Specifically, random subsequences from other examples in the same batch are scaled and added to each input example.
%each example in a batch is combined with random subsequences from other examples in the same batch.
%Mixed utterances are selected from the same training minibatch, then randomly cropped and scaled by a sampled source energy ratio.
Only the targets corresponding to the original example are predicted during pre-training. Thus, the model learns to filter out the added overlapping speech.}

Most SSL methods are trained on data \edit{where each example only contains speech from a single person}; therefore, they can perform subpar on multispeaker tasks like speaker separation and diarization. 
%The content information corresponding to the main speaker is used as the target during the masked prediction pre-training.

%The pre-training data used for WavLM extends the sources used for HuBERT and wav2vec 2.0 to reach a total of 94k hours of public audio.
The WavLM model achieved substantial improvements on the speech separation, speaker verification and speaker diarization tasks in the SUPERB benchmark, while also performing well on many other tasks compared to HuBERT and wav2vec 2.0.
%It also improved upon \edit{the} HuBERT and wav2vec 2.0 models across all ten tasks of the SUPERB benchmark while maintaining a comparable performance on ASR. 

\paragraph{data2vec} \edit{Motivated by the success of using an exponential moving average (EMA) teacher for self-supervised visual representations~\cite{grill2020byol, dino}, the data2vec model~\cite{data2vec} computes targets $Y$ using an EMA of its own parameters. The targets are constructed by averaging hidden representations of the top $k$ layers of the EMA teacher network applied to unmasked inputs. Here, we denote this jointly as $\bar{f}_2(\cdot)$.

The data2vec model was proposed for different data modalities, but for audio it uses an architecture similar to wav2vec 2.0 and HuBERT with a convolutional module $f_1(\cdot)$, a Transformer $f_2(\cdot)$ and masking applied to the Transformer input.
\begin{align}
    z_t &= f_1(X_{[t-u,t+u]}) \enspace , \\
    H &= f_2(m(Z)) \enspace , \\
    Y &= \bar{f}_2(Z) \enspace .
\end{align}
The teacher network $\bar{f}_2(\cdot)$ is a copy of the Transformer of the student network but with the parameters at training step $i$, $\theta_{\text{teacher},i}$, given by an EMA of the student parameters over all previous training steps.
\begin{equation}
    \theta_{\text{teacher},i} = \begin{cases}
        \theta_{\text{student},0} & i = 0 \\
        \gamma\theta_{\text{student},i} + (1-\gamma)\theta_{\text{teacher},i-1} & i > 0 
    \end{cases} \enspace ,
\end{equation}
\noindent where $\theta_{\text{student},i}$ are the parameters of the student network at training step $i$, updated via gradient descent, and $\gamma$ is the EMA decay rate. 

The data2vec model uses a regression loss between target and prediction. Specifically, to reduce sensitivity to outliers and prevent exploding gradients, it uses the smoothed \lone{} loss \cite{fastrcnn},
\begin{equation}
\mathcal{L}_t =
    \begin{cases}
      \frac{1}{2}(y_t - h_t)^2 / \eta\text{,} & |y_t - h_t| \leq \eta\\
      |y_t - h_t| - \frac{1}{2}\eta\text{,} & \text{otherwise}
    \end{cases} \enspace ,
\end{equation}
\noindent where the hyperparameter $\eta$ controls the transition from a squared loss to an \lone{} loss.}

\edit{The data2vec approach was shown to work well for representation learning with either speech, images or text data.} It is the first approach to achieve competitive results when trained on \edit{any} one of the three modalities. 

\subsubsection{Challenges}
The iterative nature of pre-training for the HuBERT and wavLM could present a practical inconvenience when working with large volumes of data. Another challenge for these models centers around the quality of the initial vocabulary generated from MFCC features. 
% Although \edit{a} masked prediction loss enables the models not to be \edit{limited} by the quality of their initial target units, experiments show that initial teachers with higher quality, e.g., Gaussian Mixture Models (GMMs) instead of \kmeans{}, provide non-negligible performance gains.
The data2vec approach improves over other predictive models \edit{by allowing the targets to improve continuously via the EMA teacher network}; however, \edit{student-teacher approaches inflate the existing computational challenges of very large models and may necessitate the use of methods that decrease instantaneous memory utilization such as mixed precision training, model parallelism and model sharding \cite{paszke2019pytorch}.}

\subsection{Learning from multi-modal data}
% {\color{blue} Karen}\\
\subsubsection{Motivation}
%Benefit from the naturally parallel data without annotation
Multiple modalities are useful in many settings, where each modality provides information that is complementary to other modalities.  Multi-modal work includes supervised settings, such as audio-visual ASR~\cite{petajan1984automatic,potamianos2003recent} and person identification~\cite{aleksic2006audio} which have been studied for decades.  In this section, we focus only on unsupervised representation learning from multi-modal data.

One of the motivations for learning from multiple modalities is that it can reduce the effect of noise, since noise in different modalities is likely to be largely independent or uncorrelated.  In addition, learning from speech data with accompanying signals such as images or video can help learn representations that encode more semantic information. Such ``grounding" signals can contain supplementary information that can be used by models to infer the content of the speech. Human language learning provides a proof of concept for this, as it is believed that infants benefit from the visual modality when learning language~\cite{legerstee1990infants}. Early computational models of multi-modal language learning were motivated by (and tried to emulate) human learning of language in the context of the visual surroundings~\cite{roy1999learning}.

\subsubsection{Approaches}
We define two broad classes of approaches in this area. Specifically, depending on what type of multi-modal data is involved we refer to ``intrinsic" or ``extrinsic" modalities.

%\paragraph{Intrinsic modalities} 
\edit{{\it Intrinsic modalities} are} 
%refer to 
modalities produced directly by the speech source.  Examples of intrinsic modalities (besides the speech audio) include images or video of the speaker's face~\cite{lee2004avicar,chung2016lip}, lip-movement~\cite{shi2022learning}, articulatory flesh point measurements~\cite{westbury1990x,wrench2001new}, or simultaneous \edit{magnetic resonance imaging (MRI)} scans~\cite{narayanan2011multimodal}.  Typically, learning from multiple intrinsic modalities is done so as to improve robustness to noise, since acoustic noise is likely to be uncorrelated with the other \edit{modality(ies)}.  This type of representation learning is often referred to as ``multi-view learning" because the multiple intrinsic modalities can be seen as multiple views of the same content. Some typical approaches in this category include
\begin{itemize}
    \item Multi-view autoencoders and variations~\cite{ngiam2011multimodal,badino2016integrating},
    \item Multi-modal deep Boltzmann machines~\cite{srivastava2012multimodal},
    \item Canonical correlation analysis (CCA)~\cite{hotelling1936relations} and its nonlinear extensions~\cite{andrew2013deep,wang2015deep,wang2016deep,michaeli2016nonparametric,Melzer_01a,LaiFyfe00a,LaiFyfe99a,BachJordan05a,wang_15a},
    \item Multi-view contrastive losses~\cite{HermanBlunsom14a,Huang_13b},
    \item More recently, audio-visual extensions of masked prediction methods~\cite{shi2022learning,shi2022robust}, specifically Audio-Visual HuBERT (AV-HuBERT)~\cite{shi2022learning}.
\end{itemize}

%\paragraph{Extrinsic modalities} refer to
\edit{\it Extrinsic modalities are} modalities that are not produced by the same source but still provide context for each other. A typical example is an image and its spoken caption: The image tells us what the speech is likely describing, so a representation model that takes both modalities into account will hopefully encode more of the meaning of the speech than a single-modality model. 
There has recently been \edit{a} surge of datasets collected for this purpose, usually consisting of images and spoken captions, the audio and image frames in a video, or video clips with their spoken descriptions. A recent review of datasets, as well as methods, in this category is provided by Chrupa\l{}a~\cite{chrupala2021visually}. 

Typical approaches involve learning a neural representation model for each modality, with a multi-modal contrastive loss that encourages paired examples in the two modalities to have similar representations while unpaired examples remain different~\cite{synnaeve+etal_nipsworkshop14,harwath+etal_nips16,harwath+glass_asru15,merkx2019language,rouditchenko2020avlnet,peng2022fast}. 
Other choices include training with a masked margin softmax loss~\cite{ilharco2019large,sanabria2021talk} or a masked prediction loss~\cite{MultimodelAmazonICASSP22}.  Such models are typically evaluated on cross-modal retrieval, although some work has also used the models for other downstream tasks such as the ZeroSpeech and SUPERB benchmark tasks~\cite{peng2022self}. 
Analyses of such models have found that, despite the very high-level learning objective of matching speech with a corresponding image (or other contextual modality), such models often learn multiple levels of linguistic representations from the shallowest to the deepest model layers~\cite{harwath2019learning,chrupala2017representations,scharenborg+etal_icassp18}.  They are also able to learn word-like units~\cite{harwath2018jointly,peng2022word,wang2020dnn} and can be used for cross-lingual retrieval, by considering the visual signal as an ``interlingua"~\cite{harwath2018vision,havard2019models,kamper+roth_sltu18}. 
In some settings, even in the presence of some amount of textual supervision (i.e., the speech is transcribed), visual grounding still helps learn a better representation for retrieval~\cite{pasad2019contributions}. 

There has also been growing interest in learning joint speech and text representations using paired and unpaired data. The SLAM approach~\cite{bapna2021slam} is an example where speech and text are first represented using two separate pre-trained encoders followed by a multi-modal encoder to build the joint representations. The entire model is trained using a multi-task loss including two supervised and two self-supervised tasks.

%Benefiting from natural pairing of modalities in videos. Audio and text meta-data in FB videos. \\
%oWE can add it here or in the TTS subsection:
% link in comment to avoid compilation errors: https://arxiv.org/pdf/2011.11564.pdf \\

\subsubsection{Challenges}
One of the challenges of using multi-modal approaches is that the multi-modal data they rely on is often in shorter supply than single-modality data. In addition, multi-modal data is typically drawn from specific domains, for example domains involving descriptions of visual scenes. It is not clear how well the learned speech representations apply to other speech domains that are not necessarily describing or situated in a visual scene, and this question requires further study.

\subsection{Acoustic Word Embeddings}

% {\color{blue} Karen.} \\
% \kl{I moved this section here from the zero-resource section.}\\

Most of the representation learning techniques discussed in the preceding sections are aimed at learning frame-level representations. For some purposes, however, it may be useful to explicitly represent longer spans of speech audio of arbitrary duration, such as phone, word, or phrase-level segments.  For example, searching within a corpus of recorded speech for segments that match a given (written or spoken) query can be seen as finding segments whose representations are most similar to that of the query~\cite{levin+etal_icassp15,guoguo+etal_icassp15,Chung2016AudioWord2Vec,settle2017query}; word embeddings can be defined by pooling representations of instances of a given word~\cite{chung2018speech2vec}; unsupervised segmentation and spoken term discovery can be seen as a problem of detecting and clustering segments~\cite{kamper2017embedded,Kamper17BESGMM}; and even ASR can be viewed as the problem of matching written word representations to representations of audio spans~\cite{maas+etal_icmlwrl12,bengio+heigold_interspeech14,settle2019_a2w}.  

Several lines of work have begun to address the problem of learning representations of spans of speech, especially word segments, typically referred to as {\it acoustic word embeddings}.  Early work on unsupervised acoustic word embeddings defined them as vectors of distances from the target segment to a number of pre-defined ``template" segments~\cite{levin+etal_asru13}. Later work used variants of neural autoencoders~\cite{Chung2016AudioWord2Vec,holzenberger2018learning,kamper2019truly,pengcorrespondence}. These are often evaluated on word discrimination, that is the task of determining whether two word segments correspond to the same word or not~\cite{carlin2011rapid}. This task can be thought of as a proxy for query-by-example search, since the basic operation in search is to determine whether a segment in the search database matches a query segment, and has been used for evaluation of both frame-level (e.g.,~\cite{kamper2015unsupervised}) and word-level~\cite{levin+etal_asru13,kamper2016deep} representations.  

% \kl{I left out much of the work on acoustic word embeddings, since it is in a supervised or at least weakly supervised setting.  I could add a quick note that a lot of other work exists, and add citations, but not sure that's needed for this paper.}

Since most work on acoustic word embeddings preceded the very recent wave of new self-supervised frame-level representations, one question is whether word (or more generally segment) embeddings could be derived more simply by pooling self-supervised frame-level representations, as has been done for text span embeddings by pooling over word embeddings~\cite{toshniwal2020cross,wang2021phrase}. Some initial results suggest that at least very simple pooling approaches like downsampling and mean or max pooling are not successful~\cite{van2021comparison,pengcorrespondence}, but more work is needed to reach conclusive results.

% The model learns through reconstructing a spectrogram slice from past and future slices~\cite{tagliasacchi2019self, tagliasacchi2020pre}. %Similar to DecoAR?
% The TemporalGap~\cite{tagliasacchi2019self, tagliasacchi2020pre} approach learns through estimating the temporal gap between two short audio segments extracted at random from the same audio clip. %Can this approach be considered as generative?
% Representations are learned through reconstructing the phase of the short-time Fourier transform from its magnitude~\cite{quitry2019learning}. %This is audio, not speech.

%We put the work of audio word embedding here.
%This can be a link to zero speech and further speech2semantic applications.
%Originally, audio word embedding was targeted at query-by-example, so the researchers want the embedding to contain the phoneme information (for example,  https://arxiv.org/abs/1510.01032, https://arxiv.org/abs/1603.00982). 
%Then there is some work targeting at including semantic information in the embedding (for example, https://arxiv.org/abs/1803.08976).

% should we add a section on Learning representation through multi-tasking}
 %Reviewer: Hung-yi, Daniel

\section{Benchmarks for Self-supervised Learning}
\label{section:benchmark}
% {\color{blue} Daniel}\\

% {\color{blue} Reviewer: SW, Hung-yi Lee, Abdo }\\

%\subsection{Benchmarks for self-supervised learning approaches} 

%This paper is about the size of self-supervised model. I think it can also be considered as benchmark.
%https://www.amazon.science/publications/scaling-effect-of-self-supervised-speech-models

%We have shown in the previous sections various methodologies to learn speech representations from unlabeled corpora. This section surveys the available datasets to learn the representations and evaluate the learned ones. We also summarize several works and their results to demonstrate how effective the learned representations are via providing robust model initialization for various downstream tasks.
The previous sections presented various methodologies by which to learn speech
representations from unlabeled corpora. This section surveys the datasets
available to learn and evaluate these representations. We also summarize
several studies and their results to demonstrate the usefulness of the learned
representations for various downstream tasks. 

\subsection{Datasets only for pre-training} 
\Cref{table:datasets} summarizes datasets used for pre-training SSL techniques
in the literature. These datasets are usually large but with limited or no
labels. Libri-light (LL)~\cite{kahn2020libri}, one of these datasets, is
derived from audiobooks that are part of the
LibriVox\footnote{https://librivox.org/ \label{librivox}} project. LL contains
60k hours of spoken English audio tagged with SNR, speaker ID, and genre
descriptions. The speech examples in Audioset~\cite{gemmeke2017audio}, which
consists of over 2M 10-second YouTube video clips human-annotated with 632
audio events, have also been used for pre-training. Audioset has 2.5k hours
of audio of varying quality, different languages, and sometimes multiple sound
sources. AVSpeech~\cite{ephrat2018looking} is another large-scale audio-visual
dataset used in SSL research, comprising 4.7k hours of clips from a wide
variety of languages. 
%from 150k distinct speakers
Each clip contains a visible face and audible sound originating from a single
speaker without interfering background signals. The 3100-hour audio part of
AVSpeech has been used to learn audio-only 
representations~\cite{kawakami2020learning}. The Fisher corpus~\cite{cieri2004fisher} collects
over 2k hours of conversational telephone speech, 1k hours of which is utilized
for pre-training~\cite{jiang2021further}. Industrial researchers have also
begun to build large-scale datasets for learning speech representations.
For instance, 10k hours of real-world far-field English voice commands for
self-supervised pre-training have been collected at 
Amazon~\cite{sadhu21_interspeech}. 

In addition to these English and multilingual efforts, researchers have also
collected corpora for pre-training Chinese speech representations. Didi Dictation
and Didi Callcenter~\cite{jiang2019improving, jiang2021further} are two
internal datasets containing respectively 10k hours of read speech collected
from mobile dictation application and 10k hours of spontaneous phone calls
between users and customer service staff.

\subsection{Datasets for both pre-training and evaluation}\label{sec:datasets}
Several datasets that provide both speech and associated transcripts and
speaker labels have also been used to develop SSL techniques by enabling
in-domain pre-training and evaluation. Such datasets are also listed in
\cref{table:datasets}. One of the most commonly used datasets in this category
is LibriSpeech (LS)~\cite{panayotov2015librispeech}, a labeled corpus
containing 960 hours of
%16 kHz sampled 
read English speech, which is also derived from an open-source audiobooks
project.\textsuperscript{\ref{librivox}} The corpus consists of subsets
\textit{train-clean-100}, \textit{train-clean-360}, \textit{train-other-500},
\textit{dev-clean}, \textit{dev-other}, \textit{test-clean}, and
\textit{test-other} used for training, development, and testing, respectively.
Subsets tagged with \textit{other} are more challenging utterances from
speakers that yield higher WER as measured with previously built models. LS is
used for unsupervised representation pre-training by ignoring its labels, and
can also be utilized to evaluate the performance of representation on ASR, phoneme recognition (PR),
phoneme classification (PC), and speaker identification (SID) tasks. Wall Street Journal (WSJ)~\cite{paul1992design} is another
widely adopted, labeled corpus for pre-training. Its labels can evaluate
performance for ASR, PR, PC, and SID. The original WSJ corpus contains 400 hours
of English read speech data, and today its \textit{si284} (81 hours),
\textit{dev93}, \textit{eval92} subsets are the most-used partitions for
unsupervised training, development, and test, respectively.  The \textit{si84}
(15 hours) partition is also used for training.

The speech community also utilizes multilingual corpora. These are often
large-scale, which facilitates pre-training, but are also partially labeled for ASR
evaluation (PC and PR can be enabled via phone-level forced alignment). These
corpora include Common Voice (CV)~\cite{ardila2020common}, Multilingual
LibriSpeech (MLS)~\cite{pratap20_interspeech}, VoxPopuli 
(VP)~\cite{wang-etal-2021-voxpopuli}, and BABEL (BBL)~\cite{gales2014speech}. CV is
an open-source, multi-language, growing dataset of voices containing 11k hours
of audio from 76 languages as of the date this review was written (Common Voice
corpus~7.0). Researchers usually use part of this for pre-training (e.g., 7k
hours/60 languages in \cite{babu2021xlsr} and 430 hours/29 languages in
\cite{kawakami2020learning}) or evaluation. 
%Demographic metadata like age, sex, and accent is also provided.
MLS derives content from read audiobooks of LibriVox and contains data in eight
European languages for a total of 50k hours of audio. VP comprises a total of 400k
hours of parliamentary speech from the European parliament in 23 European
languages.
The entire dataset~\cite{babu2021xlsr} or a 24k-hour 
portion~\cite{chen2021unispeechsat, chen2021wavlm} thereof has been used for pre-training. BBL
consists of 1k hours of conversational telephone speech in 17 African and Asian
languages.

Several datasets, including GigaSpeech~\cite{chen21o_interspeech}, TED-LIUM~3
(TED3)~\cite{hernandez2018ted}, TED-LIUM~2 (TED2)~\cite{rousseau2012ted},
Switchboard (SWB)~\cite{godfrey1992switchboard}, 
TIMIT~\cite{garofolo1993timit}, and VoxLingua107~\cite{valk2021voxlingua107}, are
labeled and conventionally used for evaluation, while their audio streams are
also aggregated to build diversified and large-scale corpora for unsupervised
pre-training~\cite{kawakami2020learning, song20d_interspeech, babu2021xlsr}.
GigaSpeech is a multi-domain English ASR corpus with 33k hours of audio
collected from audiobooks, podcasts, and YouTube. A subset of 10k audio is
transcribed. TED2 comes with 118 hours of English speech extracted from
TED conference talks and its transcription for evaluating ASR. Its recordings
are clear but with some reverberation. TED3 is an extension of TED2 and
comprises 450 hours of talks. SWB is a 260-hour conversational speech
recognition dataset containing two-sided telephone conversations.
%The dataset was recorded in 1990 with a lower (8k) sampling rate than the other corpora, and thus presents a challenging recognition problem.
The TIMIT corpus was designed to provide read speech data and its word and
phone-level transcriptions for acoustic-phonetic studies. It contains recordings
%from 630 speakers with 8 dialects 
in American English. Compared to the previous corpora labeled for ASR
evaluation, VoxLingua107 consists of 6.6k hours of audio in 107 languages and
is annotated for language identification. Beyond the original purpose of
evaluation, these corpora are also used in pre-training to improve the
generalizability of learned representations.

%, and each person read 10 phonetically-rich sentences.

For the purpose of pre-training and evaluating Mandarin speech representations,
the authors of \cite{jiang2019improving, jiang2021further} also compiled Open Mandarin, 
an open-source Mandarin dataset of 1.5k hours of speech from
the Linguistic Data Consortium (LDC) and
OpenSLR.\footnote{\url{https://openslr.org}} Open Mandarin consists of the HKUST
Mandarin Telephone Speech Corpus (HKUST, 200 hours of spontaneous 
speech, of which   % AMH: check
%from over 2100 speakers in mainland China. 
168 hours of audio is used for pre-training; the development and test sets are
excluded.)~\cite{liu2006hkust}, AISHELL-1~\cite{bu2017aishell} (178 hours of
read speech),
%recorded in a quiet indoor environment from 400 speakers with different accents
aidatatang 200zh (200 hours, read speech)~\cite{aidatatang}, MAGICDATA
Mandarin Chinese Read Speech Corpus (755 hours, read speech)~\cite{magic_data},
Free ST Chinese Mandarin Corpus (ST-CMDS, 100 hours, read speech)~\cite{st_cmds},
and Primewords Chinese Corpus Set 1 (100 hours, read speech)~\cite{primewords_201801}. 
Both HKUST and AISHELL-1 are labeled and are suitable for ASR evaluation.

\subsection{Datasets for evaluation}

Besides the aforementioned datasets, conventional speech processing benchmarks
are also used to evaluate self-supervised representations. Studies leverage
% Hub5'00, 
  Hub5,     % AMH: check
DIRHA, and CHiME-5 to measure the efficacy of representations in ASR.
The Hub5 evaluation (LDC2002T43 and LDC2002S09, also referred to as the NIST 2000
Hub5 English evaluation set) contains 40 transcribed English telephone
conversations only for testing, where 20 are from conversations collected in
SWB studies but not released with the SWB dataset, and the rest are from CallHome
American English Speech (LDC97S42). DIRHA~\cite{ravanelli2015dirha}, short for
Distant-speech Interaction for Robust Home Applications, is a database composed
of utterances sampled from WSJ, speech of keywords and commands, and
phonetically-rich sentences. These utterances are read by UK and US English
speakers and recorded with microphone arrays. CHiME-5~\cite{barker2018fifth} is
a challenge that aims to advance robust ASR and presents a dataset of natural
conversational speech collected under a dinner party scenario with microphone
arrays. A team at Amazon Alexa also recorded and transcribed a corpus of 1k
hours of audio for model training and evaluation~\cite{sadhu21_interspeech}.

Researchers also evaluate representations for sentiment analysis with the
INTERFACE~\cite{hozjan2002interface} and MOSEI (CMU Multimodal Opinion
Sentiment and Emotion Intensity)~\cite{zadeh2018multimodal} datasets. INTERFACE
is an emotional speech database for Slovenian, English, Spanish, and French,
and contains six emotions: anger, sadness, joy, fear, disgust, and
surprise, plus neutral. MOSEI is comprised of sentence-level sentiment
annotations of 65 hours of YouTube videos
%from 1000 speakers having 
using emotion categories similar to INTERFACE, but replacing joy with
happiness.

In addition, datasets employed to demonstrate the benefit of SSL
representations on various tasks include VCTK \cite{veaux2016superseded} and VoxCeleb1 \cite{nagrani17_interspeech} for SID/ASV (automatic speaker verification) tasks, FSC (Fluent Speech Commands) \cite{lugosch19_interspeech} for IC (intent classification), QUESST
(QUESST 2014) \cite{anguera2015quesst2014} for QbE (query by example), LS En-Fr \cite{kocabiyikoglu2018augmenting} and CoVoST-2 \cite{wang2020covost} for ST (speech translation), and ALFFA and OpenSLR-multi for multilingual ASR. The VCTK corpus includes speech
data with 109 English speakers of various accents, each reading out about 400
sentences sampled from newspapers. VoxCeleb1 is an audio-visual dataset
comprised of short YouTube clips containing human speech. It consists of 1251
unique speakers and 352 hours of audio. FSC contains utterances of spoken
English commands that one might use for a smart home or virtual assistant, and
is used to evaluate the performance of a spoken language understanding system.
The QUESST search dataset comprises spoken documents and queries in 6~languages
to measure the capability of models in spotting spoken keywords from documents.
LS En-Fr is a dataset augmenting existing LS monolingual utterances with
corresponding French translations to train and evaluate English-French machine
translators. CoVoST-2 is a multilingual speech translation benchmark based on
CV. It provides data for translating from English into 15 languages and from 21
languages into English, and has a total of 2.9k hours of speech. The ALFFA
project\footnote{\url{http://alffa.imag.fr}} collects speech of African
languages to 
% drive 
  promote    % AMH: check
the development of speech technologies in Africa, and
\cite{kawakami2020learning} leverages four African languages collected in the
project for evaluation: Amharic~\cite{tachbelie2014using}, Fongbe~\cite{laleye2016first},
Swahili~\cite{gelas2012developments}, and Wolof~\cite{gauthier2016collecting}.
In the same work~\cite{kawakami2020learning}, the authors
further select 21 phonetically diverse languages from OpenSLR to evaluate the
generalizability of SSL representations across languages. We denote the
collection as OpenSLR-multi below.

Last, \cite{tagliasacchi2020pre} puts together five datasets 
(MUSAN~\cite{snyder2015musan}, Bird Audio Detection~\cite{stowell2019automatic},
Speech Commands~\cite{warden2018speech}, Spoken Language 
Identification~\cite{oponowiczspoken}, and TUT Urban Acoustic Scenes 2018~\cite{dcase2018}) plus
an SID task built with the LS \textit{train-clean-100} subset to evaluate the
capability of representations on audio event detection.
\cite{quitry2019learning} employs the NSynth dataset~\cite{engel2017neural} on top
of the six for benchmarking. As many of the datasets are built for research in
audio processing, we here provide only a list of these datasets for reference.

\begin{table*}[ht]
  \centering
  \scriptsize %footnotesize %small, scriptsize, tiny
  \caption{Summary of datasets used in pre-training (denoted as PT) or
  evaluating (denoted as EV) SSL techniques in the literature. The languages and
  sizes of the datasets are provided in columns 3 and 4. Column 5 lists the tasks each
  dataset is used to evaluate. We use the following abbreviations:
  \textbf{EN}: English; 
% \textbf{Multi}: multilinguality; 
  \textbf{Multi}: multilingual;     % AMH: check
  \textbf{ZH}: Chinese;
  \textbf{Fr}: French; \textbf{ASR}: automatic speech recognition; \textbf{PR}:
  phoneme recognition; \textbf{PC}: phoneme classification; \textbf{SID}:
  speaker identification; \textbf{ASV}: automatic speaker verification;
  \textbf{Sentiment}: sentiment analysis; \textbf{ST}: speech translation;
  \textbf{QbE}: query by example or spoken term detection; \textbf{IC}: intent
  classification; \textbf{AED}: audio event detection; and \textbf{LID}: language
  identification. We distinguish \textbf{PR} from \textbf{PC} based on whether
  the inference is made at the phone level sequentially or the frame level separately.
  \textbf{SID} and \textbf{ASV} both evaluate model capability in encoding
  speaker information; \textbf{SID} classifies one utterance into a pre-defined
  set of speaker labels, whereas \textbf{ASV} infers whether a given pair of
  utterances was uttered by the same speaker.}
  \label{table:datasets}
  {\renewcommand*\arraystretch{1.2}
  \begin{tabular}{llllll}
    \toprule
    Dataset & Purpose & Lang. & Size [hours] & Task & License \\
    \midrule
    % \multirow{2}{*}[0em]{wav2vec 2.0}
    LibriLight (LL) & PT & EN & 60k & - & MIT License \\ \hline
    AudioSet & PT & Multi & 2.5k & - & CC BY 4.0 \\ \hline %Bidir-CPC, Audio2Vec
    AVSpeech & PT & Multi & 3.1k & - & CC BY 4.0 \\ \hline %Bidir-CPC
    Fisher & PT & EN & 2k/1k \cite{jiang2021further}  & - & Linguistic Data Consortium (LDC) \\ \hline % MPC

    Alexa-10k & PT & EN & 10k & - & Not released \\ \hline % wav2vec-c
    Didi Callcenter & PT & ZH & 10k & - & Not released \\ \hline % MPC
    Didi Dictation & PT & ZH & 10k & - & Not released \\ \hline % MPC

    LibriSpeech (LS) & PT/EV & EN & 960 & ASR/PR/PC/SID & CC BY 4.0 \\ \hline
    Wall Street Journal (WSJ) & PT/EV & EN & 81 & ASR/PR/PC/SID & Linguistic Data Consortium (LDC)  \\ \hline

    Common Voice (CV-dataset) & PT/EV & Multi & 11k/7k \cite{babu2021xlsr}/430 \cite{kawakami2020learning} & ASR/PR/PC & CC0 \\ \hline % mention the split, CPC-modified, Bidir-CPC
    Multilingual LS (MLS) & PT/EV & Multi & 50k & ASR & CC BY 4.0 \\ \hline
    VoxPopuli (VP) & PT/EV & Multi & 400k \cite{babu2021xlsr}/24k \cite{chen2021unispeechsat, chen2021wavlm} & ASR & CC0 \\ \hline
    BABEL (BBL) & PT/EV & Multi & 1k & ASR & IARPA Babel Agreement  \\ \hline

    GigaSpeech & PT/EV & EN & 40k/10k \cite{chen2021unispeechsat, chen2021wavlm} & ASR & Apache-2.0 License \\ \hline
    TED-LIUM 3 (TED3) & PT/EV & EN & 450 & ASR & CC BY-NC-ND 3.0 \\ \hline %Bidir-CPC
    TED-LIUM 2 (TED2) & PT/EV & EN & 118 & ASR & CC BY-NC-ND 3.0  \\ \hline % speech-XLNet
    Switchboard (SWB) & PT/EV & EN & 260 & ASR & Linguistic Data Consortium (LDC) \\ \hline % MPC, Bidir-CPC
    TIMIT & PT/EV & EN & 4 & ASR/PR/PC & Linguistic Data Consortium (LDC) \\ \hline
    VoxLingua107 & PT/EV & Multi & 6.6k & LID & CC BY 4.0 \\ \hline

    Open Mandarin & PT/EV & ZH & 1.5k & ASR & \makecell[l]{CC BY-NC-ND 4.0, Apache License v.2.0,\\ Linguistic Data Consortium (LDC)} \\ \hline % MPC
    HKUST & PT/EV & ZH & 168/200 & ASR & Linguistic Data Consortium (LDC) \\ \hline % MPC
    AISHELL-1 & PT/EV & ZH & 178 & ASR & Apache License v.2.0 \\ \hline % MPC
    Hub5'00 & EV & EN & 13 & ASR & Linguistic Data Consortium (LDC) \\ \hline % MPC - https://arxiv.org/pdf/1801.00059.pdf
    DIRHA & EV & EN & 11 & ASR & See link for details\textsuperscript{\ref{footnote:DIRHA_license}} \\ \hline % PASE
    % from the paper Learning Problem-agnostic Speech Representations from Multiple Self-supervised Tasks - "in Section 4.3 we use the DIRHA dataset [36]. Training and validation sets are based on the original WSJ-5k corpus (consisting of 7138 sentences uttered by 83 speakers) that is contaminated with a set of impulse re- sponses measured in a real apartment. The test set is composed of 409 WSJ sentences uttered by six American speakers and is based on real recordings in a domestic environment with a reverberation time of 0.7 s and an average signal-to-noise ratio of about 10 dB"
    CHiME-5 & EV & EN & 50 & ASR & See link for details\textsuperscript{\ref{footnote:CHiME_license}}\\ \hline % PASE+
    Alexa-eval & EV & EN & 1k & ASR & Not released \\ \hline % wav2vec-c
    INTERFACE & EV & Multi & 16 & Sentiment & No information \\ \hline % PASE
    MOSEI & EV & EN & 65 & Sentiment & See link for details\textsuperscript{\ref{footnote:MOSEI_license}}\\ \hline % MockingJay
    VCTK & EV & EN & 44 & SID/ASV & CC BY 4.0 \\ \hline % PASE
    VoxCeleb1 & EV & Multi & 352 & SID/ASV & CC BY 4.0 \\ \hline
    Fluent Speech Commands (FSC) & EV & EN & 14.7 & IC & CC BY-NC-ND 4.0 \\ \hline
    QUESST 2014 (QUESST) & EV & Multi & 23 & QbE & No information \\ \hline
    LS En-Fr & EV & En-Fr & 236 & ST & CC BY 4.0 \\ \hline % APC (tera-9), https://github.com/alicank/Translation-Augmented-LibriSpeech-Corpus
    CoVoST-2 & EV & Multi & 2.9k & ST & CC0 \\ \hline

    ALFFA & EV & Multi & 5.2--18.3 & ASR-multi & MIT License \\ \hline %Bidir-CPC
    OpenSLR-multi & EV & Multi & 4.4--265.9 & ASR-multi & \makecell[l]{CC BY-SA 3.0 US, CC BY-SA 4.0, CC BY 4.0, \\ CC BY-NC-ND 4.0, Apache License v.2.0}\\ \hline %Bidir-CPC
    AED datasets & EV & - & - & AED & \makecell[l]{CC BY 4.0 (MUSAN, Speech Commands, NSynth, Bird Audio\\ Detection),  CC0 (Spoken Language Identification),\\ Non-Commercial (TUT)} \\
    \bottomrule
  \end{tabular}}
\end{table*}

\addtocounter{footnote}{1}
\footnotetext{\label{footnote:DIRHA_license}\url{https://dirha.fbk.eu/node/107}}
\addtocounter{footnote}{1}
\footnotetext{\label{footnote:CHiME_license}\url{https://chimechallenge.github.io/chime6/download.html}}
\addtocounter{footnote}{1}
\footnotetext{\label{footnote:MOSEI_license}\url{https://github.com/A2Zadeh/CMU-MultimodalSDK/blob/master/LICENSE.txt}}

\begin{table*}[ht]
  \centering
  \scriptsize
  \caption{A summary of common experiment settings for various SSL
  evaluations (Part 1). Networks are usually pre-trained with SSL
  techniques, augmented with prediction heads, and fine-tuned (or trained) with
  labeled data in downstream tasks for benchmarking. The 
  \textbf{Pre-training corpus}, \textbf{Training (fine-tuning)}, and
  \textbf{Test} columns list the datasets used in each work, and the \textbf{Task}
  column lists the tasks performed in the corresponding papers. We follow
  the abbreviation introduced in \cref{table:datasets}. The \textbf{Transfer}
  column indicates whether the SSL technique is evaluated by its capability for
  transfer learning, i.e., different datasets are utilized for pre-training and
  fine-tuning. The \textbf{Fine-tuning labels used} column summarizes the
  amount of labeled examples used in downstream fine-tuning.}
  \label{table:method_data_setting}
  {%\tabulinesep=1mm
  \renewcommand*\arraystretch{1.4}
  \begin{tabular}{p{0.12\textwidth}p{0.12\textwidth}p{0.06\textwidth}p{0.13\textwidth}p{0.13\textwidth}p{0.06\textwidth}p{0.20\textwidth}}
    \toprule
    \multirow{2}{*}{Work} & \multirow{2}{*}{\makecell[l]{Pre-training\\ corpus}} & \multirow{2}{*}{Task} & \multicolumn{2}{c}{Dataset} & \multirow{2}{*}{Transfer} & \multirow{2}{*}{Fine-tuning labels used} \\ \cline{4-5}
     &  &  & Training (fine-tuning) & Test & & \\
    \midrule

    % CPC
    \multirow{2}{*}[0mm]{CPC \cite{oord2018representation}} & \multirow{2}{*}[0mm]{LS 100 hrs} & PC & LS 100 hrs\textsuperscript{\ref{footnote:CPC-split}} & LS 100 hrs\textsuperscript{\ref{footnote:CPC-split}} & - & 80\textsuperscript{\ref{footnote:low-resource}} hrs \\ \cline{3-7}
    & & SID & LS 100 hrs\textsuperscript{\ref{footnote:CPC-split}} & LS 100 hrs\textsuperscript{\ref{footnote:CPC-split}} & - & 80\textsuperscript{\ref{footnote:low-resource}} hrs \\ \hline
    
    % PASE
    \multirow{2}{*}[-5mm]{PASE\cite{pascual2019learning}} & \multirow{2}{*}[-5mm]{LS 50 hrs \cite{ravanelli2018learning}} & SID & VCTK\textsuperscript{\ref{footnote:official-split}} & VCTK\textsuperscript{\ref{footnote:official-split}} & \checkmark & 44 hrs \\ \cline{3-7}
    & & Sentiment & INTERFACE\textsuperscript{\ref{footnote:interface-split}} & INTERFACE\textsuperscript{\ref{footnote:interface-split}} & \checkmark & 3 hrs \\ \cline{3-7}
    & & PR & TIMIT\textsuperscript{\ref{footnote:official-split}} & TIMIT\textsuperscript{\ref{footnote:official-split}} & \checkmark & 4 hrs \\ \cline{3-7}
    & & ASR & DIRHA\textsuperscript{\ref{footnote:official-split}} & DIRHA\textsuperscript{\ref{footnote:official-split}} & \checkmark & 11 hrs \\ \hline
    % Audio2Vec
    Audio2Vec \cite{tagliasacchi2020pre} & AudioSet & AED & 6 AED datasets\textsuperscript{\ref{footnote:AED-datasets}} & 6 AED datasets\textsuperscript{\ref{footnote:AED-datasets}} & \checkmark & See \cite{tagliasacchi2020pre} for details \\ \hline
    % APC
    \multirow{3}{*}[0mm]{APC \cite{chung2019unsupervised, chung2020generative}} & \multirow{3}{*}[0mm]{LS 360 hrs} & ASR & WSJ si284\textsuperscript{\ref{footnote:91-split-dev}} & WSJ dev93 & \checkmark & 72 hrs \\ \cline{3-7}
    & & ST & LS En-Fr\textsuperscript{\ref{footnote:official-split}} & LS En-Fr\textsuperscript{\ref{footnote:official-split}} & - & 236 hrs \\ \cline{3-7}
    & & SID & WSJ si284\textsuperscript{\ref{footnote:APC-split}} & WSJ si284\textsuperscript{\ref{footnote:APC-split}} & \checkmark & 65\textsuperscript{\ref{footnote:low-resource}} hrs \\ \hline
    % wav2vec
    \multirow{2}{*}[0mm]{wav2vec \cite{schneider2019wav2vec}} & \multirow{2}{*}[-0.5mm]{\makecell[l]{LS 80/960 hrs,\\ LS 960 hrs \\ \:+ WSJ si284}} & ASR & WSJ si284 & WSJ eval92 & \checkmark & 81 hrs\vspace{1mm} \\ \cline{3-7}
    & & PR & TIMIT\textsuperscript{\ref{footnote:official-split}} & TIMIT\textsuperscript{\ref{footnote:official-split}} & \checkmark & 4 hr\vspace{1mm} \\ \hline
    % Phase predict
    PhasePredict \cite{quitry2019learning} & AudioSet & AED & 7 AED datasets\textsuperscript{\ref{footnote:AED-datasets}} & 7 AED datasets\textsuperscript{\ref{footnote:AED-datasets}} & \checkmark & See \cite{quitry2019learning} for details \\ \hline
    % Bidir-CPC
    \multirow{3}{*}{Bidir-CPC\cite{kawakami2020learning}} & \multirow{3}{*}{\makecell[l]{LS 960 hrs,\\ CPC-8k\textsuperscript{\ref{footnote:CPC-8k}}}} & ASR & \makecell[l]{WSJ si284,\\LS 960 hrs,\\TED3\textsuperscript{\ref{footnote:official-split}}} & \makecell[l]{WSJ eval92,\\LS test-clean,\\LS test-other,\\TED3\textsuperscript{\ref{footnote:official-split}}, SWB\textsuperscript{\ref{footnote:official-split}}}  & \makecell[l]{Different\\ datasets for\\ training\\ and test} & 81/960/450 hrs \\ \cline{3-7} % cross domain transferring, SWB eval2000, TED3 training set size = 450 hrs
    & & \makecell[l]{ASR-multi} & ALFFA\textsuperscript{\ref{footnote:official-split}} & ALFFA\textsuperscript{\ref{footnote:official-split}} & \checkmark & 4 languages, 5.2--18.3 hrs\\ \cline{3-7}
    & & \makecell[l]{ASR-multi} & OpenSLR-multi\textsuperscript{\ref{footnote:official-split}} & OpenSLR-multi\textsuperscript{\ref{footnote:official-split}} & \checkmark & 21 languages, 4.4--265.9 hrs \\ \hline
    % vq-VAE

    % MockingJay
    \multirow{3}{*}{MockingJay \cite{liu2020mockingjay}} & \multirow{3}{*}{\makecell[l]{LS 360 hrs}} & PC & LS 360 hrs & LS test-clean & - & 0.36\textsuperscript{\ref{footnote:low-resource}}/1.8\textsuperscript{\ref{footnote:low-resource}}/3.6\textsuperscript{\ref{footnote:low-resource}}/18\textsuperscript{\ref{footnote:low-resource}}/45\textsuperscript{\ref{footnote:low-resource}}/360  hrs \\ \cline{3-7}
    & & SID & LS 100 hrs\textsuperscript{\ref{footnote:91-split-test}} & LS 100 hrs\textsuperscript{\ref{footnote:91-split-test}} & - & 90\textsuperscript{\ref{footnote:low-resource}} hrs \\ \cline{3-7}
    & & Sentiment & MOSEI\textsuperscript{\ref{footnote:official-split}} & MOSEI\textsuperscript{\ref{footnote:official-split}} & \checkmark & 65 hrs \\ \hline

    % CPC modified
    \multirow{2}{*}{CPC modified \cite{riviere2020unsupervised}} & LS 100 hrs & PC & LS 100 hrs\textsuperscript{\ref{footnote:CPC-split}} & \makecell[l]{LS 100 hrs\textsuperscript{\ref{footnote:CPC-split}}} & - & \makecell[l]{80\textsuperscript{\ref{footnote:low-resource}} hrs} \\ \cline{2-7}
    & \makecell[l]{LS 100 hrs,\\LS 960 hrs,\\LL 60k hrs} & PC & CV-dataset\textsuperscript{\ref{footnote:official-split}} & CV-dataset\textsuperscript{\ref{footnote:official-split}} & \checkmark & 1 hrs \\ \hline
    % vq-wav2vec
    \multirow{2}{*}[0mm]{vq-wav2vec \cite{Baevski2020vq-wav2vec}} & \multirow{2}{*}[0mm]{LS 960 hrs} & ASR & WSJ si284 & \makecell[l]{WSJ eval92} & \checkmark & \makecell[l]{81 hrs} \\ \cline{3-7}
    & & PR & TIMIT\textsuperscript{\ref{footnote:official-split}} & TIMIT\textsuperscript{\ref{footnote:official-split}} & \checkmark & 4 hrs \\ \hline
    % DeCoAR
    \multirow{2}{*}[0mm]{DeCoAR \cite{ling2020deep}} & \multirow{2}{*}[0mm]{\makecell[l]{LS 100/360/460/\\\:\:\:960 hrs,\\ WSJ si284}} & ASR & WSJ si284 & WSJ eval92 & - & \makecell[l]{25\textsuperscript{\ref{footnote:low-resource}}/40\textsuperscript{\ref{footnote:low-resource}}/81 hrs} \\ \cline{3-7}
    & & ASR & \makecell[l]{LS 100/360/\\\:\:\:460/960 hrs} & \makecell[l]{LS test-clean,\\ LS test-other} & - & 100/360/460/960 hrs \\ \hline
    % MT-APC
    \multirow{2}{*}[0mm]{MT-APC\cite{chung2020improved}} & \multirow{2}{*}[0mm]{LS 360 hrs} & ASR & WSJ si284\textsuperscript{\ref{footnote:91-split-dev}} & WSJ dev93 & \checkmark & 72 hrs \\ \cline{3-7}
    & & ST & LS En-Fr\textsuperscript{\ref{footnote:official-split}} & LS En-Fr\textsuperscript{\ref{footnote:official-split}} & - & 236 hrs \\ \hline
    % PASE+
    \multirow{3}{*}[0mm]{PASE+\cite{ravanelli2020multi}} & \multirow{3}{*}[0mm]{LS 50 hrs \cite{ravanelli2018learning}} & PR & TIMIT\textsuperscript{\ref{footnote:official-split}} & TIMIT\textsuperscript{\ref{footnote:official-split}} & \checkmark & 4 hrs \\ \cline{3-7}
    & & ASR & DIRHA\textsuperscript{\ref{footnote:official-split}} & DIRHA\textsuperscript{\ref{footnote:official-split}} & \checkmark & 11 hrs \\ \cline{3-7}
    & & ASR & CHiME-5\textsuperscript{\ref{footnote:official-split}} & CHiME-5\textsuperscript{\ref{footnote:official-split}} & \checkmark & 50 hrs \\ \hline

    % AALBERT
    \multirow{2}{*}[0mm]{AALBERT \cite{chi2020audio}} & \multirow{2}{*}[0mm]{\makecell[l]{LS 360 hrs}} & PC & LS 100 hrs\textsuperscript{\ref{footnote:APC-split}} & LS 100 hrs\textsuperscript{\ref{footnote:APC-split}} & - & 80\textsuperscript{\ref{footnote:low-resource}} hrs \\ \cline{3-7}
    & & SID & LS 360 hrs\textsuperscript{\ref{footnote:APC-split}} & LS 360 hrs\textsuperscript{\ref{footnote:APC-split}} & - & 288\textsuperscript{\ref{footnote:low-resource}} hrs \\
    \bottomrule
  \end{tabular}}
\end{table*}

\begin{table*}[ht]
  \centering
  \scriptsize
  \caption{A summary of common experiment settings for various SSL
  evaluations (Part 2). See the caption of \cref{table:method_data_setting} for a
  detailed description of all the abbreviations used in this table.}
  \label{table:method_data_setting_2}
  {%\tabulinesep=1mm
  \renewcommand*\arraystretch{1.4}
  \begin{tabular}{p{0.12\textwidth}p{0.12\textwidth}p{0.05\textwidth}p{0.18\textwidth}p{0.18\textwidth}p{0.05\textwidth}p{0.15\textwidth}}
    \toprule
    \multirow{2}{*}{Work} & \multirow{2}{*}{\makecell[l]{Pre-training\\ corpus}} & \multirow{2}{*}{Task} & \multicolumn{2}{c}{Dataset} & \multirow{2}{*}{Transfer} & \multirow{2}{*}{Fine-tuning labels used} \\ \cline{4-5}
     &  &  & Training (fine-tuning) & Test & & \\
    \midrule

    % BMR
    \multirow{2}{*}[0mm]{BMR \cite{wang2020unsupervised}} & \multirow{2}{*}[0mm]{\makecell[l]{WSJ si284,\\LS 960 hrs}} & ASR & WSJ si284 & WSJ eval92 & - & 81 hrs \\ \cline{3-7}
    & & PR & WSJ si84/si284 & WSJ dev93 & - & 15/81 hrs \\ \hline

    % vq-APC
    \multirow{2}{*}[0mm]{vq-APC \cite{chung20e_interspeech}} & \multirow{2}{*}[0mm]{LS 360 hrs} & PC & WSJ si284\textsuperscript{\ref{footnote:91-split-dev}} & \makecell[l]{WSJ dev93} & \checkmark & \makecell[l]{81 hrs} \\ \cline{3-7}
    & & SID & WSJ si284\textsuperscript{\ref{footnote:APC-split}} & WSJ si284\textsuperscript{\ref{footnote:APC-split}} & \checkmark & 65\textsuperscript{\ref{footnote:low-resource}} hrs \\ \hline

    % vq-wav2vec + Discrete BERT
    \makecell[l]{vq-wav2vec + \\\:DiscreteBERT \cite{baevski2019effectiveness}} & LS 960 hrs & ASR & LS 100 hrs & \makecell[l]{LS test-clean,\\ LS test-other} & - & \makecell[l]{10 mins\textsuperscript{\ref{footnote:low-resource}},\\ 1\textsuperscript{\ref{footnote:low-resource}}/10\textsuperscript{\ref{footnote:low-resource}}/100 hrs} \\ \hline

    % speech-XLNet
    \multirow{2}{*}[-1mm]{speech-XLNet \cite{song20d_interspeech}} & \multirow{2}{*}[-1mm]{\makecell[l]{LS 960 hrs\\\;\;\;+ WSJ si284\\\;\;\;+ TED2}} & PR & TIMIT\textsuperscript{\ref{footnote:official-split}} & TIMIT\textsuperscript{\ref{footnote:official-split}} & \checkmark & 4 hrs\vspace{1mm} \\ \cline{3-7}
    & & ASR & WSJ si284 & WSJ eval92 & - & \makecell[l]{7\textsuperscript{\ref{footnote:low-resource}}/14\textsuperscript{\ref{footnote:low-resource}}/30\textsuperscript{\ref{footnote:low-resource}}/81 hrs}\vspace{1mm} \\ \hline
    
    % MPC
    \multirow{3}{*}[0mm]{MPC \cite{jiang2019improving, jiang2021further}} & \multirow{2}{*}[0mm]{\makecell[l]{Didi Callcenter,\\Didi Dictation,\\Open Mandarin}} & ASR-zh & HKUST\textsuperscript{\ref{footnote:official-split}} & HKUST\textsuperscript{\ref{footnote:official-split}} & \checkmark & 168 hrs \\ \cline{3-7}
    & & ASR-zh & AISHELL-1\textsuperscript{\ref{footnote:official-split}} & AISHELL-1\textsuperscript{\ref{footnote:official-split}} & \checkmark & 178 hrs\vspace{1mm} \\ \cline{2-7}
    & \multirow{1}{*}[-0.5mm]{\makecell[l]{SWB, Fisher 1k,\\LS 960 hrs}} & ASR & SWB & Hub5'00 & - & 260 hrs \vspace{2.5mm} \\ \hline

    % MPE
    \multirow{2}{*}[0mm]{MPE \cite{liu2020masked}} & \multirow{2}{*}[0mm]{\makecell[l]{WSJ si284,\\LS 960 hrs}} & ASR & WSJ si284 & WSJ eval92 & - & \makecell[l]{25\textsuperscript{\ref{footnote:low-resource}}/40\textsuperscript{\ref{footnote:low-resource}}/81 hrs} \\ \cline{3-7}
    & & ASR & LS 100/360/960 hrs & LS test-clean & - & 100/360/960 hrs \\ \hline

    % ConvDMM
    ConvDMM \cite{khurana20_interspeech} & \makecell[l]{LS 50\cite{ravanelli2018learning}/360/\\\:\:\:960 hrs} & PC/PR & WSJ si284 & WSJ eval92 & \checkmark & \makecell[l]{5\textsuperscript{\ref{footnote:low-resource}}/50\textsuperscript{\ref{footnote:low-resource}}/100\textsuperscript{\ref{footnote:low-resource}} mins,\\ 4\textsuperscript{\ref{footnote:low-resource}}/8\textsuperscript{\ref{footnote:low-resource}}/40\textsuperscript{\ref{footnote:low-resource}} hrs} \\ \hline
    % wav2vec 2.0
    \multirow{2}{*}{wav2vec 2.0 \cite{baevski2020wav2vec}} & \multirow{2}{*}{\makecell[l]{LS 960 hrs,\\ LL 60k hrs}} & ASR & LS 960 hrs & \makecell[l]{LS test-clean,\\ LS test-other} & - & \makecell[l]{10 mins\textsuperscript{\ref{footnote:low-resource}},\\ 1\textsuperscript{\ref{footnote:low-resource}}/10\textsuperscript{\ref{footnote:low-resource}}/100/960 hrs} \\ \cline{3-7}
    & & PR & TIMIT\textsuperscript{\ref{footnote:official-split}} & TIMIT\textsuperscript{\ref{footnote:official-split}} & \checkmark & 4 hrs \\ \hline
    % NPC
    \multirow{2}{*}[0mm]{NPC \cite{liu21l_interspeech}} & \multirow{2}{*}[0mm]{LS 360 hrs} & PC & WSJ si284\textsuperscript{\ref{footnote:91-split-dev}} & \makecell[l]{WSJ dev93} & \checkmark & \makecell[l]{81 hrs} \\ \cline{3-7}
    & & SID & WSJ si284\textsuperscript{\ref{footnote:APC-split}} & WSJ si284\textsuperscript{\ref{footnote:APC-split}} & \checkmark & 65\textsuperscript{\ref{footnote:low-resource}} hrs \\ \hline
    % DeCoAR 2.0
    DeCoAR 2.0 \cite{ling2020decoar} & LS 960 hrs & ASR & \makecell[l]{LS 100 hrs} & \makecell[l]{LS test-clean,\\ LS test-other} & - & \makecell[l]{1\textsuperscript{\ref{footnote:low-resource}}/10\textsuperscript{\ref{footnote:low-resource}}/100 hrs} \\ \hline

    % TERA
    \multirow{4}{*}[0mm]{TERA \cite{liu2021tera}} & \multirow{4}{*}[0mm]{\makecell[l]{LS 100/360/\\\:\:\:960 hrs}} & PC & LS 100 hrs\textsuperscript{\ref{footnote:CPC-split}} & LS 100 hrs\textsuperscript{\ref{footnote:CPC-split}} & - & 80\textsuperscript{\ref{footnote:low-resource}} hrs \\ \cline{3-7}
    & & SID & LS 100 hrs\textsuperscript{\ref{footnote:CPC-split}} & LS 100 hrs\textsuperscript{\ref{footnote:CPC-split}} & - & 80\textsuperscript{\ref{footnote:low-resource}} hrs \\ \cline{3-7}
    & & PR & TIMIT\textsuperscript{\ref{footnote:official-split}} & TIMIT\textsuperscript{\ref{footnote:official-split}} & \checkmark & 4 hrs \\ \cline{3-7}
    & & ASR & LS 100 hrs & LS test-clean & - & 100 hrs \\ \hline
    % HuBERT
    HuBERT \cite{hsu2021hubert} & \makecell[l]{LS 960 hrs,\\ LL 60k hrs} & ASR & LS 960 hrs & \makecell[l]{LS test-clean\\ LS test-other} & - & \makecell[l]{10 mins\textsuperscript{\ref{footnote:low-resource}},\\ 1\textsuperscript{\ref{footnote:low-resource}}/10\textsuperscript{\ref{footnote:low-resource}}/100/960 hrs} \\ \hline
    %wav2vec-c
    wav2vec-c \cite{sadhu21_interspeech} & Alexa-10k & ASR & \makecell[l]{Alexa-eval} & \makecell[l]{Alexa-eval} & \checkmark & 1k hrs \\ \hline
    UniSpeech-SAT \cite{chen2021unispeechsat} & \makecell[l]{LL 60k hrs\\\;\;\;+ GigaSpeech-10k\\\;\;\;+ VP-24k} & Multi & SUPERB & SUPERB & \checkmark & See SUPERB \cite{yang21c_interspeech} paper for details \\ \hline
    WavLM \cite{chen2021wavlm} & \makecell[l]{LL 60k hrs\\\;\;\;+ GigaSpeech-10k\\\;\;\;+ VP-24k} & Multi & SUPERB & SUPERB & \checkmark & See \cite{yang21c_interspeech} for details \\ \hline
    \multirow{4}{*}[0mm]{XLS-R \cite{babu2021xlsr}} & \multirow{4}{*}[0mm]{\makecell[l]{VP-400k + MLS\\\;\;\;+ CV-dataset-7k + VL\\\;\;\;+ BBL}} & ASR & VP, MLS, CV-dataset, BBL, LS & VP, MLS, CV-dataset, BBL, LS & - & \multirow{4}{*}[0mm]{See \cite{babu2021xlsr} for details} \\ \cline{3-6}
    & & SID & VoxCeleb1 & VoxCeleb1 & \checkmark & \\ \cline{3-6}
    & & ST & CoVoST-2 & CoVoST-2 & \checkmark & \\ \cline{3-6}
    & & LID & VL & VL & - & \\

    % more & & & & & & \\ \hline
    % https://tex.stackexchange.com/questions/66596/vertical-alignment-in-multirow-using-cells-with-1-lines
    % https://texblog.org/2019/06/03/control-the-width-of-table-columns-tabular-in-latex/
    %\multirow{2}{*}{wav2vec 2.0} & \multirow{2}{*}{wav2vec 2.0} & Population Population Population Population & 2 & 3 & 4 \\\cline{3-6}
    %http://www.emerson.emory.edu/services/latex/latex_119.html
    %& &2016&2017& 5 & 6 \\ \hline
    \bottomrule
  \end{tabular}}
\end{table*}

\addtocounter{footnote}{1}
\footnotetext{\label{footnote:CPC-split}
Train/test split made available
by \cite{oord2018representation} on Google drive
https://drive.google.com/drive/folders/1BhJ2umKH3whguxMwifaKtSra0TgAbtfb.}
\addtocounter{footnote}{1}
\footnotetext{\label{footnote:official-split}
Utilizes official training or test split.}

\addtocounter{footnote}{1}
\footnotetext{\label{footnote:interface-split}
English utterances used in experiments. The utterances correspond to
approximately 3~hours for training, 40~minutes for development, and 30~minutes
for testing.}

\addtocounter{footnote}{1}
\footnotetext{\label{footnote:AED-datasets}
The 6 AED datasets used in
\cite{tagliasacchi2020pre} are MUSAN~\cite{snyder2015musan}, Bird Audio
Detection~\cite{stowell2019automatic}, Speech Commands~\cite{warden2018speech},
Spoken Language Identification~\cite{oponowiczspoken}, TUT Urban Acoustic
Scenes 2018~\cite{dcase2018} plus an SID task built with LS
\textit{train-clean-100}. In addition to the 6 datasets,
\cite{quitry2019learning} use the NSynth dataset~\cite{engel2017neural} for
evaluation.}
\addtocounter{footnote}{1}
\footnotetext{\label{footnote:CPC-8k}
A collection of AudioSet, AVSpeech,
CV-dataset, LS, WSJ, TIMIT, Speech Accent Archive 
(SSA)~\cite{weinberger2009towards}, TED3, and SWB. SSA is a growing annotated corpus of
English speech with various accents. Among the papers studied in this review, SSA
is used in \cite{kawakami2020learning} only for pre-training, and only 1 hour
of audio is utilized. Thus, we exclude it from our discussion in
\cref{section:benchmark}.}
\addtocounter{footnote}{1}
\footnotetext{\label{footnote:low-resource}
A subset of the official training
split is sampled, usually to mimic low-resource learning conditions or to
quickly evaluate for training and testing on the same split but disjoint
subsets.}
\addtocounter{footnote}{1}
\footnotetext{\label{footnote:APC-split}
Dataset split into training,
validation, and test subsets at a ratio of 8:1:1.}
\addtocounter{footnote}{1}
\footnotetext{\label{footnote:91-split-dev}
Dataset split into training
and validation subsets at a ratio of 9:1.}
\addtocounter{footnote}{1}
\footnotetext{\label{footnote:91-split-test}
The dataset split into training
and test subsets at a ratio of 9:1.}

\subsection{Experiment settings for evaluating SSL techniques}
A common way to benchmark SSL techniques and show their efficacy is to
fine-tune a pre-trained SSL model for a supervised downstream task.
% taking a network pre-trained with SSL as initialization, and then fine-tuning and evaluating the network with labeled data in downstream tasks.
Depending on the corpora used in pre-training and fine-tuning, techniques can
be benchmarked in terms of their capability to transfer knowledge across datasets
(i.e., using pre-training corpora that differ from the fine-tuning ones), their
benefit when training with limited labeled examples (i.e., sampling a subset of
labeled examples for fine-tuning), or their improvement over a fully supervised
baseline (i.e., using the entire training split of downstream datasets for
fine-tuning). Tables~\ref{table:method_data_setting} 
and~\ref{table:method_data_setting_2} summarize experiment settings used in the
SSL literature, including the pre-training corpora, downstream tasks and datasets,
and the amount of fine-tuning labels used, which indicates the targeted
benchmarking scenario as discussed above. Note that there are a variety of ways
to fine-tune pre-trained networks (e.g., fine-tune the entire network,
freeze certain layers during fine-tuning, and add various architectures of
prediction layers to pre-trained networks). We here omit
descriptions of these choices; readers can consult the
original publications for details.

As observed in Tables~\ref{table:method_data_setting} 
and~\ref{table:method_data_setting_2}, LS and WSJ are the most commonly used
pre-training corpora. At the same time, we observe a
growing industry investment in pre-training with larger datasets, e.g.,
CPC-8k (8k hours) for Bidir-CPC~\cite{kawakami2020learning}, LL (60k hours) for
CPC modified~\cite{riviere2020unsupervised}, 
wav2vec 2.0~\cite{baevski2020wav2vec}, and HuBERT~\cite{hsu2021hubert}, Alexa internal
datasets (10k hours) for wav2vec-c~\cite{sadhu21_interspeech}, Didi internal
datasets (10k hours) for MPC~\cite{jiang2019improving, jiang2021further}, the
combination of Gigaspeech, VP-24k, and LL (94k hours in total) for
UniSpeech-SAT~\cite{chen2021unispeechsat} and WavLM~\cite{chen2021wavlm}, and
the combination of VP-400K, MLS, CV-dataset, VL and BBL (436k hours in total)
for XLS-R~\cite{babu2021xlsr}. We expect this trend to continue with the growth
in available computing power. Most studies focuses on learning representations
for English, whereas Chinese~\cite{jiang2019improving, jiang2021further} and
multilinguality~\cite{kawakami2020learning, babu2021xlsr} are also gaining
attention. Compared to pre-training, datasets used for fine-tuning are more
diverse and cover downstream tasks as varied as ASR, PR, PC, SID, AED,
Sentiment, ST, and LID. For benchmarking training scenarios covering full
supervision as well as limited resources, the amount of labeled examples used for
fine-tuning also varies from several minutes up to 1k hours. Recent
benchmarks such as SUPERB~\cite{yang21c_interspeech} that consolidate multiple 
downstream tasks have gained attention for evaluating SSL
methodologies~\cite{chen2021unispeechsat, chen2021wavlm}. The goal of such benchmarks is
to provide a holistic evaluation of the performance of learned representations; 
we discuss these in detail in \cref{sec:benchmark}. With the increasing
popularity of SSL research, we expect future experiment settings to proliferate
and cover more languages, downstream tasks, and pre-training/fine-tuning
datasets.

\begin{figure*}
\centering
\includegraphics[width=0.9\textwidth]{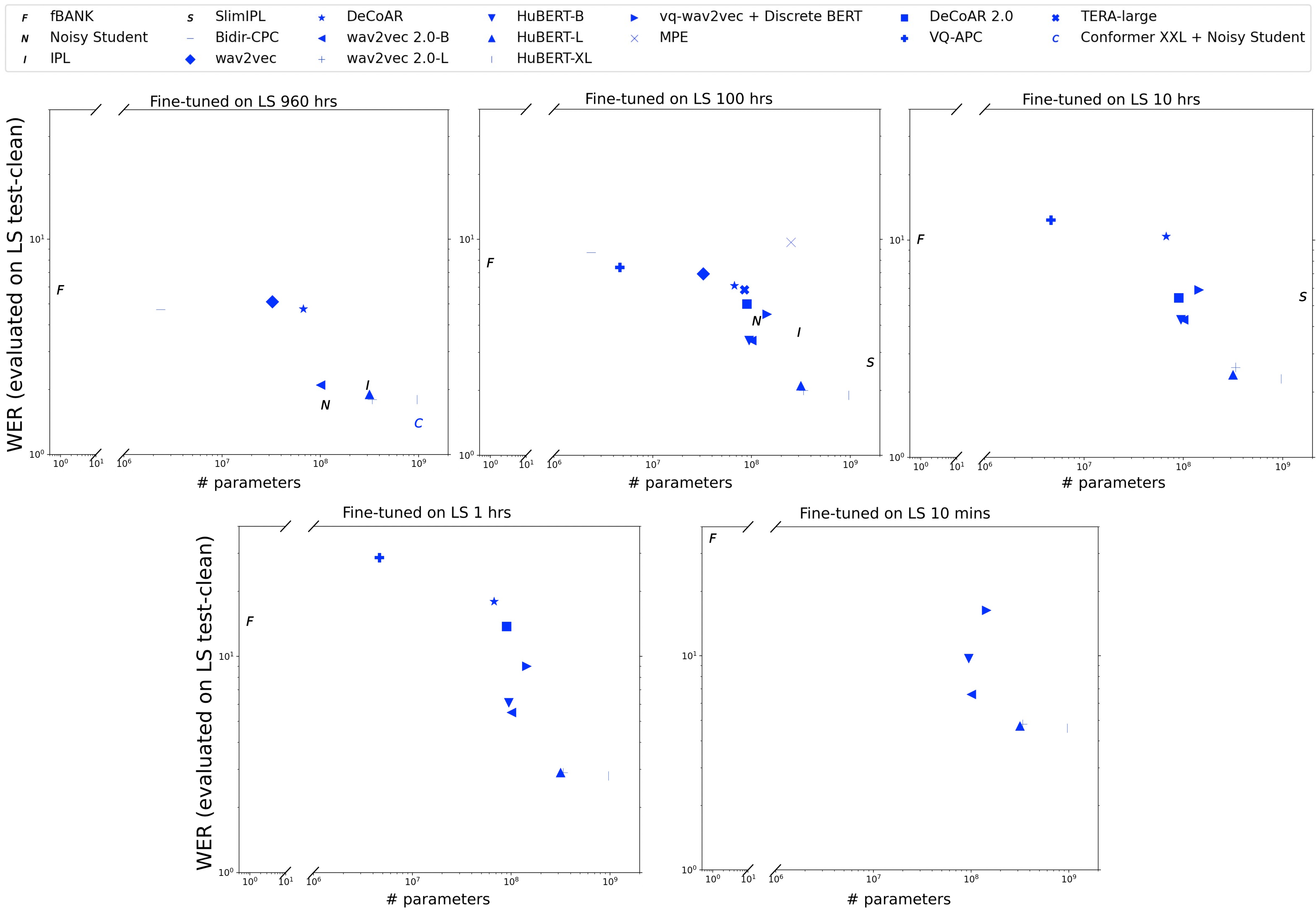}
\caption{SSL performance on ASR WER (vertical axis) evaluated with
LS \textit{test-clean} split. Techniques are sorted based on the number of
model parameters along the horizontal axis. Markers in blue correspond to models initialized with various SSL
techniques and then fine-tuned using 960, 100, 10, 1 hour(s), and 10 minutes
respectively. The 960-hour training set is the aggregation of
\textit{train-clean-100}, \textit{train-clean-360}, and
\textit{train-other-500} splits. The 100-, 10-, 1-hour, and 10-minute sets
leverage \textit{train-clean-100} or its sampling, except for Bidir-CPC, which
samples 10\% of the training examples from the entire 960-hour corpus. For simplicity,
several SSL techniques are appended with suffixes \textit{B}, \textit{L},
\textit{XL}, or \textit{XXL} indicating the \textit{Base}, \textit{Large},
\textit{X-Large}, or \textit{XX-Large} variants specified in the original
publication. We also compare with baselines including the log mel filterbank (fBANK) and semi-supervised, self-training
approaches (iterative pseudo labeling (IPL)~\cite{xu2020iterative}, 
slimIPL~\cite{likhomanenko2020slimipl}, noisy student~\cite{park2020improved}). These approaches are visualized in black. Also, note that the current state of the art---conformer XXL + noisy student~\cite{zhang2020pushing}---is a combination of self-training and SSL techniques. Given the
diversity of the listed methods in experiment settings (e.g., pre-training corpora
and objectives, whether a language model is used in decoding, whether model
parameters are frozen in fine-tuning), readers should be careful that the superiority
of methods cannot be decided only based on lower WER numbers.}
\label{fig:asr_result_no_limit}
\end{figure*}

\subsection{Benchmark results and discussion} \label{sec:benchmark} 
Given the diversity of datasets and downstream tasks used to evaluate SSL
techniques in the literature, it is infeasible to discuss all 
experiment settings in this survey. Hence, due to their wide adoption for
experiments conducted by studies in both SSL and the speech community in
general, we focus first on ASR on the LS dataset to understand the
efficacy of SSL. We examine SSL techniques which report ASR results on the LS
\textit{test-clean} split, and summarize the published WER in
\cref{fig:asr_result_no_limit}. The ASR models were obtained first by using
unlabeled speech to pre-train a model with each SSL technique. The model was
then fine-tuned on labeled data by utilizing a supervised training objective.
Respectively, 960, 100, 10, 1 hour(s), and 10 minutes of labeled LS training data
were used for fine-tuning, as indicated in different panels of
\cref{fig:asr_result_no_limit} (see
the caption of \cref{fig:asr_result_no_limit} for more details).
Semi-supervised methods such as self-training, where a model is first trained
on labeled data to annotate unlabeled speech, and then subsequently trained on
combined golden and self-annotated label-speech pairs, are gaining popularity
in the speech community and have yielded competitive results. For comparison, we also
show performance from such methods (iterative pseudo labeling 
(IPL)~\cite{xu2020iterative}, slimIPL~\cite{likhomanenko2020slimipl}, noisy 
student~\cite{park2020improved}), as well as the current state of the art---conformer XXL + noisy
student~\cite{zhang2020pushing}---which augments SSL with various advanced
techniques including self-training. Furthermore, we illustrate in the figure
the performance of a baseline system \cite{yang21c_interspeech} based on log mel filterbank (fBANK), which is one of the most commonly used features designed by domain experts.
As observed in the figure, most SSL techniques outperform fBANK
features, and with the growing investment in model size, better performance is
achieved. The largest ones, such as wav2vec 2.0-L and HuBERT-L/XL, yield
% results competitive to the state of the art 
  competitive results                         % AMH: check
when the entire 960-hour of labeled data is used in
training/fine-tuning. The benefit of SSL, especially models with more parameters
like wav2vec 2.0 and HuBERT, becomes more evident when the labeling resources
become scarce. Compared to popular semi-supervised methods such as IPL,
slimIPL, and noisy student using 100 hours of labels, wav2vec~2.0 and HuBERT
achieve lower or competitive WERs with 1 hour or even 10 minutes of labeled
examples. The results are highly favorable for low-resource use cases, for instance when
expanding systems to new domains or languages for which large amounts of unlabeled
audio are available, since collecting labels for new conditions is often prohibitively
slow or costly.
% discuss the trend
%[Pick representative tasks and datasets and show the benefit of SSL, e.g., fewer fine-tuning/labeled data is required to achieve better/comparable performance] [talk about trend as well]  The plot suggests that ASR results have been increased significantly with the progress in SSL. Near SOTA performance can be achieved with much fewer training examples.  

In addition to the ASR task, where the current state of the art is achieved by a method
combining SSL pre-training and self-training 
techniques~\cite{zhang2020pushing}, SSL models 
% approach the state of the art 
  are competitive                % AMH: check
in other tasks, including IC,
SID, ASV, and QbE. We summarize the performance of these models and previous
non-SSL methods in \cref{table:sota_performance}. The results suggest that the
benefit of SSL is generalizable among tasks that require encoding 
information such as content, speaker, and semantics. As SSL research
gains more attention, we expect that SSL pre-trained models will 
achieve state-of-the-art results on an increasing number of tasks.

\begin{table}[ht]
  \centering
  \footnotesize
  \caption{Tasks where the state of the art is models with SSL pre-training.}
  \label{table:sota_performance}
  \renewcommand*\arraystretch{1.2}
  \begin{tabular}{llllll}  
    \toprule
    Tasks & Dataset & non-SSL & SSL \\
    \midrule
    ASR (WER $\downarrow$) & LS test-clean/other & 2.1/4.0 \cite{xu2020iterative} & 1.4/2.6 \cite{zhang2020pushing} \\ \hline
    IC (Acc $\uparrow$) & FSC & 98.8 \cite{lugosch19_interspeech} & 99.3\cite{chen2021unispeechsat} \\ \hline
    SID (Acc $\uparrow$) & VoxCeleb1 & 94.8 \cite{hajibabaei2018unified} & 95.5 \cite{chen2021wavlm} \\ \hline
    ASV (EER $\downarrow$) & VoxCeleb1 & 3.1 \cite{hajavi2021siamese} & 2.4 \cite{wang2021fine} \\ \hline
    QbE (MTWV $\uparrow$) & QUESST (EN) & 10.6 \cite{rodriguez2014gtts} & 11.2\cite{chen2021unispeechsat} \\

    %wav2vec-c
    %wav2vec-c \cite{sadhu21_interspeech} & Alexa-10k & ASR & \makecell[l]{Alexa-eval} & \makecell[l]{Alexa-eval} & \checkmark & 1k hrs \\ \hline
    %UniSpeech-SAT \cite{chen2021unispeechsat} & \makecell[l]{LL 60k hrs\\\;\;\;+ GigaSpeech-10k\\\;\;\;+ VP-24k} & Multi & SUPERB & SUPERB & \checkmark & Check SUPERB \cite{yang21c_interspeech} paper for details \\ \hline
    %WavLM \cite{chen2021wavlm} & \makecell[l]{LL 60k hrs\\\;\;\;+ GigaSpeech-10k\\\;\;\;+ VP-24k} & Multi & SUPERB & SUPERB & \checkmark & Check \cite{yang21c_interspeech} for details \\ \hline
    %\multirow{4}{*}[0mm]{XLS-R \cite{babu2021xlsr}} & \multirow{4}{*}[0mm]{\makecell[l]{VP-400k + MLS\\\;\;\;+ CV-dataset-7k + VL\\\;\;\;+ BBL}} & ASR & VP, MLS, CV-dataset, BBL, LS & VP, MLS, CV-dataset, BBL, LS & - & \multirow{4}{*}[0mm]{Check \cite{babu2021xlsr} for details} \\ \cline{3-6}
    %& & SID & VoxCeleb1 & VoxCeleb1 & \checkmark & \\ \cline{3-6}
    %& & ST & CoVoST-2 & CoVoST-2 & \checkmark & \\ \cline{3-6}
    %& & LID & VL & VL & - & \\ \hline

    \bottomrule
  \end{tabular}
\end{table}

Despite the obvious trend of increasing performance as more parameters and SSL
pre-training data are being used, numbers in \cref{fig:asr_result_no_limit} 
and \cref{table:sota_performance} are less comparable than might be expected.
The task performance is obtained from the original papers and is often
achieved with different downstream fine-tuning recipes, including various
language models (used in the ASR system), prediction heads (networks added to
SSL for downstream inference), or choices between fine-tuning the whole
networks or freezing the SSL encoders. For example, in the ASR task, HuBERT-L
and wav2vec~2.0-L leverage Transformer as their language model, while a 4-gram
language model trained on LS is used in DeCoAR~2.0. The lack of common and
established mechanisms to evaluate SSL techniques in downstream applications
makes it difficult to compare techniques fairly and understand their
capabilities. To address this challenge, there are increasing efforts to establish
benchmarks with shared downstream tasks, datasets, and downstream recipes. Such
efforts include SUPERB~\cite{yang21c_interspeech}, 
LeBenchmark~\cite{evain21_interspeech}, ZeroSpeech~\cite{dunbar2020zero},
HEAR \cite{pmlr-v176-turian22a},
NOSS~\cite{shor20_interspeech}, and HARES~\cite{wang2021towards}. 

SUPERB~\cite{yang21c_interspeech} is a benchmarking platform that allows the
SSL community to train, evaluate, and compare speech representations on
diverse downstream speech processing tasks, from acoustic and speaker identity
to paralinguistic and semantics. SUPERB consolidates downstream recipes to
focus on common and straightforward settings (e.g., prediction head
architectures, language models, hyperparameter spaces) to facilitate generalizable
and reproducible benchmarking of SSL techniques. SUPERB also encourages
researchers to innovate for efficient use of model parameters and computation
resources 
to democratize SSL beyond race among Big Tech.   % AMH: what does this mean?
LeBenchmark~\cite{evain21_interspeech} shares a vision similar to SUPERB and provides a
reproducible framework for assessing SSL in French with ASR, spoken language
understanding, speech translation, and emotion recognition. 
ZeroSpeech~\cite{dunbar2020zero} (described in more detail in \cref{zero_speech})
challenges the scientific community to build speech and language understanding
systems using zero expert resources 
%with the ultimate goal of bringing the systems to the services 
for millions of users of ``low-resource" languages.
%or in abnormal condition (dyslexia, autism, etc). ZeroSpeech offers evaluation for independent units in the system as well aggregation of multiple components. 
SSL techniques are also benchmarked with the ZeroSpeech 
challenge~\cite{tjandra20_interspeech, niekerk20b_interspeech}. Apart from the speech
community, researchers have also established HEAR (holistic evaluation of audio
representations) \cite{pmlr-v176-turian22a}, NOSS (non-semantic
speech benchmark)~\cite{shor20_interspeech}, and HARES (holistic audio
representation evaluation suite)~\cite{wang2021towards} to benchmark audio
representations. These efforts promote the creation of an audio embedding
that is as holistic as the human ear in interpreting speech, environmental
sound, and music. Given the significant need to understand and compare SSL techniques
fairly and comprehensively, we expect SSL benchmarking to remain an
active research area.

%SUPERB motivation: consolidating evaluation recipes to understand the capability of various SSL in a fair way. SUPERB designates the architectures of prediction heads for each downstream tasks [common architectures] [frozen] [generalizable, comparable, reproducible results]. [Results on tasks with such constraint conditions (mainly SUPERB but also from literatures imposing similar constraint).] [discussion on the results - competitive performance with simple downstream recipes] [More challenges, ZeroSpeech, HEAR ...]

%Librispeech 100h-960h \\
%Libri-light 10m, 1h, 10h - 60k \\
%SUPERB \\
%Add more
%Hear 2021 Challenge? (focus on audio, not only speech)

 %Reviewer: SW, Hung-yi Lee, Abdo
\section{Analysis of Self-Supervised Representations}
\label{analysis}

% {\color{blue} Katrin}\\
The previous sections have shown how self-supervised learning can result in
powerful representations that provide a robust starting point for several
downstream tasks. It is natural to ask if we can gain an even deeper
understanding of the nature of these representations, in order to further
optimize them or apply them to different problems.
What is the information encoded in these representations? How robust are they
to distributional shifts, and how dependent are they on the size of the
training data? Do they generalize across languages? What are the key
ingredients for training powerful representations: input data, network
architecture, training criterion, or all three? Can we predict their
performance on downstream tasks from their training behavior? This section
tries to answer these questions by summarizing several studies that analyze
self-supervised representations.

\subsection{Information content}

In \cite{pasad2021} wav2vec~2.0 representations were analyzed with
respect to their acoustic-linguistic information content at different
network layers. Three different mechanisms were used for this
purpose. The first of these is canonical correlation analysis (CCA),
which computes similarity scores between two continuous vectors
based on the maximum correlation of their linear projections.  These
can be used to judge the similarity of embeddings at different layers
with each other, with standard acoustic representations such as mel
filterbank features, or word embeddings derived from text. The second
method clusters continuous representation vectors and computes the
discrete mutual information between cluster IDs and phone or word
labels. The third method involves probing tasks: representation
vectors extracted from the network are used to perform simple
downstream tasks, in particular determining whether two acoustic segments
correspond to the same word, and a standard benchmark of 11 word
similarity tasks~\cite{faruqi14}. These are mostly used to gauge the
amount of lexical information present in the embeddings.  Using this
battery of tests the authors compared pre-trained models of varying
sizes as well as models fine-tuned for ASR. They found that pre-trained
models show an autoencoder-style behavior, with early layers showing
strong similarity with input features, intermediate layers diverging
more, and final layers reverting back to higher similarity with input
features and early layers. Generally, the earlier layers in wav2vec~2.0 
models encode acoustic information. The next set of layers encodes
phonetic class information, followed by word meaning information,
before reverting back to encoding phonetic/acoustic information. Thus,
extracting representations from the last layers for tasks that require
phonetic or word-related information may not be the best
strategy. Indeed, the authors of \cite{baevski2021unsupervised} show that a
phone classifier trained on each of the 24 frozen layers of a wav2vec~2.0 model
showed the lowest phone error rates
for layers 10--21 and higher error rates for the other layers. 
\cite{pasad2021} further show that fine-tuning the pre-trained model with a
character-level CTC training criterion changes the behavior of the
last layers (especially the final two layers), breaking the
autoencoder-style behavior and focusing the information encoded in the
last layers on orthographic-phonetic and word information. 

The peaking of class-relevant information in intermediate layers seems to be
common across different self-supervised learners and different modalities. In
an analysis of text-based Transformers trained with a masked language model
criterion~\cite{voita2019} observed a similar compression plus reconstruction
pattern. Interestingly, similar network behavior was also recently described
for self-supervised learners in computer vision: using a contrastive
self-supervised learner (SimCLR) that optimizes for augmentation invariance,
\cite{grigg2021self} show that it is the intermediate representations that most
closely approximate information learned in a supervised way, i.e., they
provide more class information than the representations from final layers. This
is similar to the findings described above for wav2vec~2.0 without fine-tuning,
where intermediate layers provide more information about phone and word
classes.

Self-supervised representations may encode other information besides phonetic
classes or words, for example, channel, language, speaker, and sentiment
information. 
\edit{It is shown that the per-utterance mean of CPC features captures speaker information to a large extent\cite{van2021analyzing}.}
Location of information pertaining to speakers vs.\ language classes was
analyzed in \cite{ling20odyssey} for a 12-layer BERTphone model. This model
combines a self-supervised masked reconstruction loss with a phone-based CTC
loss to produce representations	for speaker recognition and language
identification. By	analyzing the weights	of a linear combination	of layer
representations for these two downstream tasks, it was	shown that language
recognition draws on representation	from higher layers (peaking at layer~10)
whereas speaker recognition benefited from layers at positions 6, 9, and 12.
This may indicate that language recognition relies more on higher-level
phonetic information whereas speaker recognition uses a combination of
acoustic and phonetic information. In a recent study~\cite{chen2021wavlm} the
same technique was used to identify layer contributions for the downstream
SUPERB benchmark tasks in the WavLM model. For a smaller model (95M parameters)
it was again confirmed that lower layers encode speaker-related information
necessary for speaker diarization and verification whereas higher layers encode
phonetic and semantic information. Another study~\cite{wang2021layersup} used
explicit self-supervised loss at the intermediate layers rather than just the
output layer of a HuBERT model in order to enforce better learning of phonetic
information. The resulting model was indeed better at downstream tasks
requiring information about phonetic content, such as phone recognition, ASR,
and keyword spotting, but worse at speaker-related tasks like speaker
diarization and verification. 

%Hung-yi Lee: My group also has a paper analyzing the attention of self-supervised speech model (https://arxiv.org/abs/2006.03265). I will include some findings in this subsection later.
Most self-supervised learning approaches rely on a Transformer architecture for
the representation model. In \cite{AnalyzeAttention} the attention patterns in
generatively trained Transformer representation models were analyzed.
Self-attention heads were grouped into three categories: diagonal, vertical,
and global. It was found that the diagonal head focuses on neighbors and is
highly correlated with phoneme boundaries, whereas the vertical head focuses on
specific phonemes in the utterance. Global heads were found to be redundant as
removing them resulted in faster inference time and higher performance.

\subsection{Training criterion}
In \cite{chung2021}, representations based on different training criteria
(masked predictive coding, contrastive predictive coding, and autoregressive
predictive coding) were compared  and analyzed with respect to the correlation
between their training loss and performance on both phone discrimination and
speaker classification probing tasks. It was observed that the autoregressive
predictive coding loss showed the strongest correlation with downstream
performance on both tasks; however, models were not further analyzed
internally. An evaluation of the similarity of representations trained
according to the three criteria above (but with different architectures and
directionality of contextual information) also showed that it is the training
criterion that most influences the information encoded in the representations,
not the architecture of the learner or the directionality of the input. 

A similar insight was obtained in \cite{zhou2020}, which compared vq-vae and
vq-wav2vec with respect to their ability to discover phonetic units.
The vq-vae model extracts continuous features from the audio signal; a
quantizer then  maps them into a discrete space, and a decoder is trained to
reconstruct the original audio conditioned on the latent discrete
representation and the past acoustic observations. By contrast, vq-wav2vec
predicts future latent discrete representations based on contextualized
embeddings of past discrete representations, in a CPC-style way. The models
were evaluated according to their ability to discover phonetic units (as
measured by phone recognition error rate on TIMIT, and the ZeroSpeech ABX task
(see \cref{sec:zero} for more details)), and it was found that the predictive vq-wav2vec
model fared better than the autoencoder-like vq-vae model, most likely due to
its superior ability to model temporal dynamics.

\subsection{Effects of data and model size}
\label{subsec:modelsize}

How does the performance of self-supervised models change in relation to the
amount of training data, and in relation to the size (number of parameters) of
the model?
Several studies have demonstrated better downstream performance when using
larger datasets~\cite{rivi20, kawakami2020learning, chen2021wavlm}. For
example,
\cite{kawakami2020learning} compared representations learned by a bidirectional
CPC model from the standard 960 hour LS corpus and a corpus of 8,000 hours of
diverse speech from multiple sources.\textsuperscript{\ref{footnote:CPC-8k}}
Not surprisingly, an ASR model trained on top of these representations
performed better when representations were learned from the larger dataset. 
Although the precise relationship between data size and performance has not
been quantified, we can assume that it follows a law of diminishing returns (or
power law), 
similar to observations for most data-intensive machine learning tasks.
In addition to the size of the dataset, the diversity of the data also seems
to play a role, although this was not quantified in this study. However, recent
experiments with larger and more diverse data collections~\cite{chen2021wavlm}
confirm this assumption, as do 
explicit investigations of domain shift robustness (see
\cref{subsec:robustness} below). 

\edit{The relation between model sizes and downstream performances have also been investigated \cite{pue2021scaling,versteegh2015zero}.}
Using the Mockingjay 
model~\cite{liu2020mockingjay}, the authors in \cite{pue2021scaling} attempt to establish a relationship between
model size and self-supervised \lone{} loss and demonstrate that it approximately
follows a power law. Model size and accuracy on downstream phone
classification and speaker recognition tasks are positively correlated but do not
exactly follow a power law; rather, the accuracy saturates as models increase in
size, possibly due to the lack of a corresponding expansion in training data
size.

\subsection{Robustness and transferability}
\label{subsec:robustness}

It is well known that traditional speech features like MFCCs lack
robustness against environmental effects such as additive noise,
reverberation, accents, etc., that cause differences in the distributions of
speech features.
Do pre-trained representations offer greater robustness against distributional
shifts? 
One study~\cite{kawakami2020learning} compared pre-trained
representations from a CPC model against MFCCs and 
found pre-trained representations to be more robust to mismatches between
training and test data. The 
training data consisted of clean, read speech (LS)
whereas test data consisted of the Switchboard corpus and TED talks. The
distributional shifts here may stem from both the acoustics (microphone, room
reverberation) as well as
lexical effects related to topic and style, as well as differences in speaker
characteristics such as accent. 
Similar problems were also investigated using HuBERT and wav2vec~2.0 models in
\cite{chang2021exploration}.
In \cite{robustw2v2} domain effects were studied in greater detail using 
datasets from six different domains. In particular, the authors focused on the
usefulness of adding out-of-domain data to pre-training. The general
conclusions are that pre-training on more and diverse domains is preferable:
models pre-trained on more domains performed better than those pre-trained on
fewer when tested on held-out domains, regardless of which additional
labeled data was used for fine-tuning. Adding in-domain unlabeled data---if
available---to pre-training improves performance robustly; however, even
out-of-domain unlabeled data is helpful and closes  66--73\% of the performance
gap between the ideal setting of in-domain labeled data and a competitive
supervised out-of-domain model. 
 
In \cite{rivi20} the effectiveness of CPC-trained representations for
phone discrimination tasks was compared across several languages. It
was found that representations pre-trained only on English successfully
enabled phone discrimination in 10 other languages, rivaling
supervised methods in accuracy in low-data regimes (1h of labeled data
per language). Thus, self-supervised pre-training enables the model to
learn contextualized speech features that generalize across different
languages. In \cite{conneau20xling}, a wav2vec~2.0 model was trained on data
from multiple different languages and different corpora (Babel, Common Voice,
and multilingual LS)
jointly, followed by fine-tuning for each individual language. The largest
model covers 53 languages in total
and consists of 56,000 hours of speech. Compared to monolingual pre-training,
even smaller models trained on only ten languages improve performance
substantially on a downstream character-based ASR task. Low-resource languages
with little labeled data improve the most under this training regime.
Multilingual representations also resulted in competitive performance (lower
character error rate than monolingual representations) for languages not
present in the training dataset, again showing that unsupervised pre-trained
representations can learn generic features of the speech signal that generalize
across different languages. The study also found that sharing data from closely
related languages is more beneficial than combining distant languages. An
analysis of language clusters in the shared discrete latent representation
space revealed that similar languages do indeed show a higher degree of sharing
of discrete tokens. 
Finally, one might ask whether the interpretation of representations extracted
from different layers of a self-supervised models also generalizes to the
multilingual setting. Experiments in \cite{baevski2021unsupervised} on phone
recognition in eight languages based on the different layers of the
multilingual wav2vec~2.0 XLSR-53 model indicate that this is indeed the case:
phone error rates showed the same pattern as in the monolingual (English)
scenario, with lower phone error rates for middle layers as opposed to
earlier/later layers.

 %newly added - just analysis, keeping 04 for benchmarking
\section{From Representation Learning to Zero Resources}
\label{sec:zero}

% {\color{black} Reviewer: TNS, Abdo, Hung-yi Lee }\\
% \kl{We discussed in our last meeting that I would add a pointer in this section to the acoustic word embedding section.  However, I'm not quite sure what the right spot for it is, so waiting for this section to stabilize a bit more first.}

In the SSL framework, speech representations can be
learned and used in various downstream tasks to achieve competitive, robust,
and transferable performance, as shown in
\crefrange{section:benchmark}{analysis}. 
However, labeled data is still required. 
For example, in ASR, utterances and their manual transcriptions are needed to learn downstream models or fine-tune representation models. 
Can a model learn without any labeled data? 
In \cref{secsec:unpaired}, we show how to learn ASR models without any paired audio and text and how SSL improves the framework.
In addition, many languages have no writing system. 
In \cref{subsec:zero}, the SSL representation is further used in scenarios where text data is unavailable.

\subsection{Unpaired text and audio} \label{secsec:unpaired}
% What are we going to do % how to match
%In the self-supervised learning framework, speech representations can be learned and used in the downstream tasks, but some labeled data for finetuning is needed.

\begin{table*}[ht]
\centering
    {\renewcommand*\arraystretch{1.4}}
\caption{
Unsupervised ASR.
TIMIT numbers are phoneme error rates (PER), while the numbers for
LibriSpeech are word error rates (WER).
SWC $=$ spoken word classifier, ST $=$ speech translation.
All speech and text are in English if not specified. 
\textcolor{black}{
The references in the table are sorted according to the date of publication.}
}
\label{fig:unsupervisedASR}
\resizebox{0.99\textwidth}{!}{
\begin{tabular}{cllllll}
    \toprule
Reference & Speech representation & Speech segmentation  & Text token/representation & Mapping approach & Refinement  & Results  \\ 
    \midrule
\cite{liu2018completely}& Audio word2vec~\cite{Wang2018SegAudioWord2Vec} & Oracle & Phoneme & \edit{Adversarial Training~\cite{gulrajani2017improved}} & - & TIMIT (PER): 63.6\%  \\ %Submitted on 1 Apr 2018
  \midrule
  \cite{chung2018unsupervised} & Speech2vec~\cite{chung2018speech2vec} & BES-GMM~\cite{Kamper17BESGMM} &  Word2Vec & \edit{Adversarial Training~\cite{conneau2017word}} & Self-training & SWC (Acc): 10.9\%  \\ %Submitted on 18 May 2018
%https://arxiv.org/pdf/1805.07467.pdf %The text embeddings were obtained by training Word2Vec on the transcriptions using the fastText implementation without subword information [3].
  \midrule
  
    \cite{chung2019towards} & \tabincell{l}{Speech2vec\\(English)} & Oracle & \tabincell{c}{Word2Vec\\(French)} & VecMap~\cite{artetxe-etal-2018-robust} & \tabincell{l}{LM rescore,\\sequence DAE} & ST (BLEU): 10.8\%   \\ %[Submitted on 4 Nov 2018]
\midrule
  
\cite{yeh2018unsupervised} & MFCC & GAS~\cite{wang2017gate} & Phoneme & Empirical-ODM~\cite{Liu2017ODM} & Self-training &  TIMIT (PER):  41.6\%  \\ %Submitted on 23 Dec 2018
  \midrule

\cite{chen2019completely} & MFCC & GAS  & Phoneme & \edit{Adversarial Training~\cite{gulrajani2017improved}} & Self-training &  TIMIT (PER):  33.1\%   \\
  \midrule
  
\cite{baevski2021unsupervised} & Wav2vec 2.0~\cite{baevski2020wav2vec} & \kmeans{} & Phoneme &\edit{Adversarial Training~\cite{gulrajani2017improved}} & Self-training & \tabincell{l}{TIMIT (PER):  18.6\%, \\ LibriSpeech (WER): 5.9\%}   \\
  \midrule
  
\textcolor{black}{\cite{Deciphering2022}} & \textcolor{black}{\tabincell{l}{Universal Phone \\ Recogniser}} & - & \textcolor{black}{Grapheme} & \textcolor{black}{Decipherment~\cite{ravi-knight-2011-deciphering}} & \textcolor{black}{Self-training} & \textcolor{black}{\tabincell{l}{GlobalPhone: 32.5\% to just 1.9\% \\ worse than supervised models} }  \\ %ubmitted on 12 Nov 2021 %m 32.5% to just 1.9% absolute worse than the equivalent fully supervised models trained on the same data
  \midrule
\textcolor{black}{\cite{w2v-u2}} & \textcolor{black}{Wav2vec 2.0~\cite{baevski2020wav2vec} }& - & \textcolor{black}{Phoneme} & \textcolor{black}{Adversarial Training~\cite{gulrajani2017improved}} & \textcolor{black}{Self-training} & \textcolor{black}{LibriSpeech (WER): 6.3\%}   \\
  \midrule

\textcolor{black}{\cite{w2v-u2}} & \textcolor{black}{Wav2vec 2.0~\cite{baevski2020wav2vec}} & - & \textcolor{black}{Grapheme} & \textcolor{black}{Adversarial Training~\cite{gulrajani2017improved}} & \textcolor{black}{Self-training} & \textcolor{black}{LJSpeech (WER): 64.0\%}  \\

%e MUSE [7] and VecMap [8], n
%The numbers in the section of unsupervised methods denoted as BLEU score (%) of VecMap /
%BLEU score (%) of MUSE 10.8 / 6.2
%11.3 / 7.3
%https://arxiv.org/pdf/1811.01307.pdf
    \bottomrule
\end{tabular}
}
\end{table*}

\subsubsection{Unsupervised ASR}
% {\color{black} Hung-yi}\\

If only unpaired speech and text are available, that is, the text is not a
manual transcription of speech, can the machine learn how to transcribe speech
into text?
This scenario is called \textit{unsupervised ASR}, and the framework is as
below. 
Given a set of unlabeled utterances $\mathcal{S}=\{S_1, S_2, ..., S_N\}$  and a
set of sentences $\mathcal{Y}=\{Y_1, Y_2, ..., Y_M\}$,\footnote{Note that
the speech and text are not paired, that is, $Y_i$ is not the transcription of
$S_i$.} a mapping function~$F$, which can take an utterance~$S$ as input and
generate its transcription, is learned from data. 
\Cref{fig:unsupervisedASR} summarizes recent work on unsupervised ASR,
including the speech representation used, the algorithm used to learn the
mapping without supervision, and the results. Below, we will discuss these
methods in more detail.

Adversarial training~\cite{goodfellow2014generative,arjovsky2017wasserstein, gulrajani2017improved} is one common way to learn such a 
mapping function. 
The framework includes a discriminator and a generator.
The mapping function~$F$ plays the role of the generator, which takes speech utterances as input and outputs text.
The discriminator learns to distinguish real text from the generated
output; the generator learns to ``fool'' the discriminator.
The generator and the discriminator are trained in an iterative, 
interleaved way. 
After the training, the generator serves as the speech recognition model.
\edit{
There is a large amount of work using gradient penalty in the objective of training discriminators~\cite{liu2018completely,chen2019completely,baevski2021unsupervised,w2v-u2}, which is inspired by Improved Wasserstein Generative Adversarial Network (WGAN)~\cite{gulrajani2017improved}.
}
\textcolor{black}{
Other ways to map speech and text include via segmental empirical output distribution matching (segmental empirical-ODM)~\cite{yeh2018unsupervised} and decipherment algorithm~\cite{Deciphering2022}.}

% Difficulty of unsupervised ASR (v.s. unsupervised MT)
Success in unsupervised neural machine translation
(MT)~\cite{artetxe2017unsupervised, conneau2017word, lample2017unsupervised}
has inspired innovative exploration of various unsupervised ASR algorithms.
If learning a translation model from unaligned sentences in two languages is
possible, considering speech and text as two different languages, learning the
mapping relationship from speech space to text space without an alignment 
% is not impossible.
  should likewise be possible.   % AMH: check
However, there are differences between unsupervised MT and unsupervised
ASR.
In unsupervised MT, most discrete source tokens can be mapped to specific
target tokens representing the same meaning. 
However, because speech has segmental structures, in unsupervised ASR, each text
token maps to a segment of consecutive acoustic features of variable length in
an utterance.
The generator is supposed to learn the segmental structure of an utterance
because information like token boundaries is not directly available.
This makes unsupervised ASR more challenging than unsupervised MT.

% how to represent audio
For unsupervised ASR to be feasible, the common idea is to make the speech and text units close to each other.
For the text side, word sequences can be transformed into phoneme sequences if a lexicon is available. 
On the other hand, we must first convert the speech signal into something close to phonemes. 
To achieve that, most studies on unsupervised ASR use a phoneme
segmentation module before the generator to segment utterances into
phoneme-level
segments~\cite{liu2018completely,chen2019completely,yeh2018unsupervised}. 
A representation vector or a token then represents each phoneme-level segment. 
It is easier for the generator to map each segment-level representation 
or token to the correct phoneme when the representation or token is highly correlated to the phonemes.
Wav2vec-U~\cite{baevski2021unsupervised} selects the input feature from different layers of wave2vec~2.0~\cite{baevski2020wav2vec}.
The selection criterion is based on analysis of the phonetic information in each layer.
\textcolor{black}{
If a universal phone recognizer trained from a diverse set of languages is available, it is another way to transcribe speech into phone-level tokens~\cite{Deciphering2022}.}
Another series of work is to transform a word into a word embedding. 
\cite{chung2018unsupervised,chung2019towards} use adversarial training to map the word-level
speech embedding space~\cite{chung2018speech2vec} to the word embedding space
and achieve promising performance on spoken word classification, speech
translation, and spoken word retrieval.
\Cref{fig:unsupervisedASR} summarizes the various ways to segment speech and represent speech and text in each reference.  

As shown in \cref{fig:unsupervisedASR}, most studies use
\textit{self-training} to refine the models.
% to achieve better performance.  % AMH: redundant
In self-training, the generator serves as the first-version phoneme recognition model.
Inputting unpaired speech to the generator generates the corresponding
``pseudo transcription''. 
We then view the speech utterances and their pseudo transcriptions as paired
data which we use to train a model in a supervised manner.
Although the pseudo transcriptions have more errors than oracle
transcriptions, experiments show that training models on pseudo
transcriptions still significantly boosts performance compared to the
first-version model.
%With the new model, we can further obtain new transcriptions.
%The iteration will continue until the performance converges. 

%Using performance to end this section.
Wav2vec-U~\cite{baevski2021unsupervised} achieved state-of-the-art results at the time, which suggests that representation learning is
essential for the success of unsupervised ASR.
It achieved an 11.3\% phoneme error rate on the TIMIT benchmark. 
On the LS benchmark, wav2vec-U achieved a 5.9\% \edit{WER} on
\emph{test-other}, rivaling some of the best published systems trained on 960 hours of labeled data from only two years earlier. 
\textcolor{black}{
And wav2vec-U 2.0~\cite{w2v-u2} further removes the requirement of the segmentation stage, so the unsupervised ASR model can be learned in an end-to-end style.}
The robustness of wav2vec-U was further analyzed with respect to 
domain-mismatch scenarios in which the domains of unpaired
speech and text were different~\cite{lin2021analyzing}.
Experimental results showed that domain mismatch leads to inferior performance, but a representation model pre-trained on the targeted speech domain extracts better representations and reduces this drop in performance.

%Hung-yi: The technique below is very important. But I think it is fine to ignore this part in a reviewe paper due to space limitation. 
\begin{comment}
In unsupervised learning, how to determine the hyperparameters is an issue. 
In a typical supervised framework, hyperparameter tuning is done by a
development set, which includes labeled data.
In an unsupervised setting, the existence of this kind of development is tricky. %Is tricky a correct term?
The existence of such kind of development set means the presence of the labeled
data.
Therefore, wav2vec-U~\cite{baevski2021unsupervised} develops a new approach to
determine the hyperparameters without a labeled development set. 
%should I provide more details? 
\end{comment}

%Hung-yi: A series of work (most from MS) focuses on meager resources but not wholly unsupervised. Perhaps we have to mention the papers.

\subsubsection{ASR-TTS}
%{\color{black} Shinji}\\
%{\color{black} Hung-yi: In the last meeting, we discussed the connection between this section and self-supervised learning. In addition to looking for papers using self-supervision to improve the ASR-TTS framework here, I think there may be another direction. Can we regard this ASR-TTS framework as a kind of self-supervised learning? ASR-TTS is similar to an autoencoder, and the output of ASR is "latent representation". In other words, this section describes a special self-supervised learning method, whose representation is expressed as "text."}\\
%{\color{black} Hung-yi: However, the above idea sounds tricky. In addition, the ``ASR-TTS'' autoencoder is learned by semi-supervised, not self-supervised.}
% AMH: to make the paper more consistent, I recommend changing all $f^{\mathrm{asr}}$ to $f_{\mathrm{asr}}$
Here we describe an alternative approach by which to train an ASR and \edit{text-to-speech (TTS)} system
based on unpaired text and audio. The ASR-TTS framework, which \edit{combines the ASR} and
TTS systems in a cascaded manner,
can be regarded as an autoencoder, where the encoder~$f$
corresponds to the ASR module and the decoder~$g$ corresponds to the TTS module.
In this framework, we consider the intermediate ASR output as a latent
representation; the framework as a whole can be regarded as a variant of
self-supervised learning.\footnote{However, to make this complicated
system work, we often require that data is paired. Therefore, in practice,
ASR-TTS and other methods described in this section are categorized as
semi-supervised learning.}

The ASR-TTS framework can jointly optimize both ASR and TTS \edit{without using paired data}~\cite{tjandra2017listening,hori2019cycle,wang2020improving}.
A speech chain~\cite{tjandra2017listening,tjandra2018machine} is one 
successful way to utilize audio-only and text-only data to train both
end-to-end ASR/TTS models.
This approach first prepares pre-trained ASR model \edit{
$f_{\mathrm{asr}}(X)$ with acoustic input $X$ and pre-trained TTS model $g_{\mathrm{tts}}(Y)$ with text input $Y$.
By following the TTS system with an ASR system, we generate new 
acoustic feature sequence~$\hat{X}$, which must be close to the original input~$X$.
Thus, we design a loss function $\mathcal{L}_{\mathrm{asr} \rightarrow \mathrm{tts}}(X, \hat{X})$, where $\hat{X}$ is generated by
%and~$\hat{X}$:
%\begin{equation}
%    \mathcal{L}_{\mathrm{asr} \rightarrow \mathrm{tts}} (X, \hat{X}),
%\end{equation}    
%where
\begin{equation}
    \hat{X} = g_{\mathrm{tts}}(f_{\mathrm{asr}}(X)).
    \label{eq:asr-tts}
\end{equation}
Thus, we train the ASR model (or both ASR and TTS models) using only
the acoustic input by minimizing $\mathcal{L}_{\mathrm{asr} \rightarrow \mathrm{tts}}$.  
%\begin{equation}
%\hat{\theta}_{\mathrm{asr}} = \arg \min _{\theta_{\mathrm{asr}}} \mathcal{L}(Y, \hat{Y}; \theta_{\mathrm{asr}}, \theta_{\mathrm{tts}}).
%\end{equation}
Note that this approach does not require the supervised text data~$Y$}.
As an analogy to the generative approach in \cref{sec:generative}, the
intermediate ASR output~$\hat{Y}$ can be regarded as the latent representation~$Z$.

The other cycle with the text-only data~$Y$ is also accomplished by the
concatenated TTS-ASR systems:\edit{
\begin{equation}
    \hat{Y} = f_{\mathrm{asr}}(g_{\mathrm{tts}}(Y)).
    \label{eq:tts-asr}
\end{equation}
Similarly, this approach does not require the supervised audio data~$X$, and the
intermediate TTS output~$\hat{X}$ can be regarded as the latent representation~$Z$}.
Although this approach initially freezes either the ASR or TTS model, 
extensions of this study~\cite{hori2019cycle,tjandra2019end,baskar2019semi}
implement the joint training of both ASR and TTS parameters using 
REINFORCE~\cite{williams1992simple} and straight-through estimators.

An emerging technique uses a well-trained TTS system to generate
speech and text data from text-only data.
This technique is a sub-problem of the TTS-ASR approach formulated in
\eqref{eq:tts-asr} in which we fix the TTS system part and estimate only the ASR
parameters. 
For example, a huge amount of text resources can be obtained from the web
and document archives without corresponding audio data.
%Self-supervised learning in this paper mostly focuses on representation learning perspectives of such unpaired data.
The typical use case scenario of such a text resource for ASR is through \edit{the
language model}. 
We combine the ASR and language model via a noisy channel 
model~\cite{jelinek1997statistical}, a weighted finite state 
transducer~\cite{mohri2002weighted}, or shallow 
fusion~\cite{gulcehre2015using,chorowski2016towards}.
However, the progress of TTS systems boosted by deep 
learning~\cite{oord2016wavenet,shen2018natural} has inspired another interesting and
straightforward research direction: \edit{artificially creating paired text and audio
data $\{\hat{X}, Y\}$ with only text data~$Y$ by generating the corresponding
audio data~$\hat{X}$ with TTS.} %\begin{equation}
%    \{\hat{X}, Y\} \text{ where } \hat{X} = g_{\mathrm{tts}} (Y).
%\end{equation}
The most straightforward approach is to simply use multi-speaker TTS to
generate the waveform with various acoustic 
variations~\cite{li2018training,ueno2019multi,rosenberg2019speech,laptev2020you,huang2020using}.
The other approaches are based on the generation of high-level (more
linguistic) features instead of generating the waveform, e.g., encoder
features~\cite{hayashi2018back} and phoneme 
features~\cite{renduchintala18_interspeech,masumura20_interspeech}.
This approach is similar to the back-translation technique developed in \edit{neural machine translation}~\cite{sennrich2016improving}.
One benefit of the above data generation approaches is that it can be used to feed
unseen word or context phrases to end-to-end ASR.

% {\color{black} Shinji: puts some notions to encourage the community to contributions to work on this direction and has more connections to the current SSL.}

\subsection{No text or lexicon} \label{subsec:zero}

% \kl{moved the acoustic word embedding subsection from here to the rep learning methods section.}\\

\subsubsection{Zero-resource speech technologies and challenges}
% {\color{black} Shinji}\\
\label{zero_speech}
Zero-resource speech technologies, which seek to discover linguistic concepts
from audio only (no text nor lexicon), are one of the most active applications
of unsupervised/self-supervised speech processing.
Zero-resource speech technologies were initially studied for acoustic and
linguistic unit discovery from speech data without linguistic resources,
e.g., transcriptions and other annotations~\cite{jansen2013summary}.
This study was motivated by unsupervised query-by-example, applications of
non-parametric Bayesian machine learning to speech processing, and low-resource
speech recognition, and was also inspired by the learning process of infants.
The goal of this type of work is to build spoken dialog systems in a zero-resource
setup for any language.
% To facilitate the zero-resource problem in the community, 
  To encourage      zero-resource research,                  % AMH: check
zero-resource speech challenges have been organized since 2015.

In this section, we describe the research directions of zero-resource speech
technologies by following the series of zero-resource speech challenges.
\begin{itemize}
	 \item \edit{Zero Resource Speech Challenge 2015~\cite{versteegh2015zero} mainly focused on building an acoustic model without using any
	 linguistic annotations based on subword unit modeling and spoken term
	 discovery tracks}.
	 \edit{For the subword unit modeling track, the ABX score for the within- and across-speaker
	 tasks was used as an evaluation metric.}
	 The spoken term discovery track used the normalized edit distance and coverage
	 scores in addition to the precision, recall, and F1 scores for types,
	 tokens, and boundaries.
    Both tracks were based on the English and Xitsonga languages.
	 \item The Zero Resource Speech Challenge 2017~\cite{dunbar2017zero} focused on 
	 unseen language and speaker aspects from the previous challenge. For
	 example, to demonstrate the robustness against unseen languages, the systems
	 were developed with English, French, and Mandarin and tested on
	 two ``surprise'' languages: German and Wolof.
	 Similarly, robustness against unseen speakers was demonstrated by varying
	 the amount of speech available for each speaker.
	 \item The Zero Resource Speech Challenge 2019~\cite{dunbar2019zero} extended a goal of
	 previous challenges by synthesizing speech without
	 text or phonetic labels but with acoustic units obtained using
	 zero-resource techniques.
	 The evaluation metrics were also extended to subjectively evaluate the
	 quality of synthesized speech, including its intelligibility, naturalness,
	 and speaker similarity.
	 \item The Zero Resource Speech Challenge 2020~\cite{dunbar2020zero} was based on two
	 tracks, revisiting previous challenges with different evaluation metrics. 
	 The first task revisited the 2019 challenge with low bit-rate subword
	 representations that optimize the quality of speech synthesis. The second
	 task revisited the 2017 challenge by focusing on the discovery of word-like
	 units from unsegmented raw speech.
	 \item The Zero Resource Speech Challenge 2021~\cite{nguyen2020zero}, the latest
	 \edit{challenge}, expanded the scope to include language modeling tasks.
	 In addition to phoneme-level ABX, the challenge includes lexical,
	 semantic, and syntactic evaluation metrics computed via a language model of
	 pseudo-acoustic labels.
\end{itemize}
These challenges have facilitated the tracking of technical trends in
zero-resource speech technologies.
For example, research directions thereof 
have expanded to various speech processing components to cover the entire
spoken dialogue systems.
To keep up with this expansion, the challenge has continued to develop
appropriate evaluation metrics for zero-resource scenarios.
Following the success of representation learning, baseline and challenge
techniques have shifted from purely generative 
models~\cite{ondel2016variational,heck2017feature}, deep 
autoencoders~\cite{tjandra19_interspeech,chorowski2019unsupervised}, and incorporation of
neural-network-based TTS/VC techniques~\cite{niekerk20b_interspeech} to
self-supervised learning~\cite{maekaku21_interspeech}.
The latest challenge included the visual modality, continuing the
expansion to include more aspects of human interaction.

\subsubsection{Textless NLP}
% {\color{black} Abdo}\\

Textless NLP is a new research direction that leverages the progress mentioned
above in self-supervised speech representation learning to model language
directly from audio, bypassing the need for text or 
labels~\cite{GSLM,SpeechResynthesis_IS21,pGSLM,emotion_conversion}. 
Not only does this open the gate for language and dialect modeling without
orthographic rules, but it also offers the opportunity to model other
non-lexical information about how speech is delivered, e.g., speaker identity,
emotion, hesitation, interruptions. 
The generative spoken language model (GSLM)~\cite{GSLM} utilizes discrete
representations from wav2vec~2.0, HuBERT, and CPC algorithms as inputs to an
autoregressive language model trained by \edit{using the cross-entropy function} to maximize the
probability of predicting the next discrete speech token. A \edit{synthesis} module
follows the language model to produce speech waveforms given the generated
discrete speech units. The generated spoken continuations compete with
supervised generations and synthesis using a character language model in
subjective human evaluations. The model completes incomplete words
(pow[..] $\rightarrow$ POWER) and continues using words in the same general mood (dark $\rightarrow$
BLACKNESS)\footnote{https://speechbot.github.io/gslm/} and has been extended to
model and generate
dialogues~\cite{dialogue_GSLM}.\footnote{https://speechbot.github.io/dgslm/}
%\cite{}. 
Given its flexibility in modeling spoken content, the GSLM has been further extended
to jointly model content and prosody~\cite{pGSLM}. This prosodic-GSLM model
introduced a multistream causal Transformer, where the input and output layers
use multiple heads to model three channels:  discrete speech units, duration,
and quantized pitch. The prosodic-GSLM model jointly generates novel content
and prosody congruently in the expressive 
style of the prompt.\footnote{https://speechbot.github.io/pgslm/}
%\cite{}. 
Going one step further, \cite{emotion_conversion} used a speech emotion
conversion framework to modify the perceived emotion of a speech utterance
while preserving its lexical content and speaker identity. Other studies have
extended the idea of textless language processing or audio discrete representation to applications such as 
spoken question answering~\cite{spoken_qa_dual}, speech separation~\cite{shi2021discretization}, TTS~\cite{hayashi2020discretalk}, and speech-to-speech
translation~\cite{textless_translation}.

 %Reviewer: TNS, Abdo, Hung-yi Lee
\section{Discussion and Conclusion}
\label{sec:conclusion}
% {\color{blue} Reviewer: Daniel, SW }\\
% {\color{blue} Abdo}\\

In this overview, we have presented the historical context of self-supervised learning and provided a thorough methodological review of important  self-supervised speech representation models. Specifically, we have categorized the approaches into three categories, generative, contrastive and predictive, differing in terms of how the pretext task is defined.
We have presented an overview of existing benchmarks and reviewed the efforts towards efficient zero-resource learning.
\edit{Although the field is progressing rapidly, with new approaches reaching higher levels of performance, a couple of patterns have emerged: (1) The solid performance of Wav2vec 2.0 for speech recognition and many downstream tasks, as well as the public availability of its pre-trained multilingual variants, enabled wide adoption in the community making it a ``standard'' go-to model. (2) The simplicity and stability of the HuBERT approach, as well as the resemblance of its training procedure to classic frame-level ASR systems, made it an easy choice for research extensions on improving representation quality, speech translation, and textless NLP.}

Below we \edit{highlight} various \edit{shortcomings of existing work and } \edit{future} research directions:
\begin{itemize}

\item \textbf{Using the representation model.} So far, there are two main ways to use representation models: Freeze the representation models and use them as feature extractors, or fine-tune the representation models \edit{on} downstream tasks. 
\edit{Some efficient methods for leveraging SSL models exist in the NLP community. 
Adapters~\cite{pmlr-v97-houlsby19a,zaken2021bitfit,guo-etal-2021-parameter} are lightweight modules inserted into SSL models, and in downstream tasks, the parameters of SSL models are frozen, and only the adapters are trained. 
The prompt/instruction learning methods~\cite{liu2021pretrain} also freeze the SSL parameters and control the output of SSL by adding additional information, which is called \textit{prompt}, in the input.
Both adapter-based methods and prompt/instruction learning yield competitive performance compared with fine-tuning in NLP applications, but there is only little related work for speech~\cite{SpeechAdapter,SpeechPrompt}.
In addition, prompt for speech SSL does not achieve comparable performance on sequence generation tasks like phoneme recognition and slot filling, so how to use prompt is still an open question.}

\item \textbf{Increasing the efficiency of the representation model.} As discussed in \cref{subsec:modelsize}, larger representation models lead to better downstream performance. Despite the success of these \edit{large} models, they \edit{incur high costs in terms of memory and time for pre-training, fine-tuning, and even when used only to extract representations without gradient calculation. This} makes them unsuitable for edge devices \edit{but also limits the ability to scale these models to very large datasets} \ci{ -- and leads to a large energy consumption}. 
Preliminary studies have been conducted on compressing speech representation models through network pruning~\cite{PARP} or knowledge distillation~\cite{chang2021distilhubert}. 
\edit{There has been quite some effort towards more efficient general neural network models via conditional computing \cite{bengio_conditional_2016} and neural network quantization \cite{gholami_survey_2021} as well as extensive work on improving the specific efficiency of Transformer models, especially with the focus on self-attention \cite{tay_efficient_2022}, but these technology has not been widely used in speech SSL.}
\edit{Because speech is intrinsically represented as sequence, one way to reduce computation is to reduce the length of speech representation sequence but still keep the vital information in speech. But we have not been aware of any publication in this direction when writing this paper.}
On the other hand, non-streaming architectures in models such as the bidirectional Transformer have hindered the representation model used in streaming scenarios, leading to studies that address these problems~\cite{StreamingW2v}. 
We anticipate research in these directions to continue in the future.

\edit{\item \textbf{Data-efficient approaches.} SOTA representation learning methods require large volumes of unlabeled speech during pre-training, going way beyond what babies need to understand language. 
Different learning approaches have different data needs, e.g., generative approaches could be more data efficient than contrastive or predictive approaches since they are constrained by more bits of information to reconstruct their inputs. Comprehensive research is needed to study the data efficiency of different approaches. }

\item \textbf{Feature Disentanglement.} 
Speech SSL models show strengths on a surprisingly wide range of tasks~\cite{yang21c_interspeech}, suggesting that representations contain different information.
One way to further improve downstream tasks is to disentangle different information from the representation.
For example, we can decompose the representation into content embedding and speaker embedding and use content embedding for ASR and speaker embedding for SID.
Some work has been in this direction~\cite{pmlr-v162-qian22b,NANCY,MultimodelAmazonIS22}.

\item \textbf{Creating robust models.} \edit{As discussed in \cref{subsec:robustness}, studies have been conducted on the robustness of representation models \cite{wu2021characterizing}. However, the failure modes of SSL models are still poorly understood, and it remains unclear whether they provide more or less robustness to adversarial attacks than fully supervised models. Due to the importance of this research direction, while writing this paper, there is already some related research about enhancing the robustness of SSL models~\cite{robustw2v2,huang2022improving,wang2022improving,zhu2022noise} and identifying their vulnerability to adversarial attack~\cite{wu2021characterizing}. }

\edit{\item \textbf{Capturing higher-level semantic information.} Although many representation learning approaches can go beyond low-level phonetic modeling to capture some lexical information~\cite{TuAnh_2022_discrete}, they still struggle in higher-level semantic tasks easily captured by word-level counterparts like BERT. One workaround is two-stage training~\cite{pGSLM, dialogue_GSLM}; however, this prevents propagating rich lexical and semantic knowledge modeled in the second stage to benefit the phonetically focused first stage. }

\item \textbf{Using text representation models to improve speech representation.} The amount of content information in speech corpora used to train speech representation models is far less than that of text representation models. Noting that the BERT training corpus exceeds 3 billion words~\cite{jacob2019BERT}, and assuming a typical speaking rate of 120 words per minute, a speech corpus containing the same content as the BERT training data would include 400,000 hours of audio, which exceeds the \edit{accumulated} training data of all current speech representation models. Therefore, to enable speech representation models to better learn human language, for instance by extracting semantic information from acoustic signals, the use of text models such as BERT and GPT  \edit{seems} key: nevertheless, how to use these to improve speech representation model pre-training remains an open question. 
%\edit{ There is already some study using both speech and text data to pre-training, but some paired data is still required to achieve good performance~\cite{bapna2021slam}.  } 

\end{itemize}

We believe SSL representation models have considerable room to grow. The relationship between representation models and downstream tasks can be compared to the relationship between operating systems and applications. Today, even individuals can build applications with desired functions on a smartphone because the smartphone's operating system handles the complex communication with the hardware and provides a convenient developer interface. Likewise, as SSL representation models learn general knowledge from human speech, it is easy to develop new speech processing applications on this basis. From this viewpoint, \edit{speech representation} models will play the role of operating systems in speech processing and further facilitate the continued development of speech technology. 
%Suppose, for instance, a speaker of an indigenous language seeks to purchase an intelligent assistant, but discovers that it does not yet support the indigenous language. Because the smart assistant has a built-in representation model, though, it can quickly learn a new language with little supervision. We believe SSL representation models will bring the benefits of speech technology to more people.
 %Reviewer: Daniel, SW

\ifCLASSOPTIONcaptionsoff
  \newpage
\fi

\bibliographystyle{IEEEtran}
\bibliography{conference,IEEEful,refs}

% Generated by IEEEtran.bst, version: 1.14 (2015/08/26)
\begin{thebibliography}{100}
\providecommand{\url}[1]{#1}
\csname url@samestyle\endcsname
\providecommand{\newblock}{\relax}
\providecommand{\bibinfo}[2]{#2}
\providecommand{\BIBentrySTDinterwordspacing}{\spaceskip=0pt\relax}
\providecommand{\BIBentryALTinterwordstretchfactor}{4}
\providecommand{\BIBentryALTinterwordspacing}{\spaceskip=\fontdimen2\font plus
\BIBentryALTinterwordstretchfactor\fontdimen3\font minus
  \fontdimen4\font\relax}
\providecommand{\BIBforeignlanguage}[2]{{%
\expandafter\ifx\csname l@#1\endcsname\relax
\typeout{** WARNING: IEEEtran.bst: No hyphenation pattern has been}%
\typeout{** loaded for the language `#1'. Using the pattern for}%
\typeout{** the default language instead.}%
\else
\language=\csname l@#1\endcsname
\fi
#2}}
\providecommand{\BIBdecl}{\relax}
\BIBdecl

\bibitem{lecun2015deep}
Y.~LeCun, Y.~Bengio, and G.~Hinton, ``Deep learning,'' \emph{Nature}, vol. 521,
  no. 7553, pp. 436--444, 2015.

\bibitem{hinton2012deep}
G.~Hinton, L.~Deng, D.~Yu, G.~E. Dahl, A.-r. Mohamed, N.~Jaitly, A.~Senior,
  V.~Vanhoucke, P.~Nguyen, T.~N. Sainath \emph{et~al.}, ``Deep neural networks
  for acoustic modeling in speech recognition: The shared views of four
  research groups,'' \emph{IEEE Signal Processing Magazine}, vol.~29, no.~6,
  pp. 82--97, 2012.

\bibitem{bourlard2012connectionist}
H.~A. Bourlard and N.~Morgan, \emph{Connectionist speech recognition: A hybrid
  approach}.\hskip 1em plus 0.5em minus 0.4em\relax Springer Science \&
  Business Media, 2012, vol. 247.

\bibitem{Kemp1999}
T.~Kemp and A.~Waibel, ``Unsupervised training of a speech recognizer: Recent
  experiments,'' \emph{Proceedings of European Conference on Speech
  Communication and Technology}, 1999.

\bibitem{csl01_limsi}
L.~Lamel, J.-L. Gauvain, and G.~Adda, ``Lightly supervised and unsupervised
  acoustic model training,'' \emph{Computer Speech \& Language}, 2002.

\bibitem{ma_bbn_06}
J.~Ma, S.~Matsoukas, O.~Kimball, and R.~Schwartz, ``Unsupervised training on
  large amounts of broadcast news data,'' in \emph{Proceedings of {IEEE}
  International Conference on Acoustics, Speech and Signal Processing}, 2006.

\bibitem{Hinton_2007}
G.~E. Hinton, ``Learning multiple layers of representation,'' \emph{Trends in
  Cognitive Sciences}, vol.~11, pp. 428--434, 2007.

\bibitem{LeCun06atutorial}
Y.~LeCun, S.~Chopra, R.~Hadsell, F.~J. Huang \emph{et~al.}, ``A tutorial on
  energy-based learning,'' in \emph{Predicting Structured Data}.\hskip 1em plus
  0.5em minus 0.4em\relax MIT Press, 2006.

\bibitem{bengio_representation_2013}
Y.~Bengio, A.~C. Courville, and P.~Vincent, ``Representation learning: {A}
  review and new perspectives,'' \emph{{IEEE} Transactions on Pattern Analysis
  Machine Intelligence}, vol.~35, no.~8, pp. 1798--1828, 2013.

\bibitem{jordan_2015}
M.~I. Jordan and T.~M. Mitchell, ``Machine learning: Trends, perspectives, and
  prospects,'' \emph{Science}, vol. 349, no. 6245, pp. 255--260, 2015.

\bibitem{vq}
R.~Gray, ``Vector quantization,'' \emph{IEEE ASSP Magazine}, vol.~1, no.~2, pp.
  4--29, 1984.

\bibitem{MoE}
M.~Jordan and R.~Jacobs, ``Hierarchical mixtures of experts and the {EM}
  algorithm,'' in \emph{Proceedings of 1993 International Conference on Neural
  Networks (IJCNN-93-Nagoya, Japan)}, vol.~2, 1993.

\bibitem{hinton_94}
G.~E. Hinton and R.~Zemel, ``Autoencoders, minimum description length and
  {H}elmholtz free energy,'' in \emph{Advances in Neural Information Processing
  Systems}, vol.~6, 1994.

\bibitem{nmf}
D.~D. Lee and H.~S. Seung, ``Learning the parts of objects by nonnegative
  matrix factorization,'' \emph{Nature}, vol. 401, pp. 788--791, 1999.

\bibitem{hinton_2006}
G.~E. Hinton and R.~R. Salakhutdinov, ``Reducing the dimensionality of data
  with neural networks,'' \emph{Science}, vol. 313, no. 5786, pp. 504--507,
  2006.

\bibitem{bommasani2021opportunities}
R.~Bommasani, D.~A. Hudson, E.~Adeli, R.~Altman, S.~Arora, S.~von Arx, M.~S.
  Bernstein, J.~Bohg, A.~Bosselut, E.~Brunskill \emph{et~al.}, ``On the
  opportunities and risks of foundation models,'' 2021.

\bibitem{ericsson2021selfsupervised}
L.~Ericsson, H.~Gouk, C.~C. Loy, and T.~M. Hospedales, ``Self-supervised
  representation learning: Introduction, advances and challenges,'' 2021.

\bibitem{LiuSSLsurvey}
X.~Liu, F.~Zhang, Z.~Hou, L.~Mian, Z.~Wang, J.~Zhang, and J.~Tang,
  ``Self-supervised learning: Generative or contrastive,'' \emph{IEEE
  Transactions on Knowledge \& Data Engineering}, no.~01, pp. 1--1, Jun 2021.

\bibitem{rogers-etal-2020-primer}
\BIBentryALTinterwordspacing
A.~Rogers, O.~Kovaleva, and A.~Rumshisky, ``A primer in {BERT}ology: What we
  know about how {BERT} works,'' \emph{Transactions of the Association for
  Computational Linguistics}, vol.~8, pp. 842--866, 2020. [Online]. Available:
  \url{https://aclanthology.org/2020.tacl-1.54}
\BIBentrySTDinterwordspacing

\bibitem{liu2021pretrain}
P.~Liu, W.~Yuan, J.~Fu, Z.~Jiang, H.~Hayashi, and G.~Neubig, ``Pre-train,
  prompt, and predict: A systematic survey of prompting methods in natural
  language processing,'' 2021.

\bibitem{xia-etal-2020-bert}
\BIBentryALTinterwordspacing
P.~Xia, S.~Wu, and B.~Van~Durme, ``Which *{BERT}? {A} survey organizing
  contextualized encoders,'' in \emph{Proceedings of the 2020 Conference on
  Empirical Methods in Natural Language Processing (EMNLP)}.\hskip 1em plus
  0.5em minus 0.4em\relax Online: Association for Computational Linguistics,
  Nov. 2020, pp. 7516--7533. [Online]. Available:
  \url{https://aclanthology.org/2020.emnlp-main.608}
\BIBentrySTDinterwordspacing

\bibitem{QiuSSLNLPsurvey}
\BIBentryALTinterwordspacing
X.~Qiu, T.~Sun, Y.~Xu, Y.~Shao, N.~Dai, and X.~Huang, ``Pre-trained models for
  natural language processing: A survey,'' \emph{Science China Technological
  Sciences}, vol.~63, no.~10, p. 1872–1897, Sep 2020. [Online]. Available:
  \url{http://dx.doi.org/10.1007/s11431-020-1647-3}
\BIBentrySTDinterwordspacing

\bibitem{JingSSLCVsurvey}
L.~Jing and Y.~Tian, ``Self-supervised visual feature learning with deep neural
  networks: A survey,'' \emph{IEEE Transactions on Pattern Analysis \& Machine
  Intelligence}, vol.~43, no.~11, pp. 4037--4058, nov 2021.

\bibitem{borgholt_22}
L.~Borgholt, J.~D. Havtorn, J.~Edin, L.~Maaløe, and C.~Igel, ``A brief
  overview of unsupervised neural speech representation learning,'' in
  \emph{The 2nd Workshop on Self-supervised Learning for Audio and Speech
  Processing (AAAI-SAS-2022)}, 2022.

\bibitem{latif2021deep}
S.~Latif, R.~Rana, S.~Khalifa, R.~Jurdak, J.~Qadir, and B.~W. Schuller, ``Deep
  representation learning in speech processing: Challenges, recent advances,
  and future trends,'' 2021.

\bibitem{Rabiner1979}
L.~Rabiner and J.~Wilpon, ``Considerations in applying clustering techniques to
  speaker independent word recognition,'' in \emph{IEEE International
  Conference on Acoustics, Speech, and Signal Processing}, vol.~4, 1979, pp.
  578--581.

\bibitem{Wilpon1985}
J.~Wilpon and L.~Rabiner, ``A modified k-means clustering algorithm for use in
  isolated work recognition,'' \emph{IEEE Transactions on Acoustics, Speech,
  and Signal Processing}, vol.~33, no.~3, pp. 587--594, 1985.

\bibitem{Gauvain1994}
J.-L. Gauvain and C.-H. Lee, ``Maximum a posteriori estimation for multivariate
  {G}aussian mixture observations of {M}arkov chains,'' \emph{IEEE Transactions
  on Speech and Audio Processing}, vol.~2, no.~2, pp. 291--298, 1994.

\bibitem{Young1994}
S.~Young and P.~Woodland, ``State clustering in hidden {M}arkov model-based
  continuous speech recognition,'' \emph{Computer Speech \& Language}, vol.~8,
  no.~4, pp. 369--383, 1994.

\bibitem{Bahl1986}
L.~Bahl, P.~Brown, P.~de~Souza, and R.~Mercer, ``Maximum mutual information
  estimation of hidden {M}arkov model parameters for speech recognition,'' in
  \emph{Proceedings of {IEEE} International Conference on Acoustics, Speech and
  Signal Processing}, vol.~11, 1986, pp. 49--52.

\bibitem{smith01}
N.~Smith and M.~Gales, ``Speech recognition using {SVM}s,'' in \emph{NIPS},
  2001.

\bibitem{venkata03}
V.~Venkataramani, S.~Chakrabartty, and W.~Byrne, ``Support vector machines for
  segmental minimum {B}ayes risk decoding of continuous speech,'' in
  \emph{ASRU}, 2003.

\bibitem{wan03}
V.~Wan and S.~Renals, ``{SVMSVM:} support vector machine speaker verification
  methodology,'' in \emph{ICASSP}, 2003, pp. 221--224.

\bibitem{dehak11a}
N.~Dehak, P.~Kenny, R.~Dehak, P.~Dumouchel, and P.~Ouellet, ``Front-end factor
  analysis for speaker verification,'' \emph{IEEE Transactions on Audio,
  Speech, and Language Processing}, vol. 19(4), pp. 788--798, 2011.

\bibitem{dehak11b}
N.~Dehak, P.~.Torres-Carrasquillo, D.~Reynolds, and R.~Dehak, ``Language
  recognition via i-vectors and dimensionality reduction,'' in
  \emph{Proceedings of the Annual Conference of the International Speech
  Communication Association}, 2011.

\bibitem{DAE}
P.~Vincent, H.~Larochelle, I.~Lajoie, Y.~Bengio, and P.-A. Manzagol, ``Stacked
  denoising autoencoders: Learning useful representations in a deep network
  with a local denoising criterion,'' \emph{Journal of Machine Learning
  Research}, 2010.

\bibitem{gutmann2012noise}
M.~U. Gutmann and A.~Hyv{\"a}rinen, ``Noise-contrastive estimation of
  unnormalized statistical models, with applications to natural image
  statistics.'' \emph{Journal of Machine Learning Research}, vol.~13, no.~2,
  2012.

\bibitem{Olshausen1996}
B.~Olshausen and D.~Field, ``Emergence of simple-cell receptive field
  properties by learning a sparse code for natural images,'' \emph{Nature},
  vol. 381, pp. 607--609, June 1996.

\bibitem{sparse_lee}
H.~Lee, A.~Battle, R.~Raina, and A.~Y. Ng, ``Efficient sparse coding
  algorithms,'' in \emph{Proceedings of the 19th International Conference on
  Neural Information Processing Systems}.\hskip 1em plus 0.5em minus
  0.4em\relax MIT Press, 2006, p. 801–808.

\bibitem{sivaram2010sparse}
G.~S. Sivaram, S.~K. Nemala, M.~Elhilali, T.~D. Tran, and H.~Hermansky,
  ``Sparse coding for speech recognition,'' in \emph{2010 IEEE International
  Conference on Acoustics, Speech and Signal Processing}.\hskip 1em plus 0.5em
  minus 0.4em\relax IEEE, 2010, pp. 4346--4349.

\bibitem{Ranzato2007}
M.~Ranzato, Y.-L. Boureau, S.~Chopra, and Y.~LeCun, ``A unified energy-based
  framework for unsupervised learning,'' in \emph{Proc. Eleventh International
  Conference on Artificial Intelligence and Statistics}, 2007, pp. 371--379.

\bibitem{hinton_cd_2002}
G.~E. Hinton, ``Training products of experts by minimizing contrastive
  divergence,'' \emph{Neural Comput.}, vol.~14, no.~8, p. 1771–1800, aug
  2002.

\bibitem{GLUE}
A.~Wang, A.~Singh, J.~Michael, F.~Hill, O.~Levy, and S.~R. Bowman, ``{GLUE}: A
  multi-task benchmark and analysis platform for natural language
  understanding,'' in \emph{Proceedings of International Conference on Learning
  Representations}, 2019.

\bibitem{yang21c_interspeech}
S.~wen Yang, P.-H. Chi, Y.-S. Chuang, C.-I.~J. Lai, K.~Lakhotia, Y.~Y. Lin,
  A.~T. Liu, J.~Shi, X.~Chang, G.-T. Lin, T.-H. Huang, W.-C. Tseng, K.~tik Lee,
  D.-R. Liu, Z.~Huang, S.~Dong, S.-W. Li, S.~Watanabe, A.~Mohamed, and
  H.~yi~Lee, ``{SUPERB: Speech Processing Universal PERformance Benchmark},''
  in \emph{Proceedings of the Annual Conference of the International Speech
  Communication Association}, 2021.

\bibitem{colorizing}
R.~Zhang, P.~Isola, and A.~A. Efros, ``Colorful image colorization,''
  \emph{CoRR}, vol. abs/1603.08511, 2016.

\bibitem{deepcluster}
M.~Caron, P.~Bojanowski, A.~Joulin, and M.~Douze, ``Deep clustering for
  unsupervised learning of visual features,'' \emph{CoRR}, vol. abs/1807.05520,
  2018.

\bibitem{context_pred}
C.~Doersch, A.~Gupta, and A.~A. Efros, ``Unsupervised visual representation
  learning by context prediction,'' \emph{CoRR}, vol. abs/1505.05192, 2015.

\bibitem{vae}
D.~P. Kingma and M.~Welling, ``Auto-encoding variational {B}ayes,'' in
  \emph{2nd International Conference on Learning Representations, {ICLR} 2014,
  Banff, AB, Canada, April 14--16, 2014, Conference Track Proceedings}, 2014.

\bibitem{rezende2014stochastic}
D.~J. Rezende, S.~Mohamed, and D.~Wierstra, ``Stochastic backpropagation and
  approximate inference in deep generative models,'' in \emph{International
  Conference on Machine Learning}.\hskip 1em plus 0.5em minus 0.4em\relax PMLR,
  2014, pp. 1278--1286.

\bibitem{Girin2021}
L.~Girin, S.~Leglaive, X.~Bie, J.~Diard, T.~Hueber, and X.~Alameda-Pineda,
  ``Dynamical variational autoencoders: A comprehensive review,''
  \emph{Foundations and Trends® in Machine Learning}, vol.~15, pp. 1--175, 12
  2021.

\bibitem{peters2018deep}
M.~E. Peters, M.~Neumann, M.~Iyyer, M.~Gardner, C.~Clark, K.~Lee, and
  L.~Zettlemoyer, ``Deep contextualized word representations,'' in
  \emph{NAACL}, 2018.

\bibitem{alex2019GPT2}
A.~Radford, J.~Wu, R.~Child, D.~Luan, D.~Amodei, and I.~Sutskever, ``Language
  models are unsupervised multitask learners,'' 2019.

\bibitem{shoeybi2020megatronlm}
M.~Shoeybi, M.~Patwary, R.~Puri, P.~LeGresley, J.~Casper, and B.~Catanzaro,
  ``{M}egatron-{LM}: Training multi-billion parameter language models using
  model parallelism,'' 2020.

\bibitem{jacob2019BERT}
J.~Devlin, M.-W. Chang, K.~Lee, and K.~Toutanova, ``{BERT}: Pre-training of
  deep bidirectional {T}ransformers for language understanding,'' in
  \emph{NAACL}, 2019.

\bibitem{Liu2019RoBERTa}
Y.~Liu, M.~Ott, N.~Goyal, J.~Du, M.~Joshi, D.~Chen, O.~Levy, M.~Lewis,
  L.~Zettlemoyer, and V.~Stoyanov, ``{RoBERTa}: {A} robustly optimized {BERT}
  pretraining approach,'' \emph{arXiv preprint arXiv:1907.11692}, 2019.

\bibitem{oord2018representation}
A.~v.~d. Oord, Y.~Li, and O.~Vinyals, ``Representation learning with
  contrastive predictive coding,'' \emph{arXiv preprint arXiv:1807.03748},
  2018.

\bibitem{pmlr-v119-chen20j}
T.~Chen, S.~Kornblith, M.~Norouzi, and G.~Hinton, ``A simple framework for
  contrastive learning of visual representations,'' in \emph{Proceedings of the
  37th International Conference on Machine Learning}, 2020.

\bibitem{he2020momentum}
K.~He, H.~Fan, Y.~Wu, S.~Xie, and R.~Girshick, ``Momentum contrast for
  unsupervised visual representation learning,'' in \emph{CVPR}, 2020.

\bibitem{chen2020improved}
X.~Chen, H.~Fan, R.~Girshick, and K.~He, ``Improved baselines with momentum
  contrastive learning,'' 2020.

\bibitem{SwAV}
M.~Caron, I.~Misra, J.~Mairal, P.~Goyal, P.~Bojanowski, and A.~Joulin,
  ``Unsupervised learning of visual features by contrasting cluster
  assignments,'' in \emph{Proceedings of Advances in Neural Information
  Processing Systems}, 2020.

\bibitem{hari_1mhour}
S.~H.~K. Parthasarathi and N.~Strom, ``Lessons from building acoustic models
  with a million hours of speech,'' \emph{CoRR}, vol. abs/1904.01624, 2019.

\bibitem{park2020improved}
D.~S. Park, Y.~Zhang, Y.~Jia, W.~Han, C.-C. Chiu, B.~Li, Y.~Wu, and Q.~V. Le,
  ``{Improved Noisy Student Training for Automatic Speech Recognition},'' in
  \emph{Proceedings of the Annual Conference of the International Speech
  Communication Association}, 2020.

\bibitem{xu2020iterative}
Q.~Xu, T.~Likhomanenko, J.~Kahn, A.~Hannun, G.~Synnaeve, and R.~Collobert,
  ``{Iterative Pseudo-Labeling for Speech Recognition},'' in \emph{Proceedings
  of the Annual Conference of the International Speech Communication
  Association}, 2020.

\bibitem{xiao_scaling_2021}
A.~Xiao, W.~Zheng, G.~Keren, D.~Le, F.~Zhang, C.~Fuegen, O.~Kalinli, Y.~Saraf,
  and A.~Mohamed, ``Scaling {ASR} improves zero and few shot learning,''
  \emph{CoRR}, vol. abs/2111.05948, 2021.

\bibitem{caruana1997multitask}
R.~Caruana, ``Multitask learning,'' \emph{Machine Learning}, vol.~28, no.~1,
  pp. 41--75, 1997.

\bibitem{Cui2015MultilingualRF}
J.~Cui, B.~Kingsbury, B.~Ramabhadran, A.~Sethy, K.~Audhkhasi, X.~Cui,
  E.~Kislal, L.~Mangu, M.~Nu{\ss}baum-Thom, M.~Picheny, Z.~T{\"u}ske, P.~Golik,
  R.~Schl{\"u}ter, H.~Ney, M.~J.~F. Gales, K.~Knill, A.~Ragni, H.~Wang, and
  P.~C. Woodland, ``Multilingual representations for low resource speech
  recognition and keyword search,'' \emph{2015 IEEE Workshop on Automatic
  Speech Recognition and Understanding (ASRU)}, 2015.

\bibitem{survey_speech_TL}
P.~Bell, J.~Fainberg, O.~Klejch, J.~Li, S.~Renals, and P.~Swietojanski,
  ``Adaptation algorithms for neural network-based speech recognition: An
  overview,'' \emph{IEEE Open Journal of Signal Processing}, vol.~2, pp.
  33--66, 2021.

\bibitem{pasad2021}
A.~Pasad, J.-C. Chou, and K.~Livescu, ``Layer-wise analysis of a
  self-supervised speech representation model,'' in \emph{Proceedings of IEEE
  Workshop on Automatic Speech Recognition and Understanding}, 2021.

\bibitem{pascual2019learning}
S.~Pascual, M.~Ravanelli, J.~Serr{\`a}, A.~Bonafonte, and Y.~Bengio, ``Learning
  problem-agnostic speech representations from multiple self-supervised
  tasks,'' in \emph{Proceedings of the Annual Conference of the International
  Speech Communication Association}, 2019.

\bibitem{chung_recurrent_2015}
J.~Chung, K.~Kastner, L.~Dinh, K.~Goel, A.~C. Courville, and Y.~Bengio, ``A
  {{Recurrent Latent Variable Model}} for {{Sequential Data}},'' in
  \emph{Proceedings of the 29th {{Conference}} on {{Neural Information
  Processing Systems}} ({{NeurIPS}})}, {Montr\'eal, Quebec, Canada}, 2015,
  p.~9.

\bibitem{fraccaro_sequential_2016}
M.~Fraccaro, S.~K. S{\o}nderby, U.~Paquet, and O.~Winther, ``Sequential
  {{Neural Models}} with {{Stochastic Layers}},'' in \emph{Proceedings of the
  30th {{Conference}} on {{Neural Information Processing Systems}}
  ({{NeurIPS}})}, {Barcelona, Spain}, 2016.

\bibitem{hsu2017learning}
W.-N. Hsu, Y.~Zhang, and J.~Glass, ``Learning latent representations for speech
  generation and transformation,'' in \emph{Proceedings of the Annual
  Conference of the International Speech Communication Association}, 2017.

\bibitem{hsu2017unsupervised}
------, ``Unsupervised learning of disentangled and interpretable
  representations from sequential data,'' in \emph{Proceedings of Advances in
  Neural Information Processing Systems}, 2017.

\bibitem{aksan_stcn_2019}
E.~Aksan and O.~Hilliges, ``{{STCN}}: {{Stochastic Temporal Convolutional
  Networks}},'' in \emph{Proceedings of the 7th {{International Conference}} on
  {{Learning Representations}} ({{ICLR}})}, {New Orleans, LA, USA}, Feb. 2019.

\bibitem{vqvae}
A.~van~den Oord, O.~Vinyals, and K.~Kavukcuoglu, ``Neural discrete
  representation learning,'' 2017.

\bibitem{SpeechResynthesis_IS21}
A.~Polyak, Y.~Adi, J.~Copet, E.~Kharitonov, K.~Lakhotia, W.-N. Hsu, A.~Mohamed,
  and E.~Dupoux, ``Speech resynthesis from discrete disentangled
  self-supervised representations,'' in \emph{Proceedings of the Annual
  Conference of the International Speech Communication Association}, 2021.

\bibitem{pGSLM}
E.~Kharitonov, A.~Lee, A.~Polyak, Y.~Adi, J.~Copet, K.~Lakhotia, T.~A. Nguyen,
  M.~Rivi{\`{e}}re, A.~Mohamed, E.~Dupoux, and W.~Hsu, ``Text-free
  prosody-aware generative spoken language modeling,'' \emph{arXiv preprint
  arXiv:2109.03264}, 2021.

\bibitem{GSLM}
K.~Lakhotia, E.~Kharitonov, W.-N. Hsu, Y.~Adi, A.~Polyak, B.~Bolte, T.-A.
  Nguyen, J.~Copet, A.~Baevski, A.~Mohamed, and E.~Dupoux, ``On generative
  spoken language modeling from raw audio,'' \emph{Transactions of the
  Association for Computational Linguistics}, vol.~9, pp. 1336--1354, 2021.

\bibitem{bengio2013estimating}
Y.~Bengio, N.~L{\'e}onard, and A.~Courville, ``Estimating or propagating
  gradients through stochastic neurons for conditional computation,''
  \emph{arXiv preprint arXiv:1308.3432}, 2013.

\bibitem{oord2016wavenet}
A.~v.~d. Oord, S.~Dieleman, H.~Zen, K.~Simonyan, O.~Vinyals, A.~Graves,
  N.~Kalchbrenner, A.~Senior, and K.~Kavukcuoglu, ``Wave{N}et: A generative
  model for raw audio,'' \emph{arXiv preprint arXiv:1609.03499}, 2016.

\bibitem{burton_generalization_1983}
D.~K. Burton, J.~E. Shore, and J.~T. Buck, ``A generalization of isolated word
  recognition using vector quantization,'' \emph{Proceedings of {IEEE}
  International Conference on Acoustics, Speech and Signal Processing}, 1983.

\bibitem{soong_vector_1985}
F.~Soong, A.~Rosenberg, and L.~R.~B. Juang, ``A vector quantization approach to
  speaker recognition,'' \emph{Proceedings of {IEEE} International Conference
  on Acoustics, Speech and Signal Processing}, 1985.

\bibitem{chorowski2019unsupervised}
J.~Chorowski, R.~J. Weiss, S.~Bengio, and A.~van~den Oord, ``Unsupervised
  speech representation learning using {W}ave{N}et autoencoders,''
  \emph{IEEE/ACM Transactions on Audio, Speech, and Language Processing},
  vol.~27, no.~12, pp. 2041--2053, 2019.

\bibitem{Gumbel-Softmax}
E.~Jang, S.~Gu, and B.~Poole, ``Categorical reparameterization with
  {G}umbel-softmax,'' in \emph{Proceedings of International Conference on
  Learning Representations}, 2017.

\bibitem{eloff2019unsupervised}
R.~Eloff, A.~Nortje, B.~van Niekerk, A.~Govender, L.~Nortje, A.~Pretorius,
  E.~van Biljon, E.~van~der Westhuizen, L.~van Staden, and H.~Kamper,
  ``Unsupervised acoustic unit discovery for speech synthesis using discrete
  latent-variable neural networks,'' \emph{Proceedings of the Annual Conference
  of the International Speech Communication Association}, 2019.

\bibitem{zeiler2013rectified}
M.~D. Zeiler, M.~Ranzato, R.~Monga, M.~Mao, K.~Yang, Q.~V. Le, P.~Nguyen,
  A.~Senior, V.~Vanhoucke, J.~Dean \emph{et~al.}, ``On rectified linear units
  for speech processing,'' in \emph{2013 IEEE International Conference on
  Acoustics, Speech and Signal Processing}.\hskip 1em plus 0.5em minus
  0.4em\relax IEEE, 2013, pp. 3517--3521.

\bibitem{badino2014auto}
L.~Badino, C.~Canevari, L.~Fadiga, and G.~Metta, ``An auto-encoder based
  approach to unsupervised learning of subword units,'' in \emph{2014 IEEE
  international conference on acoustics, speech and signal processing
  (ICASSP)}.\hskip 1em plus 0.5em minus 0.4em\relax IEEE, 2014, pp. 7634--7638.

\bibitem{badino2015discovering}
L.~Badino, A.~Mereta, and L.~Rosasco, ``Discovering discrete subword units with
  binarized autoencoders and hidden-markov-model encoders,'' in \emph{Sixteenth
  Annual Conference of the International Speech Communication Association},
  2015.

\bibitem{kamper2015unsupervised}
H.~Kamper, M.~Elsner, A.~Jansen, and S.~Goldwater, ``Unsupervised neural
  network based feature extraction using weak top-down constraints,'' in
  \emph{Proceedings of {IEEE} International Conference on Acoustics, Speech and
  Signal Processing}, 2015.

\bibitem{renshaw2015comparison}
D.~Renshaw, H.~Kamper, A.~Jansen, and S.~Goldwater, ``A comparison of neural
  network methods for unsupervised representation learning on the {Zero
  Resource Speech Challenge},'' \emph{Proceedings of the Annual Conference of
  the International Speech Communication Association}, 2015.

\bibitem{settle2019_a2w}
S.~Settle, K.~Audhkhasi, K.~Livescu, and M.~Picheny, ``Acoustically grounded
  word embeddings for improved acoustics-to-word speech recognition,'' in
  \emph{Proceedings of {IEEE} International Conference on Acoustics, Speech and
  Signal Processing}, 2019.

\bibitem{chung2019unsupervised}
Y.-A. Chung, W.-N. Hsu, H.~Tang, and J.~Glass, ``An unsupervised autoregressive
  model for speech representation learning,'' \emph{arXiv preprint
  arXiv:1904.03240}, 2019.

\bibitem{chung2020generative}
Y.-A. Chung and J.~Glass, ``Generative pre-training for speech with
  autoregressive predictive coding,'' in \emph{Proceedings of {IEEE}
  International Conference on Acoustics, Speech and Signal Processing}, 2020.

\bibitem{LPC}
D.~O'Shaughnessy, ``Linear predictive coding,'' \emph{IEEE Potentials}, vol.~7,
  no.~1, pp. 29--32, 1988.

\bibitem{wiskott2002slow}
L.~Wiskott and T.~J. Sejnowski, ``Slow feature analysis: Unsupervised learning
  of invariances,'' \emph{Neural Computation}, vol.~14, no.~4, pp. 715--770,
  2002.

\bibitem{garofolo1993timit}
\BIBentryALTinterwordspacing
J.~S. Garofolo, ``{TIMIT Acoustic-Phonetic Continuous Speech Corpus LDC93S1},''
  \emph{Linguistic Data Consortium}, 1993. [Online]. Available:
  \url{https://catalog.ldc.upenn.edu/LDC93S1}
\BIBentrySTDinterwordspacing

\bibitem{chung2020improved}
Y.-A. Chung and J.~Glass, ``Improved speech representations with multi-target
  autoregressive predictive coding,'' in \emph{Proceedings of Proceedings of
  the Annual Meeting of the Association for Computational Linguistics}, 2020.

\bibitem{chung20e_interspeech}
Y.-A. Chung, H.~Tang, and J.~Glass, ``{Vector-Quantized Autoregressive
  Predictive Coding},'' in \emph{Proceedings of the Annual Conference of the
  International Speech Communication Association}, 2020.

\bibitem{ling2020deep}
S.~Ling, Y.~Liu, J.~Salazar, and K.~Kirchhoff, ``Deep contextualized acoustic
  representations for semi-supervised speech recognition,'' in
  \emph{Proceedings of {IEEE} International Conference on Acoustics, Speech and
  Signal Processing}, 2020.

\bibitem{liu2020mockingjay}
A.~T. Liu, S.-w. Yang, P.-H. Chi, P.-c. Hsu, and H.-y. Lee, ``Mockingjay:
  Unsupervised speech representation learning with deep bidirectional
  {T}ransformer encoders,'' in \emph{Proceedings of {IEEE} International
  Conference on Acoustics, Speech and Signal Processing}, 2020.

\bibitem{jiang2019improving}
D.~Jiang, X.~Lei, W.~Li, N.~Luo, Y.~Hu, W.~Zou, and X.~Li, ``Improving
  {T}ransformer-based speech recognition using unsupervised pre-training,''
  \emph{arXiv preprint arXiv:1910.09932}, 2019.

\bibitem{liu2020masked}
L.~Liu and Y.~Huang, ``Masked pre-trained encoder base on joint
  {CTC}-{T}ransformer,'' \emph{arXiv preprint arXiv:2005.11978}, 2020.

\bibitem{wang2020unsupervised}
W.~Wang, Q.~Tang, and K.~Livescu, ``Unsupervised pre-training of bidirectional
  speech encoders via masked reconstruction,'' in \emph{Proceedings of {IEEE}
  International Conference on Acoustics, Speech and Signal Processing}, 2020.

\bibitem{jiang2021further}
D.~Jiang, W.~Li, R.~Zhang, M.~Cao, N.~Luo, Y.~Han, W.~Zou, K.~Han, and X.~Li,
  ``A further study of unsupervised pretraining for {T}ransformer based speech
  recognition,'' in \emph{Proceedings of {IEEE} International Conference on
  Acoustics, Speech and Signal Processing}, 2021.

\bibitem{yue2021pMPC}
X.~Yue and H.~Li, ``Phonetically motivated self-supervised speech
  representation learning,'' \emph{Proceedings of the Annual Conference of the
  International Speech Communication Association}, 2021.

\bibitem{liu2021tera}
A.~T. Liu, S.-W. Li, and H.-y. Lee, ``{TERA}: Self-supervised learning of
  {T}ransformer encoder representation for speech,'' \emph{IEEE/ACM
  Transactions on Audio, Speech, and Language Processing}, vol.~29, pp.
  2351--2366, 2021.

\bibitem{liu21l_interspeech}
A.~H. Liu, Y.-A. Chung, and J.~Glass, ``{Non-Autoregressive Predictive Coding
  for Learning Speech Representations from Local Dependencies},'' in
  \emph{Proceedings of the Annual Conference of the International Speech
  Communication Association}, 2021.

\bibitem{XLNet}
Z.~Yang, Z.~Dai, Y.~Yang, J.~Carbonell, R.~R. Salakhutdinov, and Q.~V. Le,
  ``{XLN}et: Generalized autoregressive pretraining for language
  understanding,'' in \emph{Proceedings of Advances in Neural Information
  Processing Systems}, H.~Wallach, H.~Larochelle, A.~Beygelzimer,
  F.~d\textquotesingle Alch\'{e}-Buc, E.~Fox, and R.~Garnett, Eds.,
  vol.~32.\hskip 1em plus 0.5em minus 0.4em\relax Curran Associates, Inc.,
  2019.

\bibitem{song20d_interspeech}
X.~Song, G.~Wang, Y.~Huang, Z.~Wu, D.~Su, and H.~Meng, ``Speech-{XLNet}:
  Unsupervised acoustic model pretraining for self-attention networks,'' in
  \emph{Proceedings of the Annual Conference of the International Speech
  Communication Association}, 2020.

\bibitem{ling2020decoar}
S.~Ling and Y.~Liu, ``{DeCoAR} 2.0: Deep contextualized acoustic
  representations with vector quantization,'' \emph{arXiv preprint
  arXiv:2012.06659}, 2020.

\bibitem{luo2021drop}
J.~Luo, J.~Wang, N.~Cheng, and J.~Xiao, ``Dropout regularization for
  self-supervised learning of {T}ransformer encoder speech representation,''
  \emph{Proceedings of the Annual Conference of the International Speech
  Communication Association}, 2021.

\bibitem{srivastava_dropout_2014}
N.~Srivastava, G.~E. Hinton, A.~Krizhevsky, I.~Sutskever, and R.~R.
  Salakhutdinov, ``Dropout: {{A Simple Way}} to {{Prevent Neural Networks}}
  from {{Overfitting}},'' \emph{Journal of Machine Learning Research}, vol.~15,
  pp. 1929--1958, 2014.

\bibitem{ravanelli2020multi}
M.~Ravanelli, J.~Zhong, S.~Pascual, P.~Swietojanski, J.~Monteiro, J.~Trmal, and
  Y.~Bengio, ``Multi-task self-supervised learning for robust speech
  recognition,'' in \emph{Proceedings of {IEEE} International Conference on
  Acoustics, Speech and Signal Processing}, 2020.

\bibitem{chung2018speech2vec}
Y.-A. Chung and J.~Glass, ``Speech2vec: A sequence-to-sequence framework for
  learning word embeddings from speech,'' \emph{Proceedings of the Annual
  Conference of the International Speech Communication Association}, 2018.

\bibitem{tagliasacchi2019self}
M.~Tagliasacchi, B.~Gfeller, F.~d.~C. Quitry, and D.~Roblek, ``Self-supervised
  audio representation learning for mobile devices,'' \emph{arXiv preprint
  arXiv:1905.11796}, 2019.

\bibitem{tagliasacchi2020pre}
M.~Tagliasacchi, B.~Gfeller, F.~de~Chaumont~Quitry, and D.~Roblek,
  ``Pre-training audio representations with self-supervision,'' \emph{IEEE
  Signal Processing Letters}, vol.~27, pp. 600--604, 2020.

\bibitem{quitry2019learning}
F.~d.~C. Quitry, M.~Tagliasacchi, and D.~Roblek, ``Learning audio
  representations via phase prediction,'' \emph{arXiv preprint
  arXiv:1910.11910}, 2019.

\bibitem{Chung2016AudioWord2Vec}
Y.-A. Chung, C.-C. Wu, C.-H. Shen, H.-Y. Lee, and L.-S. Lee, ``Audio
  word2{V}ec: Unsupervised learning of audio segment representations using
  sequence-to-sequence autoencoder,'' in \emph{Proceedings of the Annual
  Conference of the International Speech Communication Association}, 2016.

\bibitem{chi2020audio}
P.-H. Chi, P.-H. Chung, T.-H. Wu, C.-C. Hsieh, S.-W. Li, and H.-y. Lee, ``Audio
  {ALBERT}: A lite {BERT} for self-supervised learning of audio
  representation,'' \emph{Proceedings of IEEE Spoken Language Technology
  Workshop}, 2021.

\bibitem{milde2018unspeech}
B.~Milde and C.~Biemann, ``Unspeech: Unsupervised speech context embeddings,''
  in \emph{Proceedings of the Annual Conference of the International Speech
  Communication Association}, 2018.

\bibitem{schneider2019wav2vec}
S.~Schneider, A.~Baevski, R.~Collobert, and M.~Auli, ``wav2vec: Unsupervised
  pre-training for speech recognition,'' \emph{arXiv preprint
  arXiv:1904.05862}, 2019.

\bibitem{riviere2020unsupervised}
M.~Riviere, A.~Joulin, P.-E. Mazar{\'e}, and E.~Dupoux, ``Unsupervised
  pretraining transfers well across languages,'' in \emph{Proceedings of {IEEE}
  International Conference on Acoustics, Speech and Signal Processing}, 2020.

\bibitem{kawakami2020learning}
K.~Kawakami, L.~Wang, C.~Dyer, P.~Blunsom, and A.~van~den Oord, ``Learning
  robust and multilingual speech representations,'' in \emph{EMNLP}, 2020.

\bibitem{Baevski2020vq-wav2vec}
A.~Baevski, S.~Schneider, and M.~Auli, ``vq-wav2vec: Self-supervised learning
  of discrete speech representations,'' in \emph{Proceedings of International
  Conference on Learning Representations}, 2020.

\bibitem{baevski2020wav2vec}
A.~Baevski, Y.~Zhou, A.~Mohamed, and M.~Auli, ``wav2vec 2.0: A framework for
  self-supervised learning of speech representations,'' \emph{Proceedings of
  Advances in Neural Information Processing Systems}, vol.~33, 2020.

\bibitem{sadhu21_interspeech}
S.~Sadhu, D.~He, C.-W. Huang, S.~H. Mallidi, M.~Wu, A.~Rastrow, A.~Stolcke,
  J.~Droppo, and R.~Maas, ``{wav2vec-{C}: A Self-Supervised Model for Speech
  Representation Learning},'' in \emph{Proceedings of the Annual Conference of
  the International Speech Communication Association}, 2021.

\bibitem{w2vbert}
Y.~Chung, Y.~Zhang, W.~Han, C.~Chiu, J.~Qin, R.~Pang, and Y.~Wu, ``W2v-{BERT}:
  Combining contrastive learning and masked language modeling for
  self-supervised speech pre-training,'' 2021.

\bibitem{SpeechSimCLR}
D.~Jiang, W.~Li, M.~Cao, W.~Zou, and X.~Li, ``Speech {SIMCLR}: Combining
  contrastive and reconstruction objective for self-supervised speech
  representation learning,'' in \emph{Proceedings of the Annual Conference of
  the International Speech Communication Association}, 2021.

\bibitem{baevski2019effectiveness}
A.~Baevski, M.~Auli, and A.~Mohamed, ``Effectiveness of self-supervised
  pre-training for speech recognition,'' \emph{arXiv preprint
  arXiv:1911.03912}, 2019.

\bibitem{hsu2021hubert}
W.-N. Hsu, B.~Bolte, Y.-H.~H. Tsai, K.~Lakhotia, R.~Salakhutdinov, and
  A.~Mohamed, ``{HuBERT}: Self-supervised speech representation learning by
  masked prediction of hidden units,'' \emph{arXiv preprint arXiv:2106.07447},
  2021.

\bibitem{chen2021wavlm}
S.~Chen, C.~Wang, Z.~Chen, Y.~Wu, S.~Liu, Z.~Chen, J.~Li, N.~Kanda,
  T.~Yoshioka, X.~Xiao, J.~Wu, L.~Zhou, S.~Ren, Y.~Qian, Y.~Qian, J.~Wu,
  M.~Zeng, and F.~Wei, ``Wav{LM}: Large-scale self-supervised pre-training for
  full stack speech processing,'' 2021.

\bibitem{data2vec}
A.~Baevski, W.~Hsu, Q.~Xu, A.~Babu, J.~Gu, and M.~Auli, ``data2vec: {A} general
  framework for self-supervised learning in speech, vision and language,''
  2022.

\bibitem{BEST-RQ}
C.-C. Chiu, J.~Qin, Y.~Zhang, J.~Yu, and Y.~Wu, ``Self-supervised learning with
  random-projection quantizer for speech recognition,'' \emph{arXiv preprint
  arXiv:2202.01855}, 2022.

\bibitem{Schultz_Joachims_NeurIPS2004}
M.~Schultz and T.~Joachims, ``Learning a distance metric from relative
  comparisons,'' in \emph{Advances in Neural Information Processing Systems},
  S.~Thrun, L.~Saul, and B.~Sch\"{o}lkopf, Eds., 2003.

\bibitem{gutmann2010noise}
M.~Gutmann and A.~Hyv{\"a}rinen, ``Noise-contrastive estimation: A new
  estimation principle for unnormalized statistical models,''
  \emph{International Conference on Artificial Intelligence and Statistics
  (AISTATS)}, 2010.

\bibitem{ProductQuantization}
H.~Jégou, M.~Douze, and C.~Schmid, ``Product quantization for nearest neighbor
  search,'' \emph{IEEE Transactions on Pattern Analysis and Machine
  Intelligence}, vol.~33, no.~1, pp. 117--128, 2011.

\bibitem{VQVAE2}
A.~Razavi, A.~van~den Oord, and O.~Vinyals, ``Generating diverse high-fidelity
  images with {VQ-VAE}-2,'' in \emph{Proceedings of Advances in Neural
  Information Processing Systems}, H.~Wallach, H.~Larochelle, A.~Beygelzimer,
  F.~d\textquotesingle Alch\'{e}-Buc, E.~Fox, and R.~Garnett, Eds.,
  vol.~32.\hskip 1em plus 0.5em minus 0.4em\relax Curran Associates, Inc.,
  2019.

\bibitem{Caron_2018_ECCV}
M.~Caron, P.~Bojanowski, A.~Joulin, and M.~Douze, ``Deep clustering for
  unsupervised learning of visual features,'' in \emph{Proceedings of the
  European Conference on Computer Vision (ECCV)}, September 2018.

\bibitem{grill2020byol}
J.~Grill, F.~Strub, F.~Altch{\'{e}}, C.~Tallec, P.~H. Richemond,
  E.~Buchatskaya, C.~Doersch, B.~{\'{A}}. Pires, Z.~D. Guo, M.~G. Azar,
  B.~Piot, K.~Kavukcuoglu, R.~Munos, and M.~Valko, ``Bootstrap {Y}our {O}wn
  {L}atent: A new approach to self-supervised learning,'' \emph{CoRR}, vol.
  abs/2006.07733, 2020.

\bibitem{dino}
M.~Caron, H.~Touvron, I.~Misra, H.~J{\'{e}}gou, J.~Mairal, P.~Bojanowski, and
  A.~Joulin, ``Emerging properties in self-supervised vision {T}ransformers,''
  in \emph{IEEE International Conference on Computer Vision}, 2021.

\bibitem{fastrcnn}
R.~B. Girshick, ``Fast {R-CNN},'' in \emph{IEEE International Conference on
  Computer Vision}, 2015.

\bibitem{paszke2019pytorch}
A.~Paszke \emph{et~al.}, ``Py{T}orch: An imperative style, high-performance
  deep learning library,'' in \emph{Proceedings of Advances in Neural
  Information Processing Systems}, 2019.

\bibitem{petajan1984automatic}
E.~D. Petajan, ``Automatic lipreading to enhance speech recognition (speech
  reading),'' Ph.D. dissertation, University of Illinois at Urbana-Champaign,
  1984.

\bibitem{potamianos2003recent}
G.~Potamianos, C.~Neti, G.~Gravier, A.~Garg, and A.~W. Senior, ``Recent
  advances in the automatic recognition of audiovisual speech,''
  \emph{Proceedings of the IEEE}, vol.~91, no.~9, pp. 1306--1326, 2003.

\bibitem{aleksic2006audio}
P.~S. Aleksic and A.~K. Katsaggelos, ``Audio-visual biometrics,''
  \emph{Proceedings of the IEEE}, vol.~94, no.~11, pp. 2025--2044, 2006.

\bibitem{legerstee1990infants}
M.~Legerstee, ``Infants use multimodal information to imitate speech sounds,''
  \emph{Infant Behavior and Development}, vol.~13, no.~3, pp. 343--354, 1990.

\bibitem{roy1999learning}
D.~Roy, ``Learning from sights and sounds: A computational model,'' \emph{PhD
  Thesis, MIT Media Laboratory}, 1999.

\bibitem{lee2004avicar}
B.~Lee, M.~Hasegawa-Johnson, C.~Goudeseune, S.~Kamdar, S.~Borys, M.~Liu, and
  T.~Huang, ``{AVICAR}: Audio-visual speech corpus in a car environment,'' in
  \emph{Eighth International Conference on Spoken Language Processing}, 2004.

\bibitem{chung2016lip}
J.~S. Chung and A.~Zisserman, ``Lip reading in the wild,'' in \emph{Asian
  Conference on Computer Vision}, 2016.

\bibitem{shi2022learning}
B.~Shi, W.-N. Hsu, K.~Lakhotia, and A.~Mohamed, ``Learning audio-visual speech
  representation by masked multimodal cluster prediction,'' in
  \emph{International Conference on Learning Representations}, 2022.

\bibitem{westbury1990x}
J.~Westbury, P.~Milenkovic, G.~Weismer, and R.~Kent, ``X-ray microbeam speech
  production database,'' \emph{JASA}, vol.~88, no.~S1, pp. S56--S56, 1990.

\bibitem{wrench2001new}
A.~Wrench, ``A new resource for production modelling in speech technology,''
  \emph{Proc. Institute of Acoustics}, vol.~23, no.~3, pp. 207--218, 2001.

\bibitem{narayanan2011multimodal}
S.~Narayanan, E.~Bresch, P.~K. Ghosh, L.~Goldstein, A.~Katsamanis, Y.~Kim,
  A.~Lammert, M.~Proctor, V.~Ramanarayanan, and Y.~Zhu, ``A multimodal
  real-time {MRI} articulatory corpus for speech research,'' in
  \emph{Proceedings of the Annual Conference of the International Speech
  Communication Association}, 2011.

\bibitem{ngiam2011multimodal}
J.~Ngiam, A.~Khosla, M.~Kim, J.~Nam, H.~Lee, and A.~Y. Ng, ``Multimodal deep
  learning,'' in \emph{Proceedings of the 28th International Conference on
  Machine Learning}, 2011.

\bibitem{badino2016integrating}
L.~Badino, C.~Canevari, L.~Fadiga, and G.~Metta, ``Integrating articulatory
  data in deep neural network-based acoustic modeling,'' \emph{Comp. Sp. \&
  Lang.}, vol.~36, pp. 173--195, 2016.

\bibitem{srivastava2012multimodal}
N.~Srivastava and R.~R. Salakhutdinov, ``Multimodal learning with deep
  {B}oltzmann machines,'' in \emph{Proceedings of Advances in Neural
  Information Processing Systems}, 2012, pp. 2222--2230.

\bibitem{hotelling1936relations}
H.~Hotelling, ``Relations between two sets of variates,'' \emph{Biometrika},
  vol.~28, no. 3/4, pp. 321--377, 1936.

\bibitem{andrew2013deep}
G.~Andrew, R.~Arora, J.~A. Bilmes, and K.~Livescu, ``Deep canonical correlation
  analysis,'' in \emph{Proceedings of the 30th International Conference on
  Machine Learning}, 2013.

\bibitem{wang2015deep}
W.~Wang, R.~Arora, K.~Livescu, and J.~A. Bilmes, ``On deep multi-view
  representation learning,'' in \emph{ICML}, 2015.

\bibitem{wang2016deep}
W.~Wang, X.~Yan, H.~Lee, and K.~Livescu, ``Deep variational canonical
  correlation analysis,'' \emph{arXiv preprint arXiv:1610.03454}, 2016.

\bibitem{michaeli2016nonparametric}
T.~Michaeli, W.~Wang, and K.~Livescu, ``Nonparametric canonical correlation
  analysis,'' in \emph{International Conference on Machine Learning}.\hskip 1em
  plus 0.5em minus 0.4em\relax PMLR, 2016, pp. 1967--1976.

\bibitem{Melzer_01a}
T.~Melzer, M.~Reiter, and H.~Bischof, ``Nonlinear feature extraction using
  generalized canonical correlation analysis,'' in \emph{International
  Conference on Artificial Neural Networks}, pp. 353--360.

\bibitem{LaiFyfe00a}
P.~L. Lai and C.~Fyfe, ``Kernel and nonlinear canonical correlation analysis,''
  \emph{Int. J. Neural Syst.}, vol.~10, no.~5, pp. 365--377, 2000.

\bibitem{LaiFyfe99a}
------, ``A neural implementation of canonical correlation analysis,''
  \emph{Neural Networks}, vol.~12, no.~10, pp. 1391--1397, 1999.

\bibitem{BachJordan05a}
F.~R. Bach and M.~I. Jordan, ``A probabilistic interpretation of canonical
  correlation analysis,'' Department of Statistics, University of California,
  Berkeley, Tech. Rep. 688, 2005.

\bibitem{wang_15a}
W.~Wang, R.~Arora, K.~Livescu, and J.~Bilmes, ``Unsupervised learning of
  acoustic features via deep canonical correlation analysis,'' in
  \emph{Proceedings of {IEEE} International Conference on Acoustics, Speech and
  Signal Processing}, 2015, pp. 4590--4594.

\bibitem{HermanBlunsom14a}
K.~M. Hermann and P.~Blunsom, ``Multilingual distributed representations
  without word alignment,'' in \emph{Proceedings of International Conference on
  Learning Representations}.

\bibitem{Huang_13b}
P.-S. Huang, X.~He, J.~Gao, L.~Deng, A.~Acero, and L.~Heck, ``Learning deep
  structured semantic models for web search using clickthrough data,'' in
  \emph{Int. Conf. Information and Knowledge Management}, 2013.

\bibitem{shi2022robust}
B.~Shi, W.-N. Hsu, and A.~Mohamed, ``Robust self-supervised audio-visual speech
  recognition,'' in \emph{Interspeech}, 2022.

\bibitem{chrupala2021visually}
G.~Chrupa{\l{}}a, ``Visually grounded models of spoken language: A survey of
  datasets, architectures and evaluation techniques,'' \emph{arXiv preprint
  arXiv:2104.13225}, 2021.

\bibitem{synnaeve+etal_nipsworkshop14}
G.~Synnaeve, M.~Versteegh, and E.~Dupoux, ``Learning words from images and
  speech,'' in \emph{NIPS Workshop Learn. Semantics}, 2014.

\bibitem{harwath+etal_nips16}
D.~Harwath, A.~Torralba, and J.~R. Glass, ``Unsupervised learning of spoken
  language with visual context,'' in \emph{Proceedings of Advances in Neural
  Information Processing Systems}, 2016.

\bibitem{harwath+glass_asru15}
D.~Harwath and J.~R. Glass, ``Deep multimodal semantic embeddings for speech
  and images,'' in \emph{Proceedings of IEEE Workshop on Automatic Speech
  Recognition and Understanding}, 2015.

\bibitem{merkx2019language}
D.~Merkx, S.~Frank, and M.~Ernestus, ``Language learning using speech to image
  retrieval,'' \emph{Proceedings of the Annual Conference of the International
  Speech Communication Association}, 2019.

\bibitem{rouditchenko2020avlnet}
A.~Rouditchenko, A.~Boggust, D.~Harwath, B.~Chen, D.~Joshi, S.~Thomas,
  K.~Audhkhasi, H.~Kuehne, R.~Panda, R.~Feris \emph{et~al.}, ``{AVL}net:
  Learning audio-visual language representations from instructional videos,''
  in \emph{Proceedings of the Annual Conference of the International Speech
  Communication Association}, 2021.

\bibitem{peng2022fast}
P.~Peng and D.~Harwath, ``Fast-slow transformer for visually grounding
  speech,'' in \emph{Proceedings of {IEEE} International Conference on
  Acoustics, Speech and Signal Processing}, 2022.

\bibitem{ilharco2019large}
G.~Ilharco, Y.~Zhang, and J.~Baldridge, ``Large-scale representation learning
  from visually grounded untranscribed speech,'' in \emph{Proceedings of the
  23rd Conference on Computational Natural Language Learning (CoNLL)}, 2019.

\bibitem{sanabria2021talk}
R.~Sanabria, A.~Waters, and J.~Baldridge, ``Talk, don't write: A study of
  direct speech-based image retrieval,'' in \emph{Proceedings of the Annual
  Conference of the International Speech Communication Association}, 2021.

\bibitem{MultimodelAmazonICASSP22}
D.~M. Chan, S.~Ghosh, D.~Chakrabarty, and B.~Hoffmeister, ``Multi-modal
  pre-training for automated speech recognition,'' in \emph{Proceedings of
  {IEEE} International Conference on Acoustics, Speech and Signal Processing},
  2022.

\bibitem{peng2022self}
P.~Peng and D.~Harwath, ``Self-supervised representation learning for speech
  using visual grounding and masked language modeling,'' in \emph{AAAI SAS
  workshop}, 2022.

\bibitem{harwath2019learning}
D.~Harwath, W.-N. Hsu, and J.~Glass, ``Learning hierarchical discrete
  linguistic units from visually-grounded speech,'' in \emph{Proceedings of
  International Conference on Learning Representations}, 2019.

\bibitem{chrupala2017representations}
G.~Chrupa{\l{}}a, L.~Gelderloos, and A.~Alishahi, ``Representations of language
  in a model of visually grounded speech signal,'' in \emph{Proceedings of the
  55th Annual Meeting of the Association for Computational Linguistics (Volume
  1: Long Papers)}, 2017, pp. 613--622.

\bibitem{scharenborg+etal_icassp18}
O.~Scharenborg \emph{et~al.}, ``Linguistic unit discovery from multi-modal
  inputs in unwritten languages: Summary of the ``{Speaking Rosetta}'' {JSALT
  2017 Workshop},'' in \emph{Proceedings of {IEEE} International Conference on
  Acoustics, Speech and Signal Processing}, 2018.

\bibitem{harwath2018jointly}
D.~Harwath, A.~Recasens, D.~Sur{\'\i}s, G.~Chuang, A.~Torralba, and J.~Glass,
  ``Jointly discovering visual objects and spoken words from raw sensory
  input,'' in \emph{Proceedings of the European Conference on Computer Vision
  (ECCV)}, 2018, pp. 649--665.

\bibitem{peng2022word}
P.~Peng and D.~Harwath, ``Word discovery in visually grounded, self-supervised
  speech models,'' \emph{arXiv preprint arXiv:2203.15081}, 2022.

\bibitem{wang2020dnn}
L.~Wang and M.~Hasegawa-Johnson, ``A {DNN-HMM-DNN} hybrid model for discovering
  word-like units from spoken captions and image regions,'' in
  \emph{Proceedings of the Annual Conference of the International Speech
  Communication Association}, 2020.

\bibitem{harwath2018vision}
D.~Harwath, G.~Chuang, and J.~Glass, ``Vision as an interlingua: Learning
  multilingual semantic embeddings of untranscribed speech,'' in
  \emph{Proceedings of {IEEE} International Conference on Acoustics, Speech and
  Signal Processing}.\hskip 1em plus 0.5em minus 0.4em\relax IEEE, 2018, pp.
  4969--4973.

\bibitem{havard2019models}
W.~N. Havard, J.-P. Chevrot, and L.~Besacier, ``Models of visually grounded
  speech signal pay attention to nouns: A bilingual experiment on {E}nglish and
  {J}apanese,'' in \emph{Proceedings of {IEEE} International Conference on
  Acoustics, Speech and Signal Processing}.\hskip 1em plus 0.5em minus
  0.4em\relax IEEE, 2019, pp. 8618--8622.

\bibitem{kamper+roth_sltu18}
H.~Kamper and M.~Roth, ``Visually grounded cross-lingual keyword spotting in
  speech,'' in \emph{Proc. SLTU}, 2018.

\bibitem{pasad2019contributions}
A.~Pasad, B.~Shi, H.~Kamper, and K.~Livescu, ``On the contributions of visual
  and textual supervision in low-resource semantic speech retrieval,''
  \emph{Proceedings of the Annual Conference of the International Speech
  Communication Association}, 2019.

\bibitem{bapna2021slam}
A.~Bapna, Y.~an~Chung, N.~Wu, A.~Gulati, Y.~Jia, J.~H. Clark, M.~Johnson,
  J.~Riesa, A.~Conneau, and Y.~Zhang, ``{SLAM}: {A} unified encoder for speech
  and language modeling via speech-text joint pre-training,'' 2021.

\bibitem{levin+etal_icassp15}
K.~Levin, A.~Jansen, and B.~Van~Durme, ``Segmental acoustic indexing for zero
  resource keyword search,'' in \emph{Proceedings of {IEEE} International
  Conference on Acoustics, Speech and Signal Processing}, 2015.

\bibitem{guoguo+etal_icassp15}
G.~Chen, C.~Parada, and T.~N. Sainath, ``Query-by-example keyword spotting
  using long short-term memory networks,'' in \emph{Proceedings of {IEEE}
  International Conference on Acoustics, Speech and Signal Processing}, 2015.

\bibitem{settle2017query}
S.~Settle, K.~Levin, H.~Kamper, and K.~Livescu, ``Query-by-example search with
  discriminative neural acoustic word embeddings,'' in \emph{Proceedings of the
  Annual Conference of the International Speech Communication Association},
  2017.

\bibitem{kamper2017embedded}
H.~Kamper, K.~Livescu, and S.~Goldwater, ``An embedded segmental k-means model
  for unsupervised segmentation and clustering of speech,'' in
  \emph{Proceedings of IEEE Workshop on Automatic Speech Recognition and
  Understanding}, 2017.

\bibitem{Kamper17BESGMM}
\BIBentryALTinterwordspacing
H.~Kamper, A.~Jansen, and S.~Goldwater, ``A segmental framework for
  fully-unsupervised large-vocabulary speech recognition,'' \emph{Computer
  Speech \& Language}, vol.~46, pp. 154--174, 2017. [Online]. Available:
  \url{https://www.sciencedirect.com/science/article/pii/S0885230816301905}
\BIBentrySTDinterwordspacing

\bibitem{maas+etal_icmlwrl12}
A.~L. Maas, S.~D. Miller, T.~M. O'neil, A.~Y. Ng, and P.~Nguyen, ``Word-level
  acoustic modeling with convolutional vector regression,'' in \emph{Proc. ICML
  Workshop Representation Learn.}, 2012.

\bibitem{bengio+heigold_interspeech14}
S.~Bengio and G.~Heigold, ``Word embeddings for speech recognition,'' in
  \emph{Proceedings of the Annual Conference of the International Speech
  Communication Association}, 2014.

\bibitem{levin+etal_asru13}
K.~Levin, K.~Henry, A.~Jansen, and K.~Livescu, ``Fixed-dimensional acoustic
  embeddings of variable-length segments in low-resource settings,'' in
  \emph{Proceedings of IEEE Workshop on Automatic Speech Recognition and
  Understanding}, 2013.

\bibitem{holzenberger2018learning}
N.~Holzenberger, M.~Du, J.~Karadayi, R.~Riad, and E.~Dupoux, ``Learning word
  embeddings: Unsupervised methods for fixed-size representations of
  variable-length speech segments,'' in \emph{Proceedings of the Annual
  Conference of the International Speech Communication Association}.\hskip 1em
  plus 0.5em minus 0.4em\relax ISCA, 2018.

\bibitem{kamper2019truly}
H.~Kamper, ``Truly unsupervised acoustic word embeddings using weak top-down
  constraints in encoder-decoder models,'' in \emph{Proceedings of {IEEE}
  International Conference on Acoustics, Speech and Signal Processing}, 2019,
  pp. 6535--3539.

\bibitem{pengcorrespondence}
P.~Peng, H.~Kamper, and K.~Livescu, ``A correspondence variational autoencoder
  for unsupervised acoustic word embeddings,'' in \emph{Proc. NeurIPS Workshop
  on Self-Supervised Learning for Speech and Audio Processing}, 2020.

\bibitem{carlin2011rapid}
M.~A. Carlin, S.~Thomas, A.~Jansen, and H.~Hermansky, ``Rapid evaluation of
  speech representations for spoken term discovery,'' in \emph{Proceedings of
  the Annual Conference of the International Speech Communication Association},
  2011.

\bibitem{kamper2016deep}
H.~Kamper, W.~Wang, and K.~Livescu, ``Deep convolutional acoustic word
  embeddings using word-pair side information,'' in \emph{Proceedings of {IEEE}
  International Conference on Acoustics, Speech and Signal Processing}, 2016.

\bibitem{toshniwal2020cross}
S.~Toshniwal, H.~Shi, B.~Shi, L.~Gao, K.~Livescu, and K.~Gimpel, ``A cross-task
  analysis of text span representations,'' in \emph{Proceedings of the 5th
  Workshop on Representation Learning for NLP}, 2020, pp. 166--176.

\bibitem{wang2021phrase}
S.~Wang, L.~Thompson, and M.~Iyyer, ``Phrase-{BERT}: Improved phrase embeddings
  from {BERT} with an application to corpus exploration,'' in \emph{Proceedings
  of the 2021 Conference on Empirical Methods in Natural Language Processing},
  2021, pp. 10\,837--10\,851.

\bibitem{van2021comparison}
L.~van Staden and H.~Kamper, ``A comparison of self-supervised speech
  representations as input features for unsupervised acoustic word
  embeddings,'' in \emph{Proceedings of IEEE Spoken Language Technology
  Workshop}, 2021, pp. 927--934.

\bibitem{kahn2020libri}
J.~Kahn \emph{et~al.}, ``Libri-light: A benchmark for {ASR} with limited or no
  supervision,'' in \emph{Proceedings of {IEEE} International Conference on
  Acoustics, Speech and Signal Processing}, 2020.

\bibitem{gemmeke2017audio}
J.~F. Gemmeke, D.~P. Ellis, D.~Freedman, A.~Jansen, W.~Lawrence, R.~C. Moore,
  M.~Plakal, and M.~Ritter, ``Audio {S}et: An ontology and human-labeled
  dataset for audio events,'' in \emph{Proceedings of {IEEE} International
  Conference on Acoustics, Speech and Signal Processing}, 2017.

\bibitem{ephrat2018looking}
A.~Ephrat, I.~Mosseri, O.~Lang, T.~Dekel, K.~Wilson, A.~Hassidim, W.~T.
  Freeman, and M.~Rubinstein, ``Looking to listen at the cocktail party: A
  speaker-independent audio-visual model for speech separation,'' \emph{ACM
  Transactions on Graphics (TOG)}, vol.~37, no.~4, pp. 1--11, 2018.

\bibitem{cieri2004fisher}
C.~Cieri, D.~Miller, and K.~Walker, ``The {F}isher corpus: A resource for the
  next generations of speech-to-text,'' in \emph{Proceedings of International
  Conference on Language Resources and Evaluation}, vol.~4, 2004, pp. 69--71.

\bibitem{panayotov2015librispeech}
V.~Panayotov, G.~Chen, D.~Povey, and S.~Khudanpur, ``Libri{S}peech: An {ASR}
  corpus based on public domain audio books,'' in \emph{Proceedings of {IEEE}
  International Conference on Acoustics, Speech and Signal Processing}, 2015.

\bibitem{paul1992design}
D.~B. Paul and J.~Baker, ``The design for the {W}all {S}treet {J}ournal-based
  {CSR} corpus,'' in \emph{Speech and Natural Language: Proceedings of a
  Workshop Held at Harriman, New York, February 23--26}, 1992.

\bibitem{ardila2020common}
R.~Ardila, M.~Branson, K.~Davis, M.~Kohler, J.~Meyer, M.~Henretty, R.~Morais,
  L.~Saunders, F.~Tyers, and G.~Weber, ``Common {V}oice: A
  massively-multilingual speech corpus,'' in \emph{Proceedings of International
  Conference on Language Resources and Evaluation}, 2020.

\bibitem{pratap20_interspeech}
V.~Pratap, Q.~Xu, A.~Sriram, G.~Synnaeve, and R.~Collobert, ``{MLS}: A
  large-scale multilingual dataset for speech research,'' in \emph{Proceedings
  of the Annual Conference of the International Speech Communication
  Association}, 2020.

\bibitem{wang-etal-2021-voxpopuli}
C.~Wang, M.~Riviere, A.~Lee, A.~Wu, C.~Talnikar, D.~Haziza, M.~Williamson,
  J.~Pino, and E.~Dupoux, ``{V}ox{P}opuli: A large-scale multilingual speech
  corpus for representation learning, semi-supervised learning and
  interpretation,'' in \emph{ACL}, 2021.

\bibitem{gales2014speech}
M.~J. Gales, K.~M. Knill, A.~Ragni, and S.~P. Rath, ``Speech recognition and
  keyword spotting for low-resource languages: {BABEL} project research at
  {CUED},'' in \emph{Fourth International Workshop on Spoken Language
  Technologies for Under-resourced Languages (SLTU-2014)}, 2014.

\bibitem{babu2021xlsr}
A.~Babu, C.~Wang, A.~Tjandra, K.~Lakhotia, Q.~Xu, N.~Goyal, K.~Singh, P.~von
  Platen, Y.~Saraf, J.~Pino, A.~Baevski, A.~Conneau, and M.~Auli, ``{XLS-R}:
  Self-supervised cross-lingual speech representation learning at scale,''
  2021.

\bibitem{chen2021unispeechsat}
S.~Chen, Y.~Wu, C.~Wang, Z.~Chen, Z.~Chen, S.~Liu, J.~Wu, Y.~Qian, F.~Wei,
  J.~Li, and X.~Yu, ``Uni{S}peech-{SAT}: Universal speech representation
  learning with speaker aware pre-training,'' 2021.

\bibitem{chen21o_interspeech}
G.~Chen, S.~Chai, G.-B. Wang, J.~Du, W.-Q. Zhang, C.~Weng, D.~Su, D.~Povey,
  J.~Trmal, J.~Zhang, M.~Jin, S.~Khudanpur, S.~Watanabe, S.~Zhao, W.~Zou,
  X.~Li, X.~Yao, Y.~Wang, Z.~You, and Z.~Yan, ``{GigaSpeech: An Evolving,
  Multi-Domain ASR Corpus with 10,000 Hours of Transcribed Audio},'' in
  \emph{Proceedings of the Annual Conference of the International Speech
  Communication Association}, 2021.

\bibitem{hernandez2018ted}
F.~Hernandez, V.~Nguyen, S.~Ghannay, N.~Tomashenko, and Y.~Esteve, ``{TED-LIUM}
  3: Twice as much data and corpus repartition for experiments on speaker
  adaptation,'' in \emph{International Conference on Speech and
  Computer}.\hskip 1em plus 0.5em minus 0.4em\relax Springer, 2018, pp.
  198--208.

\bibitem{rousseau2012ted}
A.~Rousseau, P.~Del{\'e}glise, and Y.~Esteve, ``{TED-LIUM}: An automatic speech
  recognition dedicated corpus,'' in \emph{Proceedings of International
  Conference on Language Resources and Evaluation}, 2012, pp. 125--129.

\bibitem{godfrey1992switchboard}
J.~J. Godfrey, E.~C. Holliman, and J.~McDaniel, ``{SWITCHBOARD}: {T}elephone
  speech corpus for research and development,'' in \emph{Proceedings of {IEEE}
  International Conference on Acoustics, Speech and Signal Processing}, 1992.

\bibitem{valk2021voxlingua107}
J.~Valk and T.~Alum{\"a}e, ``Vox{L}ingua107: A dataset for spoken language
  recognition,'' in \emph{Proceedings of IEEE Spoken Language Technology
  Workshop}, 2021.

\bibitem{liu2006hkust}
Y.~Liu, P.~Fung, Y.~Yang, C.~Cieri, S.~Huang, and D.~Graff, ``{HKUST/MTS}: A
  very large scale {M}andarin telephone speech corpus,'' in \emph{Proceedings
  of International Symposium on Chinese Spoken Language Processing}, 2006.

\bibitem{bu2017aishell}
H.~Bu, J.~Du, X.~Na, B.~Wu, and H.~Zheng, ``{AISHELL}-1: An open-source
  {M}andarin speech corpus and a speech recognition baseline,'' in
  \emph{O-COCOSDA}.\hskip 1em plus 0.5em minus 0.4em\relax IEEE, 2017, pp.
  1--5.

\bibitem{aidatatang}
{Beijing DataTang Technology Co., Ltd.}, ``{aidatatang\_200zh, a free Chinese
  Mandarin speech corpus},'' \url{http://www.datatang.com}.

\bibitem{magic_data}
{Magic Data Technology Co., Ltd.}, ``{MAGICDATA} {M}andarin {C}hinese {R}ead
  {S}peech {C}orpus,'' 2019, \url{http://www.imagicdatatech.com/index.php/home/
  dataopensource/data_info/id/101}.

\bibitem{st_cmds}
Surfingtech, ``{ST-CMDS-20170001\_1, Free ST Chinese Mandarin Corpus}.''

\bibitem{primewords_201801}
{Primewords Information Technology Co., Ltd.}, ``Primewords {C}hinese {C}orpus
  {S}et 1,'' 2018, \url{https://www.primewords.cn}.

\bibitem{ravanelli2015dirha}
M.~Ravanelli, L.~Cristoforetti, R.~Gretter, M.~Pellin, A.~Sosi, and M.~Omologo,
  ``The {DIRHA}-{E}nglish corpus and related tasks for distant-speech
  recognition in domestic environments,'' in \emph{Proceedings of IEEE Workshop
  on Automatic Speech Recognition and Understanding}, 2015.

\bibitem{barker2018fifth}
J.~Barker, S.~Watanabe, E.~Vincent, and J.~Trmal, ``{The fifth `CHiME' Speech
  Separation and Recognition Challenge: Dataset, task and baselines},'' in
  \emph{Proceedings of the Annual Conference of the International Speech
  Communication Association}, 2018.

\bibitem{hozjan2002interface}
V.~Hozjan, Z.~Kacic, A.~Moreno, A.~Bonafonte, and A.~Nogueiras, ``Interface
  databases: Design and collection of a multilingual emotional speech
  database,'' in \emph{Proceedings of International Conference on Language
  Resources and Evaluation}, 2002.

\bibitem{zadeh2018multimodal}
A.~B. Zadeh, P.~P. Liang, S.~Poria, E.~Cambria, and L.-P. Morency, ``Multimodal
  language analysis in the wild: {CMU-MOSEI} dataset and interpretable dynamic
  fusion graph,'' in \emph{Proceedings of Proceedings of the Annual Meeting of
  the Association for Computational Linguistics}, 2018.

\bibitem{veaux2016superseded}
C.~Veaux, J.~Yamagishi, and K.~MacDonald, ``{CSTR VCTK corpus: English
  multi-speaker corpus for CSTR voice cloning toolkit},'' 2016.

\bibitem{nagrani17_interspeech}
A.~Nagrani, J.~S. Chung, and A.~Zisserman, ``{VoxCeleb}: A large-scale speaker
  identification dataset,'' in \emph{Proceedings of the Annual Conference of
  the International Speech Communication Association}, 2017.

\bibitem{lugosch19_interspeech}
L.~Lugosch, M.~Ravanelli, P.~Ignoto, V.~S. Tomar, and Y.~Bengio, ``{Speech
  Model Pre-Training for End-to-End Spoken Language Understanding},'' in
  \emph{Proceedings of the Annual Conference of the International Speech
  Communication Association}, 2019.

\bibitem{anguera2015quesst2014}
X.~Anguera, L.-J. Rodriguez-Fuentes, A.~Buzo, F.~Metze, I.~Sz{\"o}ke, and
  M.~Penagarikano, ``{QUESST}2014: Evaluating query-by-example speech search in
  a zero-resource setting with real-life queries,'' in \emph{Proceedings of
  {IEEE} International Conference on Acoustics, Speech and Signal Processing},
  2015.

\bibitem{kocabiyikoglu2018augmenting}
A.~C. Kocabiyikoglu, L.~Besacier, and O.~Kraif, ``Augmenting {L}ibri{S}peech
  with {F}rench translations: A multimodal corpus for direct speech translation
  evaluation,'' in \emph{Proceedings of International Conference on Language
  Resources and Evaluation}, 2018.

\bibitem{wang2020covost}
C.~Wang, A.~Wu, and J.~Pino, ``{CoVoST} 2 and massively multilingual
  speech-to-text translation,'' \emph{arXiv preprint arXiv:2007.10310}, 2020.

\bibitem{tachbelie2014using}
M.~Y. Tachbelie, S.~T. Abate, and L.~Besacier, ``Using different acoustic,
  lexical and language modeling units for {ASR} of an under-resourced
  language--{A}mharic,'' \emph{Speech Communication}, vol.~56, pp. 181--194,
  2014.

\bibitem{laleye2016first}
F.~A.~A. Laleye, L.~Besacier, E.~C. Ezin, and C.~Motamed, ``First automatic
  {F}ongbe continuous speech recognition system: Development of acoustic models
  and language models,'' in \emph{FedCSIS}.\hskip 1em plus 0.5em minus
  0.4em\relax IEEE, 2016, pp. 477--482.

\bibitem{gelas2012developments}
H.~Gelas, L.~Besacier, and F.~Pellegrino, ``Developments of {S}wahili resources
  for an automatic speech recognition system,'' in \emph{Spoken Language
  Technologies for Under-Resourced Languages}, 2012.

\bibitem{gauthier2016collecting}
E.~Gauthier, L.~Besacier, S.~Voisin, M.~Melese, and U.~P. Elingui, ``Collecting
  resources in sub-{S}aharan {A}frican languages for automatic speech
  recognition: A case study of {W}olof,'' in \emph{LREC 2016}, 2016.

\bibitem{snyder2015musan}
D.~Snyder, G.~Chen, and D.~Povey, ``{MUSAN}: A music, speech, and noise
  corpus,'' \emph{arXiv preprint arXiv:1510.08484}, 2015.

\bibitem{stowell2019automatic}
D.~Stowell, M.~D. Wood, H.~Pamu{\l}a, Y.~Stylianou, and H.~Glotin, ``Automatic
  acoustic detection of birds through deep learning: The first {B}ird {A}udio
  {D}etection {C}hallenge,'' \emph{Methods in Ecology and Evolution}, vol.~10,
  no.~3, pp. 368--380, 2019.

\bibitem{warden2018speech}
P.~Warden, ``{S}peech {C}ommands: A dataset for limited-vocabulary speech
  recognition,'' \emph{arXiv preprint arXiv:1804.03209}, 2018.

\bibitem{oponowiczspoken}
T.~Oponowicz, ``Spoken language identification, 2018,''
  \emph{https://www.kaggle.com/toponowicz/spoken-language-identification}.

\bibitem{dcase2018}
A.~Mesaros, T.~Heittola, and T.~Virtanen, ``Detection and classification of
  acoustic scenes and events,'' \emph{Detection and Classification of Acoustic
  Scenes and Events 2018 Workshop (DCASE2018)}, pp. 9--13, 2018.

\bibitem{engel2017neural}
J.~Engel, C.~Resnick, A.~Roberts, S.~Dieleman, M.~Norouzi, D.~Eck, and
  K.~Simonyan, ``Neural audio synthesis of musical notes with {W}ave{N}et
  autoencoders,'' in \emph{International Conference on Machine Learning}.\hskip
  1em plus 0.5em minus 0.4em\relax PMLR, 2017, pp. 1068--1077.

\bibitem{ravanelli2018learning}
M.~Ravanelli and Y.~Bengio, ``Learning speaker representations with mutual
  information,'' \emph{arXiv preprint arXiv:1812.00271}, 2018.

\bibitem{khurana20_interspeech}
S.~Khurana, A.~Laurent, W.-N. Hsu, J.~Chorowski, A.~Lancucki, R.~Marxer, and
  J.~Glass, ``{A Convolutional Deep Markov Model for Unsupervised Speech
  Representation Learning},'' in \emph{Proceedings of the Annual Conference of
  the International Speech Communication Association}, 2020.

\bibitem{weinberger2009towards}
S.~H. Weinberger and S.~Kunath, ``Towards a typology of {E}nglish accents,''
  \emph{AACL Abstract Book}, vol. 104, 2009.

\bibitem{likhomanenko2020slimipl}
T.~Likhomanenko, Q.~Xu, J.~Kahn, G.~Synnaeve, and R.~Collobert, ``{slimIPL:
  Language-Model-Free Iterative Pseudo-Labeling},'' in \emph{Proceedings of the
  Annual Conference of the International Speech Communication Association},
  2021.

\bibitem{zhang2020pushing}
Y.~Zhang, J.~Qin, D.~S. Park, W.~Han, C.-C. Chiu, R.~Pang, Q.~V. Le, and Y.~Wu,
  ``Pushing the limits of semi-supervised learning for automatic speech
  recognition,'' in \emph{Workshop on Self-Supervised Learning for Speech and
  Audio Processing, NeurIPS}, 2020.

\bibitem{hajibabaei2018unified}
M.~Hajibabaei and D.~Dai, ``Unified hypersphere embedding for speaker
  recognition,'' \emph{arXiv preprint arXiv:1807.08312}, 2018.

\bibitem{hajavi2021siamese}
A.~Hajavi and A.~Etemad, ``Siamese capsule network for end-to-end speaker
  recognition in the wild,'' in \emph{Proceedings of {IEEE} International
  Conference on Acoustics, Speech and Signal Processing}, 2021.

\bibitem{wang2021fine}
Y.~Wang, A.~Boumadane, and A.~Heba, ``A fine-tuned wav2vec 2.0/{HuBERT}
  benchmark for speech emotion recognition, speaker verification and spoken
  language understanding,'' \emph{arXiv preprint arXiv:2111.02735}, 2021.

\bibitem{rodriguez2014gtts}
L.~J. Rodr{\'\i}guez-Fuentes, A.~Varona, M.~Penagarikano, G.~Bordel, and
  M.~Diez, ``{GTTS-EHU} systems for {QUESST} at {MediaEval} 2014,'' in
  \emph{MediaEval}, 2014.

\bibitem{evain21_interspeech}
S.~Evain, H.~Nguyen, H.~Le, M.~Z. Boito, S.~Mdhaffar, S.~Alisamir, Z.~Tong,
  N.~Tomashenko, M.~Dinarelli, T.~Parcollet, A.~Allauzen, Y.~Estève,
  B.~Lecouteux, F.~Portet, S.~Rossato, F.~Ringeval, D.~Schwab, and L.~Besacier,
  ``{LeBenchmark: A Reproducible Framework for Assessing Self-Supervised
  Representation Learning from Speech},'' in \emph{Proceedings of the Annual
  Conference of the International Speech Communication Association}, 2021.

\bibitem{dunbar2020zero}
E.~Dunbar, J.~Karadayi, M.~Bernard, X.-N. Cao, R.~Algayres, L.~Ondel,
  L.~Besacier, S.~Sakti, and E.~Dupoux, ``The {Z}ero {R}esource {S}peech
  {C}hallenge 2020: Discovering discrete subword and word units,'' in
  \emph{Proceedings of the Annual Conference of the International Speech
  Communication Association}, 2020.

\bibitem{pmlr-v176-turian22a}
J.~Turian, J.~Shier, H.~R. Khan, B.~Raj, B.~W. Schuller, C.~J. Steinmetz,
  C.~Malloy, G.~Tzanetakis, G.~Velarde, K.~McNally, M.~Henry, N.~Pinto,
  C.~Noufi, C.~Clough, D.~Herremans, E.~Fonseca, J.~Engel, J.~Salamon,
  P.~Esling, P.~Manocha, S.~Watanabe, Z.~Jin, and Y.~Bisk, ``Hear: Holistic
  evaluation of audio representations,'' in \emph{Proceedings of the NeurIPS
  2021 Competitions and Demonstrations Track}, vol. 176, 2022, pp. 125--145.

\bibitem{shor20_interspeech}
J.~Shor, A.~Jansen, R.~Maor, O.~Lang, O.~Tuval, F.~de~Chaumont~Quitry,
  M.~Tagliasacchi, I.~Shavitt, D.~Emanuel, and Y.~Haviv, ``{Towards Learning a
  Universal Non-Semantic Representation of Speech},'' in \emph{Proceedings of
  the Annual Conference of the International Speech Communication Association},
  2020.

\bibitem{wang2021towards}
L.~Wang, P.~Luc, Y.~Wu, A.~Recasens, L.~Smaira, A.~Brock, A.~Jaegle, J.-B.
  Alayrac, S.~Dieleman, J.~Carreira, and A.~van~den Oord, ``Towards learning
  universal audio representations,'' \emph{arXiv preprint arXiv:2111.12124},
  2021.

\bibitem{tjandra20_interspeech}
A.~Tjandra, S.~Sakti, and S.~Nakamura, ``{Transformer VQ-VAE for Unsupervised
  Unit Discovery and Speech Synthesis: ZeroSpeech 2020 Challenge},'' in
  \emph{Proceedings of the Annual Conference of the International Speech
  Communication Association}, 2020.

\bibitem{niekerk20b_interspeech}
B.~van Niekerk, L.~Nortje, and H.~Kamper, ``{Vector-Quantized Neural Networks
  for Acoustic Unit Discovery in the ZeroSpeech 2020 Challenge},'' in
  \emph{Proceedings of the Annual Conference of the International Speech
  Communication Association}, 2020.

\bibitem{faruqi14}
M.~Faruqi and C.~Dyer, ``Community evaluation and exchange of word vectors and
  wordvectors.org,'' in \emph{Proceedings of the 52nd Annual Meeting of the
  Association for Computational Linguistics: System Demonstration}, 2014.

\bibitem{baevski2021unsupervised}
A.~Baevski, W.-N. Hsu, A.~Conneau, and M.~Auli, ``Unsupervised speech
  recognition,'' \emph{arXiv preprint arXiv:2105.11084}, 2021.

\bibitem{voita2019}
E.~Voita, R.~Sennrich, and I.~Titov, ``The bottom-up evolution of
  representations in the {T}ransformer: A study with machine translation and
  language modeling objectives,'' in \emph{NAACL}, 2019.

\bibitem{grigg2021self}
T.~G. Grigg, D.~Busbridge, J.~Ramapuram, and R.~Webb, ``Do self-supervised and
  supervised methods learn similar visual representations?'' in
  \emph{https://arxiv.org/abs/2110.00528}, 2021.

\bibitem{van2021analyzing}
B.~van Niekerk, L.~Nortje, M.~Baas, and H.~Kamper, ``Analyzing speaker
  information in self-supervised models to improve zero-resource speech
  processing,'' \emph{Proceedings of the Annual Conference of the International
  Speech Communication Association}, 2021.

\bibitem{ling20odyssey}
S.~Ling, J.~Salazar, Y.~Liu, and K.~Kirchhoff, ``{BERTphone}:
  Phonetically-aware encoder representations for utterance-level speaker and
  language recognition,'' in \emph{Proceedings of Odyssey: The Speaker and
  Language Recognition Workshop}, 2020, pp. 9--16.

\bibitem{wang2021layersup}
C.~Wang \emph{et~al.}, ``Self-supervised learning for speech recognition with
  intermediate layer supervision,'' \emph{arXiv e-print 2112.08778}, 2021.

\bibitem{AnalyzeAttention}
S.~wen Yang, A.~T. Liu, and H.~yi~Lee, ``Understanding self-attention of
  self-supervised audio {T}ransformers,'' in \emph{Proceedings of the Annual
  Conference of the International Speech Communication Association}, 2020.

\bibitem{chung2021}
Y.-A. Chung, Y.~Belinkov, and J.~Glass, ``Similarity analysis of
  self-supervised speech representations,'' in \emph{Proceedings of {IEEE}
  International Conference on Acoustics, Speech and Signal Processing}, 2021.

\bibitem{zhou2020}
H.~Zhou, A.~Baevski, and M.~Auli, ``A comparison of discrete latent variable
  models for speech representation learning,'' in
  \emph{https://arxiv.org/abs/2010.14230}, 2020.

\bibitem{rivi20}
M.~Rivière, A.~Joulin, P.-E. Mazaré, and E.~Dupoux, ``Unsupervised
  pretraining transfers well across languages,'' in \emph{Proceedings of {IEEE}
  International Conference on Acoustics, Speech and Signal Processing}, 2020.

\bibitem{pue2021scaling}
J.~Pu, Y.~Yang, R.~Li, O.~Elibol, and J.~Droppo, ``Scaling effect of
  self-supervised models,'' in \emph{Proceedings of the Annual Conference of
  the International Speech Communication Association}, 2021.

\bibitem{versteegh2015zero}
M.~Versteegh, R.~Thiolliere, T.~Schatz, X.~N. Cao, X.~Anguera, A.~Jansen, and
  E.~Dupoux, ``{The Zero Resource Speech Challenge 2015},'' in \emph{Sixteenth
  Annual Conference of the International Speech Communication Association},
  2015.

\bibitem{chang2021exploration}
X.~Chang, T.~Maekaku, P.~Guo, J.~Shi, Y.-J. Lu, A.~S. Subramanian, T.~Wang,
  S.-w. Yang, Y.~Tsao, H.-y. Lee \emph{et~al.}, ``An exploration of
  self-supervised pretrained representations for end-to-end speech
  recognition,'' \emph{arXiv preprint arXiv:2110.04590}, 2021.

\bibitem{robustw2v2}
W.-N. Hsu, A.~Sriram, A.~Baevski, T.~Likhomanenko, Q.~Xu, V.~Pratap, J.~Kahn,
  A.~Lee, R.~Collobert, G.~Synnaeve, and M.~Auli, ``Robust wav2vec 2.0:
  Analyzing domain shift in self-supervised pre-training,'' in
  \emph{Proceedings of the Annual Conference of the International Speech
  Communication Association}, 2021.

\bibitem{conneau20xling}
A.~Conneau, A.~Baevski, R.~Collobert, A.~Mohamed, and M.~Auli, ``Unsupervised
  cross-lingual representation learning for speech recognition,'' \emph{CoRR,
  abs/2006.13979}, 2020.

\bibitem{liu2018completely}
D.-R. Liu, K.-Y. Chen, H.-Y. Lee, and L.-s. Lee, ``Completely unsupervised
  phoneme recognition by adversarially learning mapping relationships from
  audio embeddings,'' \emph{Proceedings of the Annual Conference of the
  International Speech Communication Association}, pp. 3748--3752, 2018.

\bibitem{Wang2018SegAudioWord2Vec}
Y.-H. Wang, H.~yi~Lee, and L.~shan Lee, ``Segmental audio word2{V}ec:
  Representing utterances as sequences of vectors with applications in spoken
  term detection,'' in \emph{Proceedings of {IEEE} International Conference on
  Acoustics, Speech and Signal Processing}, 2018.

\bibitem{gulrajani2017improved}
I.~Gulrajani, F.~Ahmed, M.~Arjovsky, V.~Dumoulin, and A.~Courville, ``Improved
  training of {W}asserstein {GAN}s,'' \emph{Proceedings of the 31st
  International Conference on Neural Information Processing Systems}, pp.
  5769--5779, 2017.

\bibitem{chung2018unsupervised}
Y.-A. Chung, W.-H. Weng, S.~Tong, and J.~Glass, ``Unsupervised cross-modal
  alignment of speech and text embedding spaces,'' \emph{Proceedings of
  Advances in Neural Information Processing Systems}, vol.~31, pp. 7354--7364,
  2018.

\bibitem{conneau2017word}
A.~Conneau, G.~Lample, M.~Ranzato, L.~Denoyer, and H.~J{\'e}gou, ``Word
  translation without parallel data,'' \emph{Proceedings of International
  Conference on Learning Representations}, 2018.

\bibitem{chung2019towards}
Y.-A. Chung, W.-H. Weng, S.~Tong, and J.~Glass, ``Towards unsupervised
  speech-to-text translation,'' in \emph{Proceedings of {IEEE} International
  Conference on Acoustics, Speech and Signal Processing}, 2019, pp. 7170--7174.

\bibitem{artetxe-etal-2018-robust}
\BIBentryALTinterwordspacing
M.~Artetxe, G.~Labaka, and E.~Agirre, ``A robust self-learning method for fully
  unsupervised cross-lingual mappings of word embeddings,'' in
  \emph{Proceedings of the 56th Annual Meeting of the Association for
  Computational Linguistics (Volume 1: Long Papers)}.\hskip 1em plus 0.5em
  minus 0.4em\relax Melbourne, Australia: Association for Computational
  Linguistics, Jul. 2018, pp. 789--798. [Online]. Available:
  \url{https://aclanthology.org/P18-1073}
\BIBentrySTDinterwordspacing

\bibitem{yeh2018unsupervised}
C.-K. Yeh, J.~Chen, C.~Yu, and D.~Yu, ``Unsupervised speech recognition via
  segmental empirical output distribution matching,'' \emph{Proceedings of
  International Conference on Learning Representations}, 2018.

\bibitem{wang2017gate}
Y.-H. Wang, C.-T. Chung, and H.-y. Lee, ``Gate activation signal analysis for
  gated recurrent neural networks and its correlation with phoneme
  boundaries,'' \emph{Proceedings of the Annual Conference of the International
  Speech Communication Association}, 2017.

\bibitem{Liu2017ODM}
Y.~Liu, J.~Chen, and L.~Deng, ``Unsupervised sequence classification using
  sequential output statistics,'' in \emph{Proceedings of Advances in Neural
  Information Processing Systems}, 2017.

\bibitem{chen2019completely}
K.-Y. Chen, C.-P. Tsai, D.-R. Liu, H.-Y. Lee, and L.-s. Lee, ``Completely
  unsupervised speech recognition by a generative adversarial network
  harmonized with iteratively refined hidden {M}arkov models,''
  \emph{Proceedings of the Annual Conference of the International Speech
  Communication Association}, pp. 1856--1860, 2019.

\bibitem{Deciphering2022}
O.~Klejch, E.~Wallington, and P.~Bell, ``Deciphering speech: a zero-resource
  approach to cross-lingual transfer in {ASR},'' \emph{arXiv preprint
  arXiv:2111.06799}, 2022.

\bibitem{ravi-knight-2011-deciphering}
\BIBentryALTinterwordspacing
S.~Ravi and K.~Knight, ``Deciphering foreign language,'' in \emph{Proceedings
  of the 49th Annual Meeting of the Association for Computational Linguistics:
  Human Language Technologies}.\hskip 1em plus 0.5em minus 0.4em\relax
  Portland, Oregon, USA: Association for Computational Linguistics, Jun. 2011,
  pp. 12--21. [Online]. Available: \url{https://aclanthology.org/P11-1002}
\BIBentrySTDinterwordspacing

\bibitem{w2v-u2}
A.~H. Liu, W.-N. Hsu, M.~Auli, and A.~Baevski, ``Towards end-to-end
  unsupervised speech recognition,'' \emph{arXiv preprint arXiv:2204.02492},
  2022.

\bibitem{goodfellow2014generative}
I.~Goodfellow, J.~Pouget-Abadie, M.~Mirza, B.~Xu, D.~Warde-Farley, S.~Ozair,
  A.~Courville, and Y.~Bengio, ``Generative adversarial nets,'' in
  \emph{Proceedings of Advances in Neural Information Processing Systems},
  2014, pp. 2672--2680.

\bibitem{arjovsky2017wasserstein}
M.~Arjovsky, S.~Chintala, and L.~Bottou, ``Wasserstein {GAN},''
  \emph{Proceedings of the 34th International Conference on Machine Learning},
  pp. 214--223, 2017.

\bibitem{artetxe2017unsupervised}
M.~Artetxe, G.~Labaka, E.~Agirre, and K.~Cho, ``Unsupervised neural machine
  translation,'' \emph{Proceedings of International Conference on Learning
  Representations}, 2018.

\bibitem{lample2017unsupervised}
G.~Lample, A.~Conneau, L.~Denoyer, and M.~Ranzato, ``Unsupervised machine
  translation using monolingual corpora only,'' \emph{Proceedings of
  International Conference on Learning Representations}, 2018.

\bibitem{lin2021analyzing}
G.-T. Lin, C.-J. Hsu, D.-R. Liu, H.-Y. Lee, and Y.~Tsao, ``Analyzing the
  robustness of unsupervised speech recognition,'' 2021.

\bibitem{tjandra2017listening}
A.~Tjandra, S.~Sakti, and S.~Nakamura, ``Listening while speaking: Speech chain
  by deep learning,'' in \emph{Proceedings of IEEE Workshop on Automatic Speech
  Recognition and Understanding}.\hskip 1em plus 0.5em minus 0.4em\relax IEEE,
  2017, pp. 301--308.

\bibitem{hori2019cycle}
T.~Hori, R.~Astudillo, T.~Hayashi, Y.~Zhang, S.~Watanabe, and J.~Le~Roux,
  ``Cycle-consistency training for end-to-end speech recognition,'' in
  \emph{Proceedings of {IEEE} International Conference on Acoustics, Speech and
  Signal Processing}.\hskip 1em plus 0.5em minus 0.4em\relax IEEE, 2019, pp.
  6271--6275.

\bibitem{wang2020improving}
G.~Wang, A.~Rosenberg, Z.~Chen, Y.~Zhang, B.~Ramabhadran, Y.~Wu, and P.~Moreno,
  ``Improving speech recognition using consistent predictions on synthesized
  speech,'' in \emph{Proceedings of {IEEE} International Conference on
  Acoustics, Speech and Signal Processing}.\hskip 1em plus 0.5em minus
  0.4em\relax IEEE, 2020, pp. 7029--7033.

\bibitem{tjandra2018machine}
A.~Tjandra, S.~Sakti, and S.~Nakamura, ``Machine speech chain with one-shot
  speaker adaptation,'' \emph{arXiv preprint arXiv:1803.10525}, 2018.

\bibitem{tjandra2019end}
------, ``End-to-end feedback loss in speech chain framework via
  straight-through estimator,'' in \emph{Proceedings of {IEEE} International
  Conference on Acoustics, Speech and Signal Processing}.\hskip 1em plus 0.5em
  minus 0.4em\relax IEEE, 2019, pp. 6281--6285.

\bibitem{baskar2019semi}
M.~K. Baskar, S.~Watanabe, R.~Astudillo, T.~Hori, L.~Burget, and
  J.~{\v{C}}ernock{\`y}, ``Semi-supervised sequence-to-sequence {ASR} using
  unpaired speech and text,'' \emph{arXiv preprint arXiv:1905.01152}, 2019.

\bibitem{williams1992simple}
R.~J. Williams, ``Simple statistical gradient-following algorithms for
  connectionist reinforcement learning,'' \emph{Machine Learning}, vol.~8,
  no.~3, pp. 229--256, 1992.

\bibitem{jelinek1997statistical}
F.~Jelinek, \emph{Statistical Methods for Speech Recognition}.\hskip 1em plus
  0.5em minus 0.4em\relax MIT press, 1997.

\bibitem{mohri2002weighted}
M.~Mohri, F.~Pereira, and M.~Riley, ``Weighted finite-state transducers in
  speech recognition,'' \emph{Computer Speech \& Language}, vol.~16, no.~1, pp.
  69--88, 2002.

\bibitem{gulcehre2015using}
C.~Gulcehre, O.~Firat, K.~Xu, K.~Cho, L.~Barrault, H.-C. Lin, F.~Bougares,
  H.~Schwenk, and Y.~Bengio, ``On using monolingual corpora in neural machine
  translation,'' \emph{arXiv preprint arXiv:1503.03535}, 2015.

\bibitem{chorowski2016towards}
J.~Chorowski and N.~Jaitly, ``Towards better decoding and language model
  integration in sequence to sequence models,'' \emph{arXiv preprint
  arXiv:1612.02695}, 2016.

\bibitem{shen2018natural}
J.~Shen, R.~Pang, R.~J. Weiss, M.~Schuster, N.~Jaitly, Z.~Yang, Z.~Chen,
  Y.~Zhang, Y.~Wang, R.~Skerrv-Ryan \emph{et~al.}, ``Natural {TTS} synthesis by
  conditioning {W}ave{N}et on mel spectrogram predictions,'' in
  \emph{Proceedings of {IEEE} International Conference on Acoustics, Speech and
  Signal Processing}.\hskip 1em plus 0.5em minus 0.4em\relax IEEE, 2018, pp.
  4779--4783.

\bibitem{li2018training}
J.~Li, R.~Gadde, B.~Ginsburg, and V.~Lavrukhin, ``Training neural speech
  recognition systems with synthetic speech augmentation,'' \emph{arXiv
  preprint arXiv:1811.00707}, 2018.

\bibitem{ueno2019multi}
S.~Ueno, M.~Mimura, S.~Sakai, and T.~Kawahara, ``Multi-speaker
  sequence-to-sequence speech synthesis for data augmentation in
  acoustic-to-word speech recognition,'' in \emph{Proceedings of {IEEE}
  International Conference on Acoustics, Speech and Signal Processing}.\hskip
  1em plus 0.5em minus 0.4em\relax IEEE, 2019, pp. 6161--6165.

\bibitem{rosenberg2019speech}
A.~Rosenberg, Y.~Zhang, B.~Ramabhadran, Y.~Jia, P.~Moreno, Y.~Wu, and Z.~Wu,
  ``Speech recognition with augmented synthesized speech,'' in
  \emph{Proceedings of IEEE Workshop on Automatic Speech Recognition and
  Understanding}.\hskip 1em plus 0.5em minus 0.4em\relax IEEE, 2019, pp.
  996--1002.

\bibitem{laptev2020you}
A.~Laptev, R.~Korostik, A.~Svischev, A.~Andrusenko, I.~Medennikov, and
  S.~Rybin, ``You do not need more data: Improving end-to-end speech
  recognition by text-to-speech data augmentation,'' in \emph{2020 13th
  International Congress on Image and Signal Processing, BioMedical Engineering
  and Informatics (CISP-BMEI)}.\hskip 1em plus 0.5em minus 0.4em\relax IEEE,
  2020, pp. 439--444.

\bibitem{huang2020using}
Y.~Huang, L.~He, W.~Wei, W.~Gale, J.~Li, and Y.~Gong, ``Using personalized
  speech synthesis and neural language generator for rapid speaker
  adaptation,'' in \emph{Proceedings of {IEEE} International Conference on
  Acoustics, Speech and Signal Processing}.\hskip 1em plus 0.5em minus
  0.4em\relax IEEE, 2020, pp. 7399--7403.

\bibitem{hayashi2018back}
T.~Hayashi, S.~Watanabe, Y.~Zhang, T.~Toda, T.~Hori, R.~Astudillo, and
  K.~Takeda, ``Back-translation-style data augmentation for end-to-end {ASR},''
  in \emph{Proceedings of IEEE Spoken Language Technology Workshop}.\hskip 1em
  plus 0.5em minus 0.4em\relax IEEE, 2018, pp. 426--433.

\bibitem{renduchintala18_interspeech}
A.~Renduchintala, S.~Ding, M.~Wiesner, and S.~Watanabe, ``{Multi-Modal Data
  Augmentation for End-to-end ASR},'' in \emph{Proceedings of the Annual
  Conference of the International Speech Communication Association}, 2018.

\bibitem{masumura20_interspeech}
R.~Masumura, N.~Makishima, M.~Ihori, A.~Takashima, T.~Tanaka, and S.~Orihashi,
  ``{Phoneme-to-Grapheme Conversion Based Large-Scale Pre-Training for
  End-to-End Automatic Speech Recognition},'' in \emph{Proceedings of the
  Annual Conference of the International Speech Communication Association},
  2020.

\bibitem{sennrich2016improving}
R.~Sennrich, B.~Haddow, and A.~Birch, ``Improving neural machine translation
  models with monolingual data,'' in \emph{Proceedings of the 54th Annual
  Meeting of the Association for Computational Linguistics (Volume 1: Long
  Papers)}, 2016, pp. 86--96.

\bibitem{jansen2013summary}
A.~Jansen, E.~Dupoux, S.~Goldwater, M.~Johnson, S.~Khudanpur, K.~Church,
  N.~Feldman, H.~Hermansky, F.~Metze, R.~Rose \emph{et~al.}, ``A summary of the
  2012 {JHU} {CLSP} workshop on zero resource speech technologies and models of
  early language acquisition,'' in \emph{2013 IEEE International Conference on
  Acoustics, Speech and Signal Processing}.\hskip 1em plus 0.5em minus
  0.4em\relax IEEE, 2013, pp. 8111--8115.

\bibitem{dunbar2017zero}
E.~Dunbar, X.~N. Cao, J.~Benjumea, J.~Karadayi, M.~Bernard, L.~Besacier,
  X.~Anguera, and E.~Dupoux, ``{The Zero Resource Speech Challenge 2017},'' in
  \emph{Proceedings of IEEE Workshop on Automatic Speech Recognition and
  Understanding}.\hskip 1em plus 0.5em minus 0.4em\relax IEEE, 2017, pp.
  323--330.

\bibitem{dunbar2019zero}
E.~Dunbar, R.~Algayres, J.~Karadayi, M.~Bernard, J.~Benjumea, X.-N. Cao,
  L.~Miskic, C.~Dugrain, L.~Ondel, A.~W. Black \emph{et~al.}, ``{The Zero
  Resource Speech Challenge 2019: TTS without T},'' \emph{arXiv preprint
  arXiv:1904.11469}, 2019.

\bibitem{nguyen2020zero}
T.~A. Nguyen, M.~de~Seyssel, P.~Roz{\'e}, M.~Rivi{\`e}re, E.~Kharitonov,
  A.~Baevski, E.~Dunbar, and E.~Dupoux, ``{The Zero Resource Speech Benchmark
  2021}: Metrics and baselines for unsupervised spoken language modeling,''
  \emph{arXiv preprint arXiv:2011.11588}, 2020.

\bibitem{ondel2016variational}
L.~Ondel, L.~Burget, and J.~{\v{C}}ernock{\`y}, ``Variational inference for
  acoustic unit discovery,'' \emph{Procedia Computer Science}, vol.~81, pp.
  80--86, 2016.

\bibitem{heck2017feature}
M.~Heck, S.~Sakti, and S.~Nakamura, ``Feature optimized {DPGMM} clustering for
  unsupervised subword modeling: A contribution to {Z}ero{S}peech 2017,'' in
  \emph{Proceedings of IEEE Workshop on Automatic Speech Recognition and
  Understanding}.\hskip 1em plus 0.5em minus 0.4em\relax IEEE, 2017, pp.
  740--746.

\bibitem{tjandra19_interspeech}
A.~Tjandra, B.~Sisman, M.~Zhang, S.~Sakti, H.~Li, and S.~Nakamura, ``{VQVAE
  Unsupervised Unit Discovery and Multi-Scale Code2Spec Inverter for ZeroSpeech
  Challenge 2019},'' in \emph{Proceedings of the Annual Conference of the
  International Speech Communication Association}, 2019.

\bibitem{maekaku21_interspeech}
T.~Maekaku, X.~Chang, Y.~Fujita, L.-W. Chen, S.~Watanabe, and A.~Rudnicky,
  ``{Speech Representation Learning Combining Conformer CPC with Deep Cluster
  for the ZeroSpeech Challenge 2021},'' in \emph{Proceedings of the Annual
  Conference of the International Speech Communication Association}, 2021.

\bibitem{emotion_conversion}
F.~Kreuk, A.~Polyak, J.~Copet, E.~Kharitonov, T.~A. Nguyen, M.~Rivi{\`{e}}re,
  W.~Hsu, A.~Mohamed, E.~Dupoux, and Y.~Adi, ``Textless speech emotion
  conversion using decomposed and discrete representations,'' \emph{arXiv
  preprint arXiv:2111.07402}, 2021.

\bibitem{dialogue_GSLM}
T.~A. Nguyen, E.~Kharitonov, J.~Copet, Y.~Adi, W.-N. Hsu, A.~Elkahky,
  P.~Tomasello, R.~Algayres, B.~Sagot, A.~Mohamed, and E.~Dupoux, ``Generative
  spoken dialogue language modeling,'' 2022.

\bibitem{spoken_qa_dual}
G.-T. Lin, Y.-S. Chuang, H.-L. Chung, S.-w. Yang, H.-J. Chen, S.~Dong, S.-W.
  Li, A.~Mohamed, H.-y. Lee, and L.-s. Lee, ``{DUAL}: Discrete spoken unit
  adaptive learning for textless spoken question answering,'' \emph{CoRR},
  2022.

\bibitem{shi2021discretization}
J.~Shi, X.~Chang, T.~Hayashi, Y.-J. Lu, S.~Watanabe, and B.~Xu,
  ``Discretization and re-synthesis: an alternative method to solve the
  cocktail party problem,'' \emph{arXiv preprint arXiv:2112.09382}, 2021.

\bibitem{hayashi2020discretalk}
T.~Hayashi and S.~Watanabe, ``Discretalk: Text-to-speech as a machine
  translation problem,'' \emph{arXiv preprint arXiv:2005.05525}, 2020.

\bibitem{textless_translation}
A.~Lee, H.~Gong, P.~Duquenne, H.~Schwenk, P.~Chen, C.~Wang, S.~Popuri, J.~Pino,
  J.~Gu, and W.~Hsu, ``Textless speech-to-speech translation on real data,''
  \emph{CoRR}, 2021.

\bibitem{pmlr-v97-houlsby19a}
N.~Houlsby, A.~Giurgiu, S.~Jastrzebski, B.~Morrone, Q.~De~Laroussilhe,
  A.~Gesmundo, M.~Attariyan, and S.~Gelly, ``Parameter-efficient transfer
  learning for {NLP},'' in \emph{International Conference on Machine Learning},
  K.~Chaudhuri and R.~Salakhutdinov, Eds., vol.~97, 09--15 Jun 2019, pp.
  2790--2799.

\bibitem{zaken2021bitfit}
E.~B. Zaken, S.~Ravfogel, and Y.~Goldberg, ``{BitFit}: Simple
  parameter-efficient fine-tuning for {T}ransformer-based masked
  language-models,'' 2021.

\bibitem{guo-etal-2021-parameter}
\BIBentryALTinterwordspacing
D.~Guo, A.~Rush, and Y.~Kim, ``Parameter-efficient transfer learning with diff
  pruning,'' in \emph{Proceedings of the 59th Annual Meeting of the Association
  for Computational Linguistics and the 11th International Joint Conference on
  Natural Language Processing (Volume 1: Long Papers)}.\hskip 1em plus 0.5em
  minus 0.4em\relax Online: Association for Computational Linguistics, Aug.
  2021, pp. 4884--4896. [Online]. Available:
  \url{https://aclanthology.org/2021.acl-long.378}
\BIBentrySTDinterwordspacing

\bibitem{SpeechAdapter}
B.~Thomas, S.~Kessler, and S.~Karout, ``Efficient adapter transfer of
  self-supervised speech models for automatic speech recognition,'' in
  \emph{Proceedings of {IEEE} International Conference on Acoustics, Speech and
  Signal Processing}, 2022.

\bibitem{SpeechPrompt}
K.-W. Chang, W.-C. Tseng, S.-W. Li, and H.~yi~Lee, ``{SpeechPrompt}: An
  exploration of prompt tuning on generative spoken language model for speech
  processing tasks,'' \emph{Proceedings of the Annual Conference of the
  International Speech Communication Association}, 2022.

\bibitem{PARP}
C.-I.~J. Lai, Y.~Zhang, A.~H. Liu, S.~Chang, Y.-L. Liao, Y.-S. Chuang, K.~Qian,
  S.~Khurana, D.~Cox, and J.~Glass, ``{PARP}: Prune, adjust and re-prune for
  self-supervised speech recognition,'' in \emph{Proceedings of Advances in
  Neural Information Processing Systems}, 2021.

\bibitem{chang2021distilhubert}
H.-J. Chang, S.~wen Yang, and H.~yi~Lee, ``{DistilHuBERT}: Speech
  representation learning by layer-wise distillation of hidden-unit {BERT},''
  2021.

\bibitem{bengio_conditional_2016}
E.~Bengio, P.-L. Bacon, J.~Pineau, and D.~Precup, ``Conditional {{Computation}}
  in {{Neural Networks}} for faster models,'' \emph{arXiv:1511.06297 [cs]},
  Jan. 2016.

\bibitem{gholami_survey_2021}
A.~Gholami, S.~Kim, Z.~Dong, Z.~Yao, M.~W. Mahoney, and K.~Keutzer, ``A
  {{Survey}} of {{Quantization Methods}} for {{Efficient Neural Network
  Inference}},'' Jun. 2021.

\bibitem{tay_efficient_2022}
Y.~Tay, M.~Dehghani, D.~Bahri, and D.~Metzler, ``Efficient {{Transformers}}:
  {{A Survey}},'' Mar. 2022.

\bibitem{StreamingW2v}
S.~Cao, Y.~Kang, Y.~Fu, X.~Xu, S.~Sun, Y.~Zhang, and L.~Ma, ``Improving
  streaming {T}ransformer based {ASR} under a framework of self-supervised
  learning,'' in \emph{Proceedings of the Annual Conference of the
  International Speech Communication Association}, 2021.

\bibitem{pmlr-v162-qian22b}
K.~Qian, Y.~Zhang, H.~Gao, J.~Ni, C.-I. Lai, D.~Cox, M.~Hasegawa-Johnson, and
  S.~Chang, ``{C}ontent{V}ec: An improved self-supervised speech representation
  by disentangling speakers,'' in \emph{Proceedings of International Conference
  on Machine Learning}, 2022.

\bibitem{NANCY}
H.-S. Choi, J.~Lee, W.~Kim, J.~H. Lee, H.~Heo, and K.~Lee, ``Neural analysis
  and synthesis: Reconstructing speech from self-supervised representations,''
  in \emph{NeurIPS}, 2021.

\bibitem{MultimodelAmazonIS22}
D.~M. Chan and S.~Ghosh, ``Content-context factorized representations for
  automated speech recognition,'' in \emph{Proceedings of the Annual Conference
  of the International Speech Communication Association}, 2022.

\bibitem{wu2021characterizing}
H.~Wu, B.~Zheng, X.~Li, X.~Wu, H.~yi~Lee, and H.~Meng, ``Characterizing the
  adversarial vulnerability of speech self-supervised learning,'' 2021.

\bibitem{huang2022improving}
K.~P. Huang, Y.-K. Fu, Y.~Zhang, and H.-y. Lee, ``Improving distortion
  robustness of self-supervised speech processing tasks with domain
  adaptation,'' in \emph{Proceedings of the Annual Conference of the
  International Speech Communication Association}, 2022.

\bibitem{wang2022improving}
H.~Wang, Y.~Qian, X.~Wang, Y.~Wang, C.~Wang, S.~Liu, T.~Yoshioka, J.~Li, and
  D.~Wang, ``Improving noise robustness of contrastive speech representation
  learning with speech reconstruction,'' in \emph{ICASSP 2022-2022 IEEE
  International Conference on Acoustics, Speech and Signal Processing
  (ICASSP)}.\hskip 1em plus 0.5em minus 0.4em\relax IEEE, 2022, pp. 6062--6066.

\bibitem{zhu2022noise}
Q.-S. Zhu, J.~Zhang, Z.-Q. Zhang, M.-H. Wu, X.~Fang, and L.-R. Dai, ``A
  noise-robust self-supervised pre-training model based speech representation
  learning for automatic speech recognition,'' in \emph{ICASSP 2022-2022 IEEE
  International Conference on Acoustics, Speech and Signal Processing
  (ICASSP)}.\hskip 1em plus 0.5em minus 0.4em\relax IEEE, 2022, pp. 3174--3178.

\bibitem{TuAnh_2022_discrete}
T.~A. Nguyen, B.~Sagot, and E.~Dupoux, ``Are discrete units necessary for
  spoken language modeling?'' 2022.

\end{thebibliography}

\begin{IEEEbiography}
[{\includegraphics[width=1in,height=1.25in,clip,keepaspectratio]{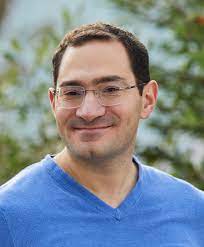} }]
{Abdelrahman Mohamed} is a research scientist at Fundamental AI Research (FAIR), Meta. He was a principal scientist/manager in Amazon Alexa and a researcher in Microsoft Research. Abdelrahman was in the team that started the Deep Learning revolution in Spoken Language Processing in 2009 and received the IEEE Signal Processing Society Best Paper Award in 2016. He has been focusing on improving speech representation learning, where he co-organized multiple workshops, tutorials, and special sessions.
\end{IEEEbiography}

\begin{IEEEbiography}
[{\includegraphics[width=1in,height=1.25in,clip,keepaspectratio]{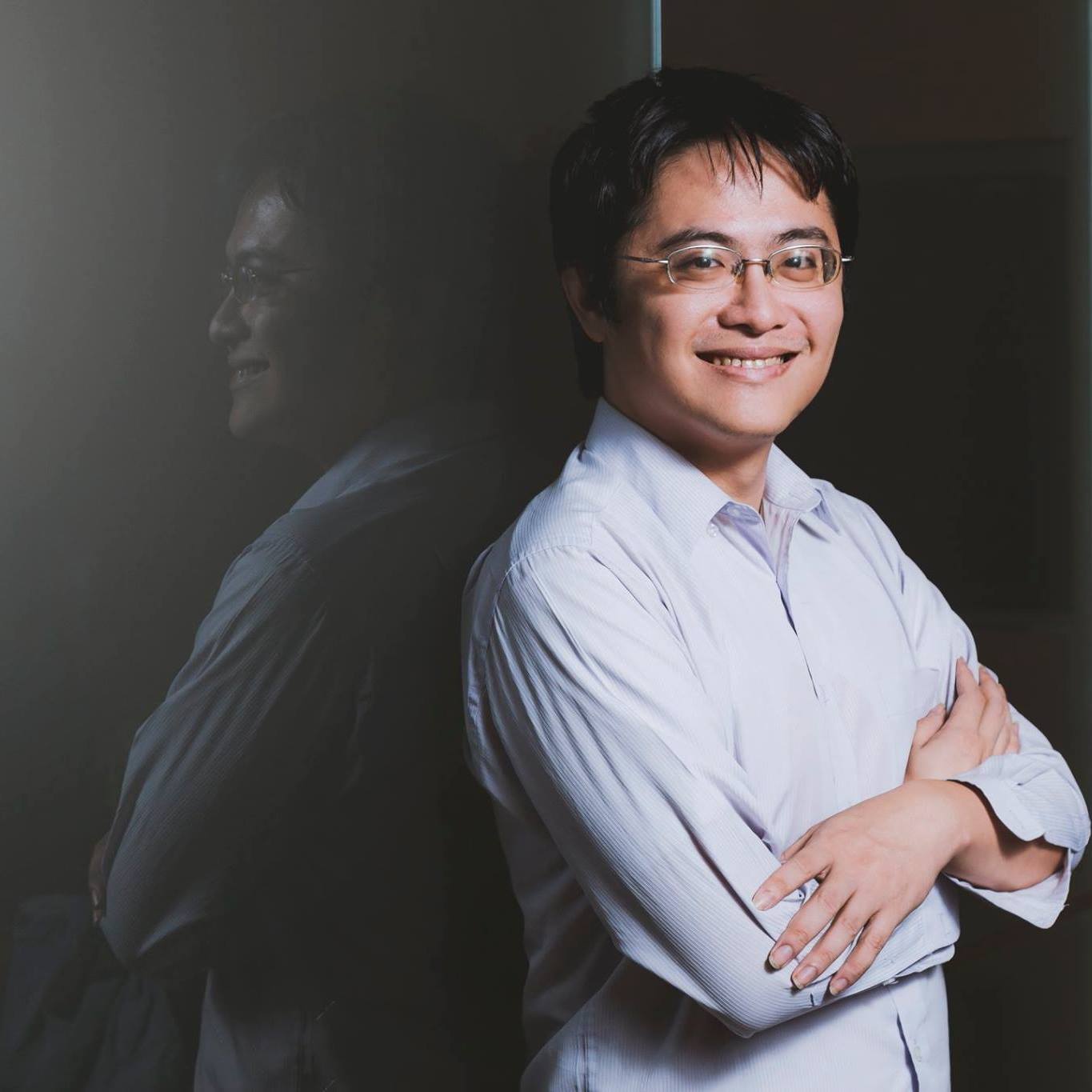} }]
{Hung-yi Lee} is an associate professor in the Department of Electrical Engineering of National Taiwan University (NTU), with a joint appointment at the Department of Computer Science \& Information Engineering. He is the co-organizer of the special session on "New Trends in self-supervised speech processing" at Interspeech (2020) and the workshop on "Self-Supervised Learning for Speech and Audio Processing" at NeurIPS (2020) and AAAI (2022).
\end{IEEEbiography}

\begin{IEEEbiography}
[{\includegraphics[width=1in,height=1.25in,clip,keepaspectratio]{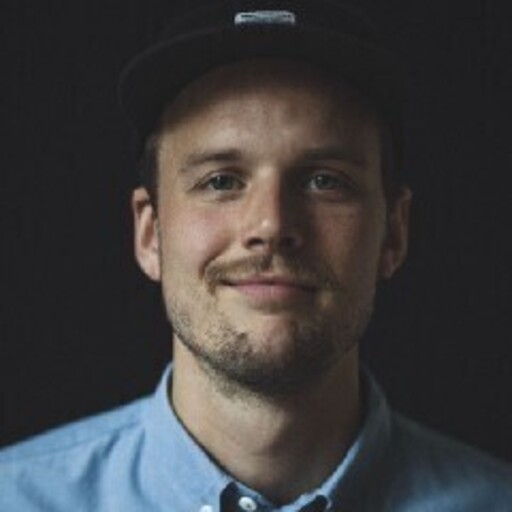} }]
{Lasse Borgholt} received his Ph.D. degree from University of Copenhagen in 2022. He was supervised by Professors Christian Igel and Anders Søgaard and Adjunct Associate Professor Lars Maaløe. He currently works as an industrial machine learning researcher at the Danish company Corti. His primary research interests are speech recognition and representation learning for speech. 
\end{IEEEbiography}

\begin{IEEEbiography}
[{\includegraphics[width=1in,height=1.25in,clip,keepaspectratio]{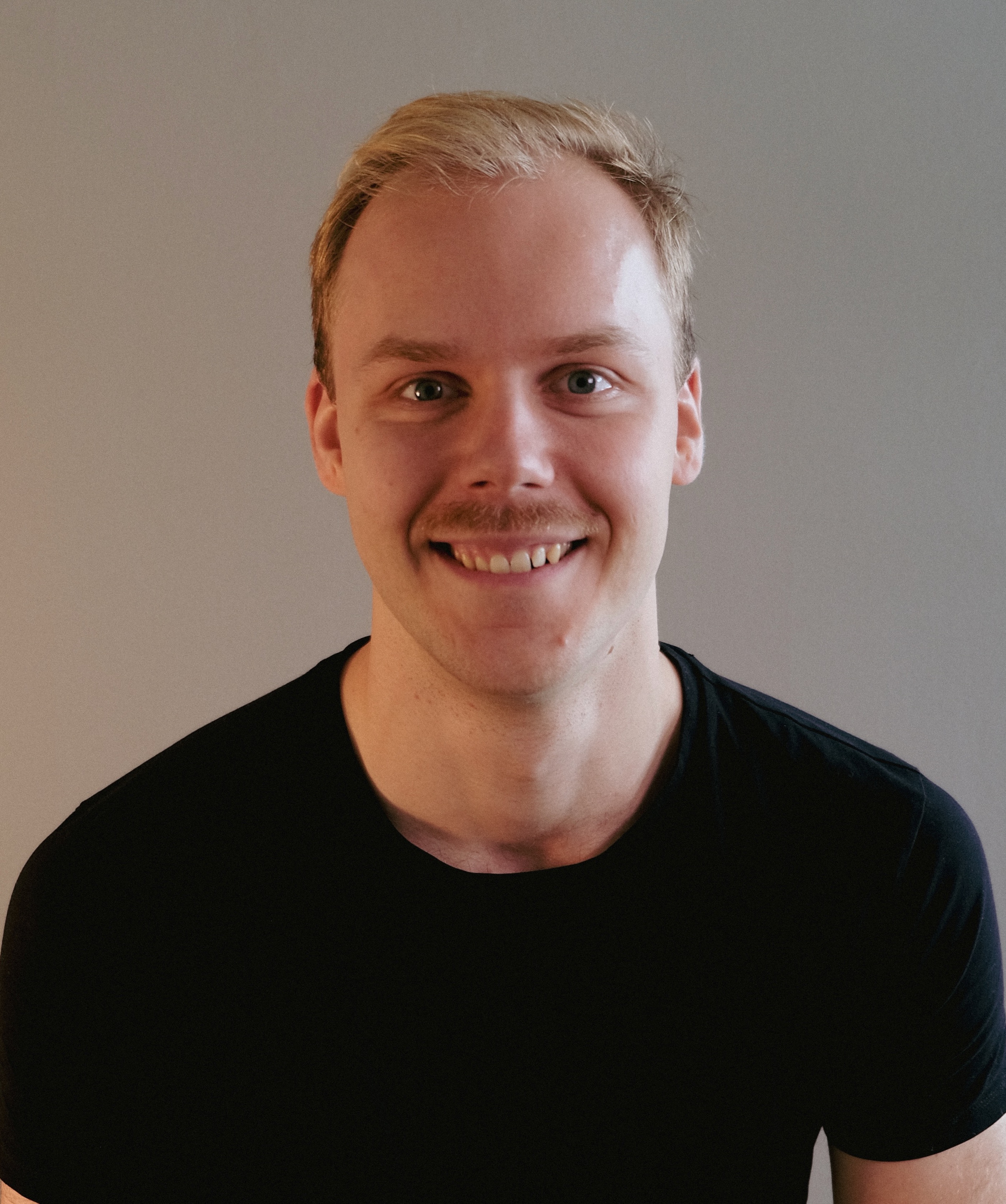}}]
{Jakob D. Havtorn} received his bachelor's degree in physics and master's degree in mathematical modeling and computation from the Technical University of Denmark (DTU), Copenhagen, in 2016 and 2018, respectively. 
He is currently an industrial Ph.D. student at Corti and DTU, supervised by Associate Professor Jes Frellsen, Adjunct Associate Professor Lars Maaløe and Professors Søren Hauberg and Ole Winther. 
His primary research interests are deep generative latent variable models, uncertainty quantification and representation learning for speech.
\end{IEEEbiography}

\begin{IEEEbiography}
[{\includegraphics[width=1in,height=1.25in,clip,keepaspectratio]{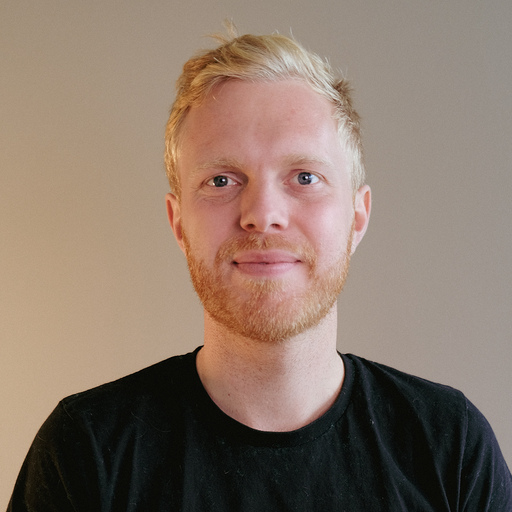} }]
{Joakim Edin} received his master's degree in biomedical engineering from the Technical University of Denmark, Copenhagen, in 2019. He worked as an industrial machine learning researcher at the Danish company Corti prior to starting in his current occupation as a Ph.D. student at University of Copenhagen. He is supervised by Professor Søren Brunak and Adjunct Associate Professor Lars Maaløe. His primary research interests are representation learning from medical conversations and text.
\end{IEEEbiography}

\begin{IEEEbiography}
[{\includegraphics[width=1in,height=1.25in,clip,keepaspectratio]{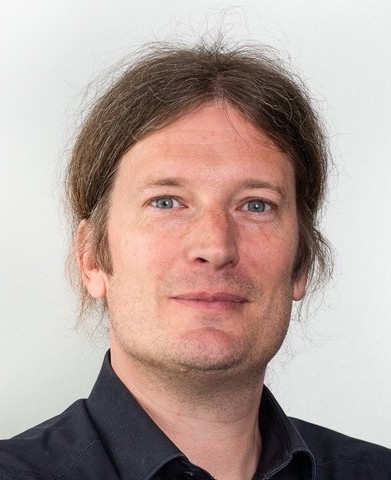} }]
{Christian Igel} received the Diploma degree in computer science from the Technical University Dortmund, Germany, in 1997,
the Dr.~rer.~nat.~degree from the Ruhr-University Bochum, Germany, in 2002, and the Habilitation degree from the Ruhr-University Bochum, Germany, in 2010.
From 2003 to 2010, he was a Juniorprofessor at the Institute for Neural Computation, Ruhr-University Bochum, Germany.
He joined DIKU, the Department of Computer Science at the University of Copenhagen (UCPH), Denmark, in 2010.
He is a full professor at DIKU and currently the Director of the SCIENCE AI Centre at UCPH. 
He is a Fellow of the European Lab for Learning and Intelligent Systems (ELLIS).
\end{IEEEbiography}

\begin{IEEEbiography}
[{\includegraphics[width=1in,height=1.25in,clip,keepaspectratio]{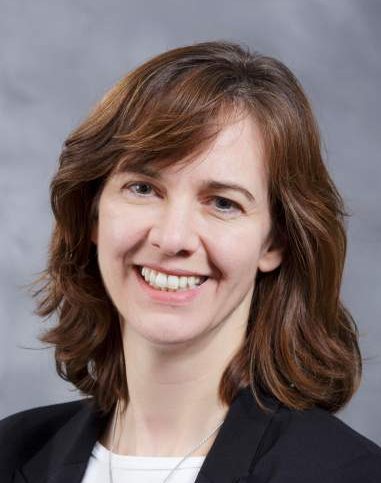} }]
{Katrin Kirchhoff} is a Director of Applied Science at Amazon Web Services, where she heads several teams in speech and audio processing. She was a Research Professor at the UW, Seattle, for 17 years, where she co-founded the Signal, Speech and Language Interpretation Lab. She served on the editorial boards of Speech Communication and Computer, Speech, and Language, and was a member of the IEEE Speech Technical Committee.
\end{IEEEbiography}

\begin{IEEEbiography}
[{\includegraphics[width=1in,height=1.25in,clip,keepaspectratio]{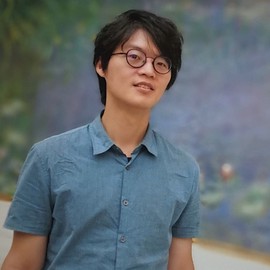} }]
{Shang-Wen Li} is a Research and Engineering Manager at Meta AI. He worked at Apple Siri, Amazon Alexa and AWS. He completed his PhD in 2016 from the Spoken Language Systems group of MIT CSAIL. He co-organized the workshop of "Self-Supervised Learning for Speech and Audio Processing" at NeurIPS (2020) and AAAI (2022). His recent research focuses on self-supervised learning in speech and its application to language understanding.
\end{IEEEbiography}

\begin{IEEEbiography}
[{\includegraphics[width=1in,height=1.25in,clip,keepaspectratio]{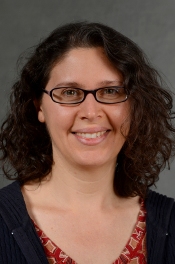} }]
{Karen Livescu}  is a Professor at TTI-Chicago. She completed her PhD at MIT in the Spoken Language Systems group. She is an ISCA Fellow and an IEEE Distinguished Lecturer, and has served as a program chair for ICLR, Interspeech, and ASRU. She works on a variety of topics in speech and language processing and machine learning.  She has co-organized multiple workshops on machine learning for speech processing, multi-view representation learning, and self-supervised representation learning. 
\end{IEEEbiography}

\begin{IEEEbiography}
[{\includegraphics[width=1in,height=1.25in,clip,keepaspectratio]{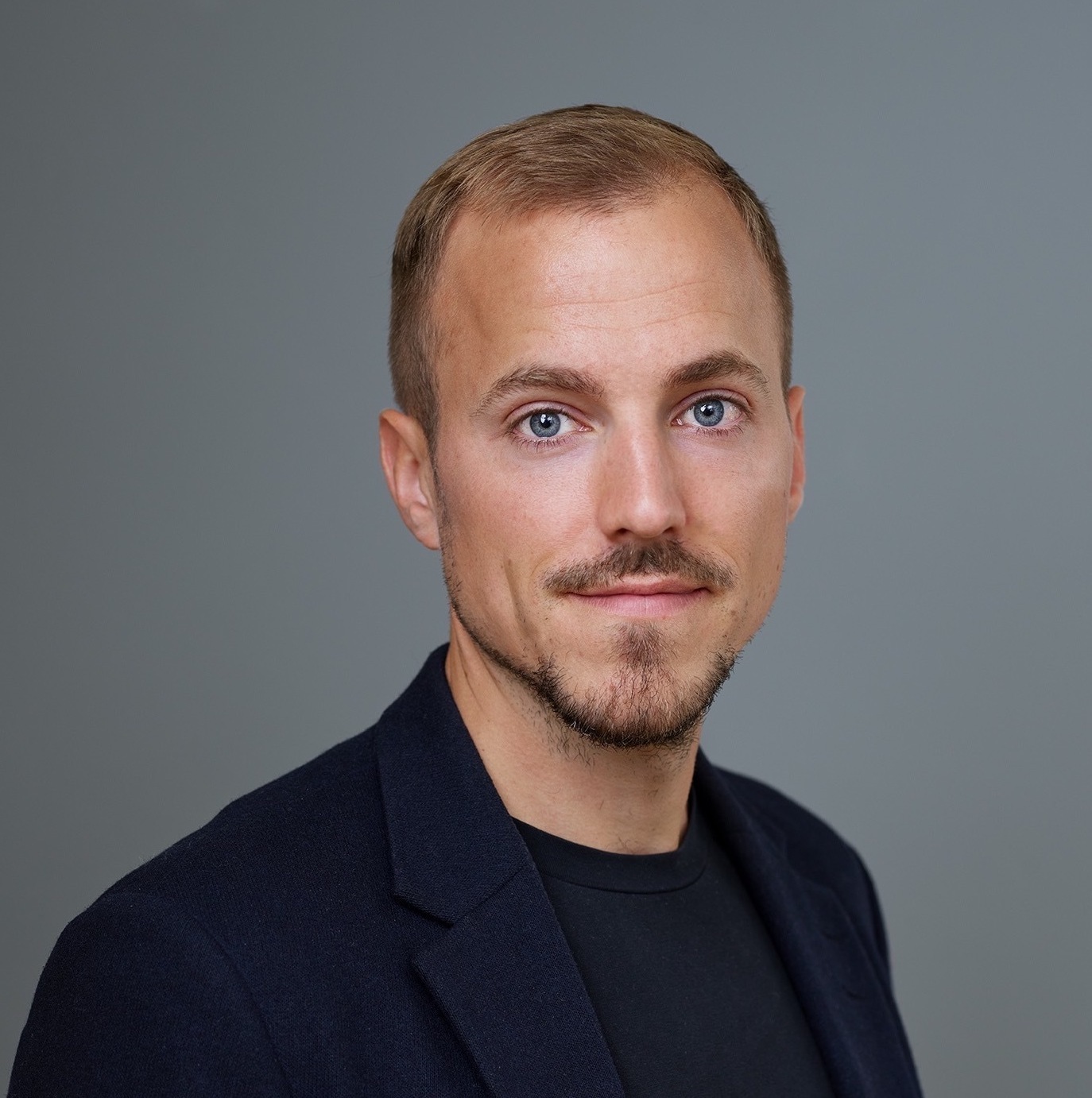} }]
{Lars Maaløe} received his Ph.D. degree from the Technical University of Denmark (DTU), Copenhagen, in 2018. He was nominated Ph.D. of the year by the Department for Applied Mathematics and Computer Science. In the past, he worked in the financial sector and with various tech companies such as Apple. He is a co-founder and the current CTO of Corti, a voice-based digital assistant powered by artificial intelligence. He also holds an appointment as Adjunct Associate Professor at DTU.
\end{IEEEbiography}

\begin{IEEEbiography}
[{\includegraphics[width=1in,height=1.25in,clip,keepaspectratio]{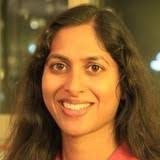} }]
{Tara N. Sainath} is a Principal Research scientist at Google. She received her PhD from MIT in the Spoken Language Systems Group. She is an IEEE and ISCA Fellow and the recipient of the 2021 IEEE SPS Industrial Innovation Award. Her research involves applications of deep neural networks for automatic speech recognition, and has been very active in the community organizing workshops and special sessions on this topic.
\end{IEEEbiography}

\begin{IEEEbiography}
[{\includegraphics[width=1in,height=1.25in,clip,keepaspectratio]{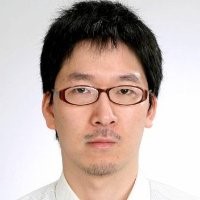} }]
{Shinji Watanabe} is an Associate Professor at CMU. He was a research scientist at NTT, Japan, a visiting scholar in Georgia Tech, a senior principal research scientist at MERL, and an associate research professor at JHU. He has published over 300 peer-reviewed papers. He serves as a Senior Area Editor of the IEEE TASLP. He was/has been a member of several technical committees, including the APSIPA SLA, IEEE SPS SLTC, and MLSP.
\end{IEEEbiography}

\end{document}